\documentclass{article}

\usepackage{microtype}
\usepackage{graphicx}
\usepackage{booktabs} 

\usepackage{hyperref}

\usepackage[accepted]{icml2023}

\usepackage{amsmath}
\usepackage{amssymb}
\usepackage{mathtools}
\usepackage{amsthm}

\usepackage[capitalize,noabbrev]{cleveref}

\theoremstyle{plain}

\theoremstyle{definition}

\theoremstyle{remark}

\usepackage[textsize=tiny]{todonotes}

\icmltitlerunning{Total Variation Graph Neural Networks}

\usepackage{caption}
\usepackage{subcaption}
\usepackage{stix}

\makeatletter
\newcommand*\myfootnotesize{%
  \@setfontsize\myfootnotesize{8.8}{10.56}%
}
\makeatother

\newcommand{\meanstd}[2]{
$\text{#1} {\scriptstyle \pm \text{#2}}$}

\newcommand\s{\boldsymbol s}
\newcommand\cb{\boldsymbol c}
\newcommand\y{\boldsymbol y}
\newcommand\x{\boldsymbol x}
\newcommand\z{\boldsymbol z}
\newcommand\Sb{\mathbf S}
\newcommand\Lb{\mathbf L}
\newcommand\D{\mathbf D}
\newcommand\A{\mathbf A}
\newcommand\X{\mathbf X}
\newcommand\I{\mathbf I}
\newcommand\R{\mathbb R}

\definecolor{red}{HTML}{B81919}
\definecolor{black}{HTML}{000000}
\definecolor{gray}{HTML}{929090}

\begin{document}

\twocolumn[
\icmltitle{Total Variation Graph Neural Networks}



\icmlsetsymbol{equal}{*}

\begin{icmlauthorlist}
\icmlauthor{Jonas Berg Hansen}{equal,uit}
\icmlauthor{Filippo Maria Bianchi}{equal,uit,nor}
\end{icmlauthorlist}

\icmlaffiliation{uit}{Department of Mathematics and Statistics, UiT the Arctic University of Norway}
\icmlaffiliation{nor}{NORCE, The Norwegian Research Centre AS}

\icmlcorrespondingauthor{Filippo Maria Bianchi}{filippo.m.bianchi@uit.no}

\icmlkeywords{Graph neural networks, graph pooling, graph coarsening, graph clustering}

\vskip 0.3in
]



\printAffiliationsAndNotice{\icmlEqualContribution} 

\begin{abstract}
Recently proposed Graph Neural Networks (GNNs) for vertex clustering are trained with an unsupervised minimum cut objective, approximated by a Spectral Clustering (SC) relaxation. 
However, the SC relaxation is loose and, while it offers a closed-form solution, it also yields overly smooth cluster assignments that poorly separate the vertices. 
In this paper, we propose a GNN model that computes cluster assignments by optimizing a tighter relaxation of the minimum cut based on graph total variation (GTV). 
The cluster assignments can be used directly to perform vertex clustering or to implement graph pooling in a graph classification framework. 
Our model consists of two core components: 
i) a message-passing layer that minimizes the $\ell_1$ distance in the features of adjacent vertices, which is key to achieving sharp transitions between clusters; 
ii) an unsupervised loss function that minimizes the GTV of the cluster assignments while ensuring balanced partitions.  
Experimental results show that our model outperforms other GNNs for vertex clustering and graph classification.


\end{abstract}


\section{Introduction}
\label{sec: introduction}
\begin{figure}[!t]
    \centering
    \begin{subfigure}{\columnwidth}
        \centering
        \includegraphics[width=.75\textwidth]{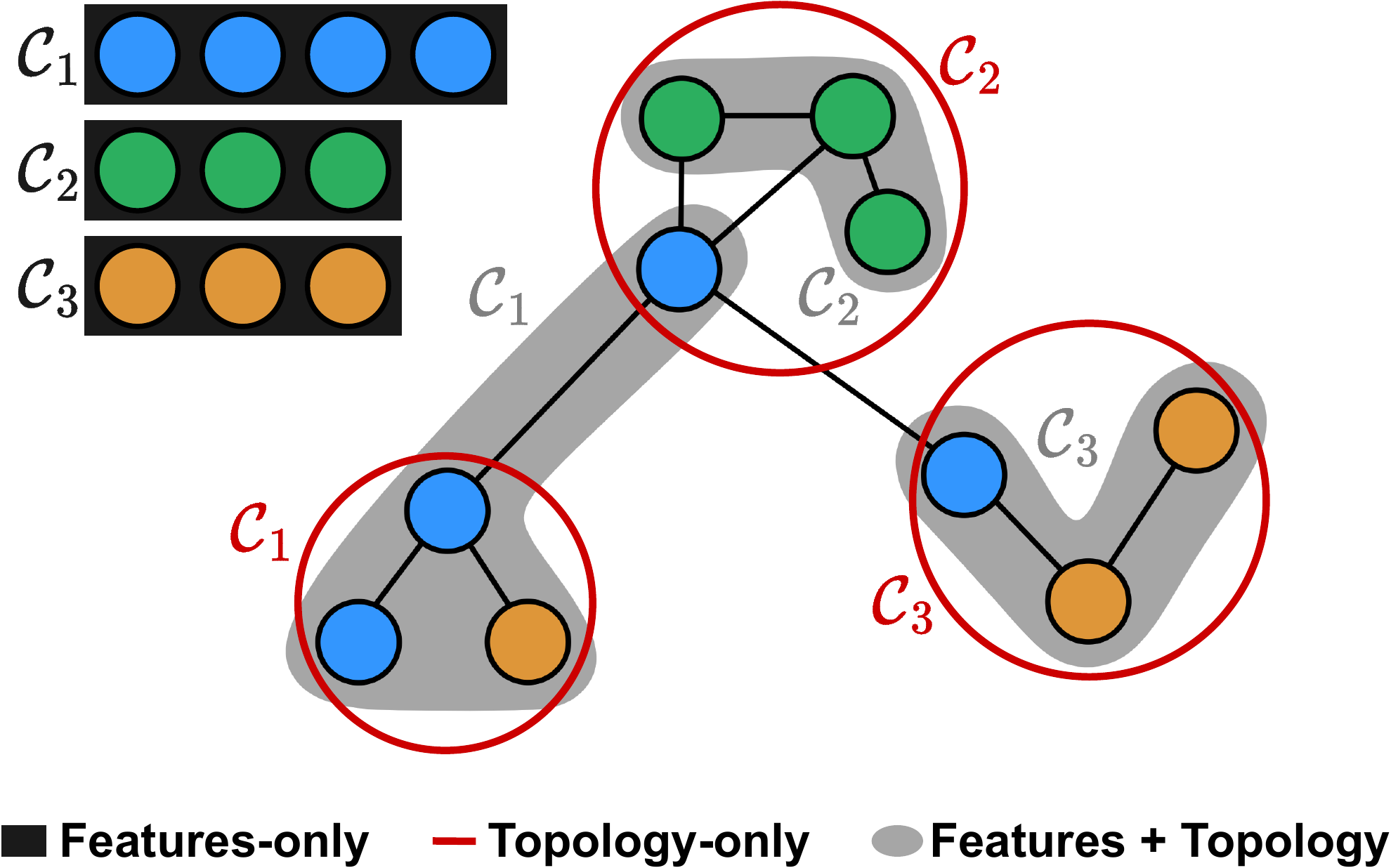}
        \caption{}
    \end{subfigure}
    \begin{subfigure}{\columnwidth}
        \centering
        \includegraphics[width=.65\textwidth]{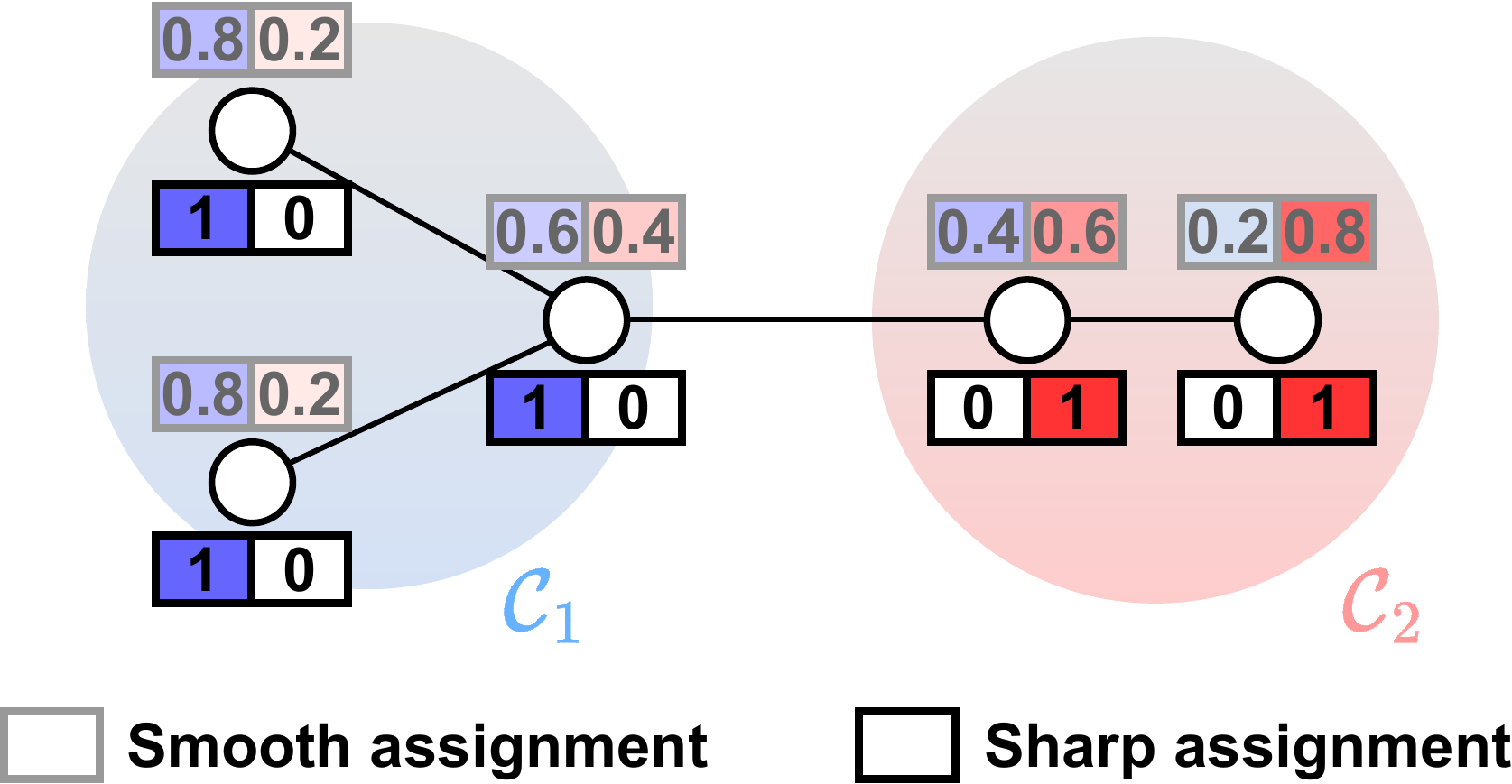}
        \caption{}
    \end{subfigure}
    \caption{\small \textbf{a)} Most clustering methods partition the data only based on the features (${\color{black}\mdblksquare}$). SC partitions the vertices of the graph based on its topology (${\color{red}\mdblksquare}$). GNNs account both for the vertex features and the graph topology(${\color{gray}\mdblksquare}$). \textbf{b)} Difference between smooth and sharp cluster assignments. Especially at the edge of a cluster, the smooth assignments give large weights to more than one cluster.}
    \label{fig:intro_sketch}
\end{figure}

Traditional clustering techniques partition samples based on their features or on suitable data representations computed, for example, with deep learning models~\citep{tian2014learning, min2018survey, su2022comprehensive}.
Spectral clustering (SC)~\citep{von2007tutorial} is a popular technique that first encodes the similarity of the data features into a graph and then creates a partition based on the graph topology.
Such a graph is just a convenient representation of the similarity among the samples and has no attributes on its vertices.
On the other hand, an attributed graph can represent both the relationships among samples and their features.
Graph Neural Networks (GNNs) are deep learning architectures specifically designed to process and make inference on such data~\citep{hamilton2020graph}.
Therefore, contrarily to traditional clustering methods, a GNN-based approach for clustering can account for both the features and the relationships among samples to generate partitions (see Fig.\ref{fig:intro_sketch}a).

Similarly to other deep learning architectures for clustering~\citep{shaham2018spectralnet, kampffmeyer2019deep}, GNNs can be trained end-to-end and return soft cluster assignments as part of the output.
Several existing GNN clustering approaches compute cluster assignments from the vertex representations generated by message passing (MP) layers and, then, optimize the assignments with an unsupervised loss inspired by SC~\citep{bianchi2020spectral, tsitsulin2020graph, duval2022higher}.
The SC objective benefits from the smoothing operations performed by the MP layers, which minimize the local quadratic variation of adjacent vertex features.
However, this approach produces smooth cluster assignment vectors that are less informative as they do not separate well the samples (see Fig.\ref{fig:intro_sketch}b).
Indeed, the SC objective is known to give a loose approximation of the optimal partition defined in terms of the minimum cut~\citep{rangapuram2014tight}.

\textbf{Contributions.}
We propose a novel GNN-based clustering approach that generates cluster assignments by optimizing the graph total variation (GTV). 
Compared to SC, GTV trades a closed-form solution with a tighter continuous relaxation of the optimal minimum cut partition~\citep{hein2011beyond}.
Notably, an optimization objective with a closed-form solution is not particularly useful for a GNN trained with gradient descent.

We design an unsupervised loss function for minimizing the GTV that is GNN-friendly as it avoids numerical issues  during gradient descent and assumes values in a well-defined range, making it suitable to be combined with other losses.
Our GNN yields sharp cluster assignments by minimizing the $\ell_1$ norm of their differences.
Clearly, the minimization of GTV is hindered when cluster assignments are derived from smooth representations computed by traditional MP layers.
To address this issue, we propose a new MP layer that minimizes the $\ell_1$ norm of the difference between adjacent vertex features. 

By optimizing the proposed loss, we can train a GNN end-to-end to perform vertex clustering. 
In addition, by coarsening the graph according to the learned partition we can perform hierarchical graph pooling~\citep{grattarola2022understanding} in deep GNN architectures for graph-level tasks, such as graph classification.
In this case, the proposed loss is combined with an additional supervised loss, such as the cross-entropy.
Experiments show the superiority of the proposed approach compared to other GNN methods for clustering and graph pooling.

\section{Background}
A graph is represented by a tuple $\mathcal{G}=(\mathcal{V}, \mathcal{E})$ where $\mathcal{V}$ and $\mathcal{E}$ are the vertex and edge sets, respectively. The cardinality of the sets are given as $|\mathcal{V}|=N$ and $|\mathcal{E}|=E$. 
The adjacency matrix $\mathbf{A}\in\mathbb{R}^{N\times N}$ with elements $a_{ij}\in\{0, 1\}$ defines the graph connectivity. In an attributed graph, each vertex $i$ is associated with a feature vector $\x_i \in\mathbb{R}^F$. Feature vectors are often grouped in a matrix $\mathbf{X}\in\mathbb{R}^{N\times F}$. 
The soft cluster assignment matrix is $\mathbf{S}\in\mathbb{R}^{N\times K}$, where $K$ is the number of clusters and $s_{ij}\in [0, 1]$ is the membership of vertex $i$ to cluster $j$.
The combinatorial Laplacian is $\Lb = \D - \A$ and $\mathbf{\tilde{A}}=\mathbf{D}^{-1/2}\mathbf{A}\mathbf{D}^{-1/2}$ is the symmetric degree normalization of $\A$.

\subsection{Graph cuts}
The task of finding $K$ clusters of similar size can be cast into the balanced $K$-cut problem, defined as a ratio of two set functions:
\begin{equation}
    \label{eq:balanced_k_cut}
    \mathcal{C} = \min_{C_1, \dots, C_K} \sum_{k=1}^K \frac{\text{cut}(C_k, \bar{C}_k)}{\hat{S}(C_k)} \;\; \text{s.t.}\;\; C_i \cap C_j = \emptyset,
\end{equation}
where $\text{cut}(C_i, C_j)$ counts the volume of edges connecting the two sets of vertices $C_i, C_j \subset \mathcal{V}$, $\bar{C_k}$ is the complement of set $C_k$, and $\hat{S}(\cdot): 2^\mathcal{V}\to\mathbb{R}_+$ is a submodular set function that balances the size of the clusters in the partition~\citep{hein2011beyond}. 
Depending on the choice of $\hat{S}(\cdot)$ different cuts are obtained, such as the ratio cut for $\hat{S}(C_k)=|C_k|$, and the normalized cut for $\hat{S}(C_k)=\text{vol}(C_k)$~\citep{von2007tutorial}. 
Of particular interest for us is the Cheeger cut, also known as Balanced cut or Ratio Cheeger cut, where $\hat{S}(C_k)=\min\{|C_k|, |\bar{C_k}|\}$, which penalizes the formation of very large clusters. 
A variant of the Cheeger Cut, called \textit{Asymmetric Cheeger cut}, encourages an even partition by letting $\hat{S}(C_k)=\min\{(K-1)|C_k|, |\bar{C_k}|\}$~\citep{bresson2013multiclass}.

\subsection{Tight relaxation of the Asymmetric Cheeger cut}
Let the numerator in ~\eqref{eq:balanced_k_cut} be expressed in matrix form as
\begin{equation}
    \text{cut}(C_k, \bar{C}_k) = \mkern-18mu \sum \limits_{i \in C_k, j \in \bar C_k}\mkern-18mu a_{ij}(1- z_i z_j) = \z^T \Lb \z,
\end{equation}
with $z_i, z_j \in \{ -1, 1 \}$ (derivation in \ref{appendix:cut}).
The common relaxation done in spectral clustering (SC) is:
\begin{equation}
    \label{eq:general_sc_relax}
    \min_{\z \in \{ -1, 1 \}^{N}} \frac{\z^T \Lb \z}{\hat{S}(C_k)} \rightarrow \min_{\s \in \R^{N}} \frac{\s^T \Lb \s}{S(C_k)}
\end{equation}
where $S(C_k)$ is the continuous counterpart of $\hat{S}(C_k)$.
What the SC relaxation actually does, is to apply Laplacian smoothing to a graph signal $\s$ by minimizing its \textit{local quadratic variation} (LQV), i.e., the quadratic variation of $\s$ across adjacent vertices.
The LQV defined in terms of the combinatorial Laplacian reads
\begin{equation}
    \label{eq:lap_smooth_comb}
   \s^T\Lb\s = \frac{1}{2} \sum_{(i,j) \in \mathcal{E}} a_{ij}(s_i - s_j)^2.
\end{equation}
From the graph signal processing perspective, Laplacian smoothing applies to a graph signal a low-pass filter with response $(1-\lambda_i)$, where $\lambda_i$ is the $i$-th eigenvalue of the Laplacian~\citep{tremblay2018design}. Despite offering a closed-form solution, SC gives a loose approximation of the solution of the discrete optimization problem~\citep{hein2011beyond}.

Let $\s_k\in\mathbb{R}^{N}$ be the soft assignment vector for the $k$-th cluster.
A tighter continuous relaxation of the Asymmetric Cheeger cut problem is given by~\cite{bresson2013multiclass}:
\begin{equation}
    \label{eq:relaxed_asymcheeger}
    \min_{\s_1, \dots, \s_K \in\mathbb{R}^N}\sum_{k=1}^K \frac{||\s_{k}||_\text{GTV}}{||\s_{k} - \textrm{quant}_\rho (\s_{k})||_{1, \rho}}\;\text{s.t.}\;\sum_{k=1}^K \s_k = \boldsymbol{1}_N.
\end{equation}
In the numerator, $||\s_k||_\text{GTV} = \sum_{i, j} a_{ij}|s_{i, k} - s_{j, k}|$ measures the graph total variation (GTV) of the soft assignments to cluster $k$. 
In the denominator, $\text{quant}_\rho(\s_k)$ denotes the $\rho$-quantile of $\s_k$, i.e., the $(q+1)^{\small\textrm{st}}$ largest value in $\s_k$ with $q = \lfloor N/(\rho + 1) \rfloor$, while $||\cdot||_{1,\rho}$ denotes an asymmetric $\ell_1$ norm, which for a vector $\x\in\mathbb{R}^{N}$ is defined as
\begin{equation}
 \label{eq:asymmetric_l1_norm}
 ||\x||_{1,\rho} = \sum_{i=1}^N |x_{i}|_\rho, \,\textrm{where}\, |x_i|_\rho = \begin{cases}\rho x_i, & x_i\geq 0\\ -x_i, & x_i < 0 \end{cases}. 
\end{equation}

When $\rho=K-1$, the denominator in \eqref{eq:relaxed_asymcheeger} encourages balanced partitions, i.e., clusters having similar sizes, and prevents the two common \textit{degenerate solutions}: i) all samples collapsing into the same cluster (e.g., $\s_i = [1, 0, \dots, 0], \forall i$), and ii) samples uniformly assigned to all clusters ($\s_i = [1/K, 1/K, \dots, 1/K], \forall i$).

Differently from SC, \eqref{eq:relaxed_asymcheeger} minimizes the $\ell_1$ rather than the $\ell_2$ distance between components of $\s$ that are adjacent on the graph. Minimizing GTV yields sharper cluster assignments and achieves a tighter relaxation of the balanced $K$-cut problem compared to SC~\citep{rangapuram2014tight, bresson2013multiclass}.
However, contrarily to SC, \eqref{eq:relaxed_asymcheeger} it is non-convex and is optimized through iterative updates.

\subsection{Clustering with Graph Neural Networks}
\label{sec:gnn_background}

To leverage the graph topology for learning vertex representations, GNNs implement message-passing (MP) layers. Typically, an MP layer collects information from the neighbours to update the representation of each vertex~\citep{hamilton2020graph}. 
An example of an MP layer is the Graph Convolutional Network (GCN) by \cite{kipf2016semi}:
\begin{equation}
    \mathbf{X}^{(\text{out})} = \sigma(\mathbf{\tilde{A}'}\mathbf{X}^{(\text{in})}\mathbf{\Theta}_\text{MP})
    \label{eq: gcn_layer}
\end{equation}
where $\mathbf{A'} = \A + \I_N$ and $\mathbf{\Theta}_\text{MP}\in\mathbb{R}^{F_\text{in}\times F_\text{out}}$ are learnable parameters. It can be shown that a GCN layer applies Laplacian smoothing on the vertex features $\mathbf{X}$; see~\cite{wu2019simplifying, nt2019revisiting, li2018deeper, bianchi2021graph} for detailed discussions. 
For example, \cite{ma2021unified} shows that the $l$-th GCN layer updates of the vertex features as
\begin{equation}
    \label{eq:gcn_update}
    \X^{(l+1)} = (\I - \delta \mathbf{\tilde A}')\X^{(l)},
\end{equation}
which is actually one gradient descent step of size $\delta$ in the optimization of the problem
\begin{equation}
    \label{eq:prob1}
    \min_{\X} \text{Tr}(\X^T(\mathbf{I} - \mathbf{\tilde A}')\X).
\end{equation}
%

After applying a stack of $L$ MP layers, the soft assignment matrix $\Sb$ is obtained by passing the updated vertex features to some function $f$, typically a multi-layer perceptron (MLP) with a \texttt{Softmax} activation, that produces a $K$-dimensional vector for each vertex.
Importantly, since the MP layers perform Laplacian smoothing and $f$ is a smooth function, the cluster assignments generated from $\X^{(L)}$ comply with the SC definition.
Nevertheless, minimizing the LQV of $\X^{(L)}$ is not the same as minimizing the LQV of $\Sb$ directly.
In addition, the size of the clusters must be balanced and degenerate solutions avoided.
For these reasons, the assignments $\Sb$ are further optimized through unsupervised loss functions. 
For instance, MinCutPool \citep{bianchi2020spectral} optimizes the following loss
\begin{equation}
\label{eq:mincut}
    \underbrace{-\frac{\text{Tr}(\mathbf{S}^T\mathbf{\tilde{A}}\mathbf{S})}{\text{Tr}(\mathbf{S}^T\mathbf{\tilde{D}}\mathbf{S})}}_{\mathcal{L}_c} 
    + 
    \underbrace{\left\| \frac{\mathbf{S}^T\mathbf{S}}{\|\mathbf{S}^T\mathbf{S}\|_F} - \frac{\mathbf{I}_K}{\sqrt{K}} \right\|_F}_{\mathcal{L}_o},
\end{equation}
where $\|\cdot\|_F$ indicates the Frobenius norm and $\tilde \D = \text{diag}(\tilde \A \mathbf{1})$.
The $\mathcal{L}_c$ term encourages strongly connected components to be clustered together, while $\mathcal{L}_o$ is a balancing term that promotes equally-sized clusters and helps to avoid degenerate solutions.
Similarly, DMoN~\citep{tsitsulin2020graph} optimizes a two-termed loss
\begin{equation}
\label{eq:dmon}
    \underbrace{- \frac{\text{Tr}(\Sb^T\A\Sb - \Sb^T\mathbf{d}^T\mathbf{d}\Sb)}{2|\mathcal{E}|}}_{\mathcal{L}_m} 
    + 
    \underbrace{\frac{\sqrt{K}}{N} \left\| \sum_i \Sb_i^T \right\|_F -1}_{\mathcal{L}_r},
\end{equation}
where $\mathbf{d}$ is the degree vector of $\A$, $\mathcal{L}_m$ pushes strongly connected components to the same cluster, and $\mathcal{L}_r$ is a regularization term that penalizes the degenerate solutions.

The losses in \eqref{eq:mincut} and \eqref{eq:dmon} are closely related to SC, from which the problem of smooth cluster assignments is inherited.
In addition, being trained with gradient descent, the GNNs do not exploit the closed-form solution of SC.
This motivates relying on GTV to perform clustering with a GNN.

\section{Total Variation Graph Neural Network}

In this section, we present the two core components of the Total Variation Graph Neural Network (TVGNN).
First, we introduce the unsupervised clustering loss inspired by the GTV relaxation.
Then, we present a novel MP layer to be used in conjunction with the proposed loss.
We conclude by showing two specific TVGNN architectures that can be used for vertex clustering and for graph classification. 

\subsection{The loss function}
\label{sec:loss}

In principle, the optimization of the objective in \eqref{eq:relaxed_asymcheeger} yields a balanced partition with sharp transitions in the assignment vectors of adjacent vertices belonging to different clusters.
However, the relaxed Asymmetric Cheeger Cut expression is ill-suited for the stochastic gradient descent used to train a GNN, due to potential numerical issues in the proximity of the degenerate solutions.
Specifically, all cluster assignments become similar when a degenerate solution is approached, bringing both the numerator and the denominator in \eqref{eq:relaxed_asymcheeger} close to zero and creating numerical instability in the gradients.
Adding small constants to avoid zero division can mitigate the issue only partially as it hinders the effect of the balancing term, which is what actually prevents degenerate solutions.
Finally, we desire a loss that could be easily combined with other losses (e.g., the cross-entropy) when the cluster assignments are used to implement graph pooling in a deep GNN for graph classification (see Sec.~\ref{sec:architectures}).
For this purpose, it is desirable to control the range of possible values it can assume and, therefore, a denominator taking arbitrary small values should be avoided.

To satisfy these requirements while retaining the desired properties of the objective function, we construct a loss by adding the GTV and balancing terms rather than taking their ratio. First, we define the GTV loss term as
\begin{equation}
    \label{eq:unscaled_tv_component}
    \mathcal{L}_\text{GTV}^* = {||\mathbf{S}||_\text{GTV}} = \sum_{k=1}^K \sum_{i=1}^N \sum_{j=i}^N a_{ij} |s_{ik} - s_{jk}|,
\end{equation}
Then, we define the asymmetrical norm term as
\begin{equation}
    \label{eq:unscaled_an_component}
    \mathcal{L}_\text{AN}^* = \sum_{k=1}^K ||\s_{:k} - \textrm{quant}_\rho (\s_{:k})||_{1, \rho}. 
\end{equation}
To control the range of values of the loss, the two terms are rescaled as follows:
\begin{align}
        \mathcal{L}_\text{GTV} &= \frac{\mathcal{L}_\text{GTV}^*}{2E} \in [0, 1],\\
        \mathcal{L}_\text{AN} &= \frac{\beta -  \mathcal{L}_\text{AN}^*}{\beta} \in [0, 1], \hspace{0.25cm}
\end{align}
where $E$ is the number of edges in the graph and 
\begin{equation}
    \beta = 
    \begin{cases}
        N\rho & \textrm{when}\; \rho = K-1, \\
        N\rho\min(1, K/(\rho+1)) & \textrm{otherwise.}
    \end{cases}
\end{equation}
%
The final loss reads:
\begin{equation}
    \label{eq:objective_loss}
    \mathcal{L} = \alpha_1\mathcal{L}_\text{GTV} + \alpha_2\mathcal{L}_\text{AN},
\end{equation}
where $\alpha_1, \alpha_2\in\mathbb{R}$ are hyperparameters that weigh the relative and total (in case there are other losses) contribution of the loss components.
For the experiments in Sec.~\ref{sec: experiments} we set $\rho = K - 1$, for which $\mathcal{L}_\text{AN}$ will be minimized for balanced clusters. Without prior knowledge, a balanced clustering is a reasonable bias. 
Nevertheless, the balancing term only poses a soft constraint that can be violated if needed, \textit{e.g.}, if the data show a clustering structure that is clearly uneven (see the additional results in \ref{appendix:additional_plots}).

The cluster assignments are computed by an $\texttt{MLP}$ fed with the vertex representations produced by a stack of $L$ MP layers
\begin{equation}
    \label{eq:mlp}
    \Sb = \texttt{Softmax}(\texttt{MLP}(\X^{(L)}; \mathbf{\Theta}_\text{MLP}))
\end{equation}
As discussed in Sec.~\ref{sec:gnn_background}, common MP layers minimize the LQV of vertex features and the \texttt{MLP}, which is a smooth function, will naturally preserve the Laplacian smoothing effect when computing $\mathbf{S}$. 
While this was suitable for an SC objective, the optimization problem expressed by $\mathcal{L}_\text{GTV}$ entails minimizing the $\ell_1$ norm $\|\s_i - \s_j\|_1$ of each pair of adjacent vertices $i$ and $j$. 
Thus, the MP layers should ideally minimize the discrepancy in the features of adjacent vertices in a $\ell_1$ norm sense.
To satisfy this requirement, we design a new MP layer starting from the definition of the gradient of an approximated GTV function.

\subsection{The GTVConv layer}
\label{sec:gtvconv_layer}

Our goal is to design an MP layer that minimizes the GTV of the vertex features.
While the GTV can be expressed by means of an incidence matrix~\citep{JMLR:v17:15-147} or a (nonlinear) $\ell_1$-Laplacian operator~\citep{bai2018graph, zhou2005regularization}, these formulations are difficult to integrate within a standard MP layer that operates on a connectivity matrix. Furthermore, GTV is non-differentiable for $x_i=x_j$.

To address these issues, we define the GTV Laplacian as 
\begin{equation}
    \label{eq:l1_laplacian_univariate}
    \mathbf{L}_\mathbf{\Gamma} = \mathbf{D}_{\mathbf{\Gamma}} - \mathbf{\Gamma}, \;\; \text{with} \;\; \mathbf{D}_{\mathbf{\Gamma}}=\text{diag}(\mathbf{\Gamma}\boldsymbol{1}),
\end{equation}
where $\mathbf{\Gamma}$ is a connectivity matrix matching the sparsity pattern of the adjacency matrix, with elements
\begin{equation}
    \label{eq:gamma_element_univariate}
    [\mathbf{\Gamma}]_{ij} = \gamma_{ij} = \frac{a_{ij}}{\max\{|x_i-x_j|, \epsilon\}},
\end{equation}
where $\epsilon$ is a small constant, whose purpose will be clarified soon. 
To minimize the GTV, we compute the partial derivative of $||\x||_\text{GTV}$ with respect to vertex $k$ for some $\x\in\mathbb{R}^{N}$. According to the definition of $\mathbf{\Gamma}$ and assuming that the graph is undirected, the partial derivative reads
\begin{equation}
    \label{eq:gtv_univariate_subdifferential}
    \frac{\partial}{\partial x_k}(||\x||_\text{GTV})_\epsilon = 2\sum_{j=1}^N \gamma_{kj}\cdot (x_k - x_j),
\end{equation}
and the full GTV gradient is
\begin{equation}
    \label{eq:gtv_gradient_univariate}
    \nabla (||\x||_\text{GTV})_\epsilon = 2\mathbf{D_\Gamma}\x - 2\mathbf{\Gamma}\x = 2(\mathbf{D_\Gamma} - \mathbf{\Gamma})\x = 2\mathbf{L_\Gamma}\x. 
\end{equation}

By looking at \eqref{eq:gtv_univariate_subdifferential}, we now see that the constant $\epsilon$ ensures numerical stability by avoiding the discontinuity in the derivative of $||\x||_\text{GTV}$.
\begin{figure}[!ht]
    \centering
    \includegraphics[width=\columnwidth]{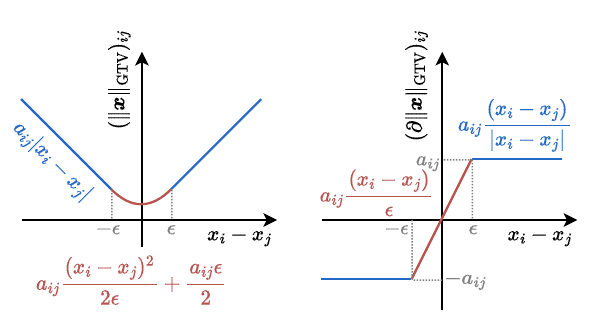}
    \caption{\small The GTV function (in blue) is modified (parts in red) near zero to avoid the discontinuity in the derivative.}
    \label{fig:functions}
\end{figure}
Indeed, using $\max\{|x_i-x_j|, \epsilon\}$ rather than $|x_i-x_j|$ as the denominator in \eqref{eq:gamma_element_univariate} corresponds to modifying the GTV function and its derivative in the proximity of $|x_i-x_j|=0$, as shown in Fig.~\ref{fig:functions}.
A detailed discussion and the derivations are deferred to Appendix~\ref{appendix:GTVConv}.

Based on \eqref{eq:gtv_gradient_univariate}, the $\epsilon$-approximation of $||\x||_\text{GTV}$ is minimized by taking the following gradient descent update 
\begin{equation}
    \label{eq:gtv_update_univariate}
    \x^{(t+1)} =  \left(\mathbf{I} - 2\delta\mathbf{L_\Gamma}^{(t)} \right)\x^{(t)}
\end{equation}
where $\delta$ is the step size.
The update in \eqref{eq:gtv_update_univariate} closely resembles the update in \eqref{eq:gcn_update} and it can be implemented by a GCN layer operating on the connectivity matrix $\mathbf{I} - 2\delta \mathbf{L}_\mathbf{\Gamma}^{(t)}$. We note that the superscript $(t)$ in $\mathbf{L}_\mathbf{\Gamma}^{(t)}$ indicates the dependency on the features $\x^{(t)}$ in the denominator of \eqref{eq:gamma_element_univariate}.
Without loss of generality, from now on we replace index $t$, which indicates the $t$-th step in the gradient descent update, with layer index $l$, meaning that each MP layer in the GNN performs a gradient descent update to minimize the GTV.  

The aggregation procedure in \eqref{eq:gtv_update_univariate} is only valid for univariate vertex features $\x \in \R^{N}$. For a graph with multi-dimensional features $\mathbf{X}\in\mathbb{R}^{N\times F}$ the partial derivative is
\begin{equation}
\label{eq:gamma_multi}
    \frac{\partial}{\partial x_{kf}}(||\mathbf{X}||_\text{GTV})_\epsilon =\sum_{j=1}^N \gamma_{kjf}\cdot(x_{kf}-x_{jf}),
\end{equation}
where
\begin{equation}
    \gamma_{kjf} = \frac{a_{kj}}{\max\{|x_{kf}-x_{jf}|,\epsilon\}}, \; f=1,\dots,F.
\end{equation}
The dependence on $f$ implies that we need a distinct $\mathbf{\Gamma}_f$ for each feature and that the update in \eqref{eq:gtv_update_univariate} must be done feature-wise. 
Notably, each time an MP layer maps the vertex features in a new $F'$-dimensional space, $F'$ different connectivity matrices are required.
Clearly, this introduces two major drawbacks. 
First, building, storing, and applying many $\mathbf{\Gamma}_f$ matrices is computationally expensive, especially for large graphs.
Second, since each feature is updated independently, the minimization of the approximated $\| \X \|_\text{GTV}$ can exhibit erratic behaviors and fail to converge to the optimal solution. 
See \ref{appendix:single_vs_multi} for a detailed discussion and an example.

To address these issues, we modify the gradient descent step by defining a single operator that computes the $\ell_1$ distance over the full feature vectors:
\begin{equation}
    \label{eq:gamma_single}
    \hat{\gamma}_{ij} = \frac{a_{ij}}{\max\{||\x_i - \x_j ||_1, \epsilon\}}.
\end{equation}
By letting
\begin{equation}
    \label{eq:l1_laplacian_approx}
    \hat{\mathbf{L}}_\mathbf{\Gamma} = \mathbf{\hat{D}_\Gamma} - \mathbf{\hat{\Gamma}}, \;\; \text{with} \;\; [\mathbf{\hat{\Gamma}}]_{ij} = \hat{\gamma}_{ij} \;\; \text{and}\;\;
    \hat{\mathbf{D}}_\mathbf{\Gamma} = \text{diag}(\mathbf{\hat{\Gamma}}\boldsymbol{1}),
\end{equation}
the vertex features update at step/layer $l+1$ becomes
\begin{equation}
    \label{eq:update_gtv_single}
    \mathbf{X}^{(l+1)} = \left(\mathbf{I} - 2\delta{\hat{\Lb}_\mathbf{\Gamma}}^{(l)}\right) \mathbf{X}^{(l)}.
\end{equation}

By referring to the notation in \eqref{eq: gcn_layer}, the proposed GTVConv layer reads
\begin{equation}
    \mathbf{X}^{(l+1)} = \sigma\left[\left(\mathbf{I} - 2\delta{\hat{\Lb}_\mathbf{\Gamma}^{(l)}}\right) \mathbf{X}^{(l)} \mathbf{\Theta}_\text{MP}\right]
    \label{eq:gtvconv}
\end{equation}

\paragraph{Remarks}
There is a clear analogy between the LQV in \eqref{eq:lap_smooth_comb}, minimized by common MP layers, and the GTV defined in terms of $\hat{\Lb}_\mathbf{\Gamma}$, minimized by GTVConv.
While the MP layers based on Laplacian smoothing perform low-pass filtering~\citep{bo2021beyond}, GTVConv is closely related to graph trend filtering~\citep{JMLR:v17:15-147, pmlr-v139-liu21k}, which implements a total variation smoother based on the $\ell_1$ Laplacian.
While linear smoothers cannot handle heterogeneous smoothness, a total variation smoother encompasses both globally smooth functions, said to have homogeneous smoothness, and functions with different levels of smoothness at different graph locations~\citep{sadhanala2016total}. From the graph signal processing perspective, it means applying low- and high-pass filtering on the graph signal at the same time~\citep{fu2022p}.
In our case, when driven by the $\mathcal{L}_\text{GTV}$ term in \eqref{eq:objective_loss}, GTVConv applies low-pass filtering to central vertices in the cluster and high-pass filtering at the clusters' boundary, enabling sharp cluster transitions. 

Similarly to attention-based GNNs~\citep{velivckovic2017graph}, GTVConv can learn edge weights in a data-driven fashion: by looking at \eqref{eq:gamma_single}, we notice that $\hat{\gamma}_{ij}$ depend on the features $\x_i, \x_j$.
Therefore, the output features at layer $l$ will influence the edge weights at layer $l+1$.

A variant to the proposed GTVConv layer is obtained from a GTV weighted by the vertex degrees, which gives an expression related to the LQV defined in terms of the symmetric normalized Laplacian (details in \ref{appendix:GTVConv_weight}).

\subsection{TVGNN architectures for clustering and classification}
\label{sec:architectures}

In the following, we describe the GNN architectures we used in two downstream tasks: unsupervised vertex clustering and supervised graph classification. 

\begin{figure}[!ht]
    \centering
    \begin{subfigure}{0.4\columnwidth}
        \centering
        \includegraphics[width=\textwidth]{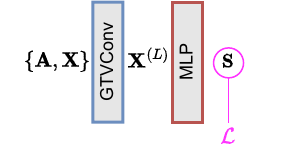}
        \caption{Clustering}
    \end{subfigure}
    
    \vspace{.5cm}
    
    \begin{subfigure}{\columnwidth}
        \centering
        \includegraphics[width=\textwidth]{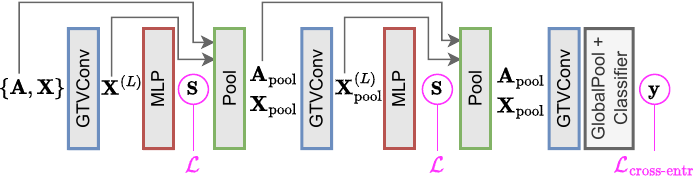}
        \caption{Classification}
    \end{subfigure}
    \caption{\small Schematic depiction of the architectures used for vertex clustering and graph classification.}
    \label{fig:architectures}
\end{figure}

\paragraph{Vertex clustering}
The GNN used for clustering is depicted in Fig.~\ref{fig:architectures}a.
The architecture is rather simple: a stack of $L$ GTVConv layers ($L \geq 1$) generates the feature vectors $\X^{(L)}$ that are used by an MLP to compute the cluster assignments $\Sb$. Since clustering is an unsupervised task, the GNN is trained using only the loss function $\mathcal{L}$ defined in \eqref{eq:objective_loss}.

\paragraph{Graph classification}
Graph classification is a graph-level task, where a class label $y_i$ is assigned to the $i$-th graph $\{ \A_i, \X_i \}$.
GNN architectures for graph classification often alternate MP layers with graph pooling layers, which gradually distill the global label information from the vertex representations~\citep{du2021multi}.

The key challenge in graph pooling is to generate a coarsened graph that summarizes well the properties of the original one~\citep{bianchi2023expressive}.
Similarly to previous work~\citep{ying2018hierarchical, bianchi2020spectral}, the cluster assignment matrix $\Sb$ computed in \eqref{eq:mlp} is used to coarsen the adjacency matrix and to compute pooled vertex features as
\begin{equation}
    \mathbf{A}^\text{pool} = \mathbf{S}^T \mathbf{A} \mathbf{S} \in\mathbb{R}^{K\times K}; \, \mathbf{X}^\text{pool}=\mathbf{S}^T\mathbf{X} \in\mathbb{R}^{K\times F}.
\end{equation}
According to a recently proposed taxonomy by \cite{grattarola2022understanding}, ours is a trainable, dense, fixed, hierarchical pooling method.

Graph pooling can be applied multiple times, to obtain smaller and smaller coarsened graphs.
The features on the final coarsened graph are globally pooled and passed to a classifier that predicts the class label.
A different instance of the loss in \eqref{eq:objective_loss} is used to optimize the cluster assignments at each pooling layer.
The total loss is given by combining the clustering losses, which in this case act as regularizers, and a supervised cross-entropy loss $\mathcal{L}_\text{cross-entr}$ between true and predicted class labels.
Fig.~\ref{fig:architectures}b shows a schematic depiction.

By construction, each entry $a^\text{pool}_{ij}$ of $\mathbf{A}^\text{pool}$ is the volume of edges across clusters $i$ and $j$.
Since the minimization of $\mathcal{L}_\text{GTV}$ pushes connected components to the same cluster, $\mathbf{A}^\text{pool}$ gradually turns into a diagonally dominant matrix.
This would limit the contribution of any MP layer operating on $\mathbf{A}^\text{pool}$, since the vertices will share information mostly with themselves.  
However, in the Laplacian \eqref{eq:l1_laplacian_approx} used by the GTVConv layer, the diagonal of the coarsened adjacency is removed, which avoids this potential issue.

\section{Experiments}
\label{sec: experiments}

We evaluate the proposed TVGNN model on unsupervised vertex clustering and supervised graph classification tasks.
The code to implement TVGNN is publicly available\footnote{
\url{https://github.com/FilippoMB/Total-variation-graph-neural-networks}
}.

\subsection{Unsupervised vertex clustering}

\begin{table*}[!ht]
    \myfootnotesize
    \centering
    \caption{\small NMI and ACC results for vertex clustering. The highest averages are in bold and the second highest are underlined.}
    \begin{tabular}{l c c c c c c c c | c c}
        \toprule
        \textbf{Method} & \multicolumn{2}{c}{\textbf{Cora}} & \multicolumn{2}{c}{\textbf{Citeseer}} & \multicolumn{2}{c}{\textbf{Pubmed}} & \multicolumn{2}{c}{\textbf{DBLP}} & \multicolumn{2}{c}{\textbf{Tot. Average}}\\
        \midrule
        & NMI & ACC & NMI & ACC & NMI & ACC & NMI & ACC & NMI & ACC\\
        SC 
             & \meanstd{0.029}{0.017} & \meanstd{29.8}{0.7} 
             & \meanstd{0.014}{0.003} & \meanstd{21.7}{0.3} 
             & \meanstd{0.183}{0.000} & \meanstd{59.0}{0.0} 
             & \meanstd{0.023}{0.005} & \meanstd{45.8}{0.2} 
             & 0.062 & 39.1                                 
             \\
        DeepWalk 
            & \meanstd{0.064}{0.024} & \meanstd{23.0}{2.1} 
            & \meanstd{0.005}{0.001} & \meanstd{19.4}{0.3} 
            & \meanstd{0.001}{0.000} & \meanstd{36.1}{0.1} 
            & \meanstd{0.001}{0.000} & \meanstd{26.7}{0.1} 
            & 0.018 & 26.3
            \\
        node2vec
            & \meanstd{0.060}{0.030} & \meanstd{23.0}{2.5} 
            & \meanstd{0.004}{0.001} & \meanstd{19.5}{0.3} 
            & \meanstd{0.001}{0.000} & \meanstd{36.2}{0.1} 
            & \meanstd{0.001}{0.000} & \meanstd{27.3}{0.1} 
            & 0.017 & 26.5
            \\
        NetMF
            & \meanstd{0.251}{0.000} & \meanstd{38.9}{0.0} 
            & \meanstd{0.127}{0.000} & \meanstd{27.7}{0.0} 
            & \meanstd{0.059}{0.000} & \meanstd{44.8}{0.0} 
            & \meanstd{0.037}{0.000} & \meanstd{45.6}{0.0} 
            & 0.119 & 39.3
            \\
        \midrule
        TADW
            & \meanstd{0.012}{0.000} & \meanstd{19.3}{0.0} 
            & \meanstd{0.002}{0.000} & \meanstd{18.8}{0.0} 
            & \meanstd{0.031}{0.000} & \meanstd{42.8}{0.0} 
            & \meanstd{0.012}{0.000} & \meanstd{29.8}{0.0} 
            & 0.014 & 27.7
            \\
        BANE
            & \meanstd{0.291}{0.000} & \meanstd{49.5}{0.0} 
            & \meanstd{0.260}{0.000} & \meanstd{49.4}{0.0} 
            & \meanstd{0.121}{0.000} & \meanstd{50.9}{0.0} 
            & \meanstd{0.177}{0.000} & \meanstd{50.6}{0.0} 
            & 0.212 & 50.1 
            \\
        TENE
            & \meanstd{0.115}{0.000} & \meanstd{25.5}{0.0} 
            & \meanstd{0.005}{0.000} & \meanstd{20.8}{0.0} 
            & \meanstd{0.002}{0.000} & \meanstd{39.9}{0.0} 
            & \meanstd{0.003}{0.000} & \meanstd{40.2}{0.0} 
            & 0.125 & 31.6 
            \\
        GAE
            & \meanstd{0.328}{0.051} & \meanstd{46.5}{6.2} 
            & \meanstd{0.163}{0.029} & \meanstd{38.1}{3.8} 
            & \meanstd{\underline{0.235}}{0.044} & \meanstd{58.9}{7.2} 
            & \meanstd{0.111}{0.029} & \meanstd{41.6}{3.5} 
            & 0.209 & 46.3
            \\
        VGAE
            & \meanstd{\underline{0.437}}{0.029} & \meanstd{\underline{57.2}}{5.4} 
            & \meanstd{0.156}{0.034} & \meanstd{36.0}{3.9} 
            & \meanstd{\textbf{0.245}}{0.043} & \meanstd{\textbf{61.1}}{6.1} 
            & \meanstd{0.213}{0.021} & \meanstd{50.7}{4.7} 
            & 0.263 & 51.3
            \\
        
        \midrule
        DiffPool 
            & \meanstd{0.307}{0.006} & \meanstd{47.3}{1.0} 
            & \meanstd{0.180}{0.008} & \meanstd{33.6}{0.8} 
            & \meanstd{0.084}{0.002} & \meanstd{41.8}{0.3} 
            & \meanstd{0.045}{0.044} & \meanstd{37.1}{4.3} 
            & 0.154 & 40.0
            \\
        MinCutPool
            & \meanstd{0.406}{0.029} & \meanstd{53.4}{4.1} 
            & \meanstd{\underline{0.295}}{0.029} & \meanstd{\underline{49.8}}{4.9} 
            & \meanstd{0.209}{0.015} & \meanstd{57.3}{3.5} 
            & \meanstd{0.297}{0.025} & \meanstd{53.8}{3.4} 
            & \underline{0.302} & \underline{53.6}
            \\
        DMoN
            & \meanstd{0.357}{0.043} & \meanstd{48.8}{6.4} 
            & \meanstd{0.196}{0.030} & \meanstd{36.4}{4.3} 
            & \meanstd{0.193}{0.049} & \meanstd{55.9}{4.2} 
            & \meanstd{\underline{0.335}}{0.027} & \meanstd{\underline{59.0}}{4.0} 
            & 0.270 & 50.0 
            \\
                    TVGNN
            & \meanstd{\textbf{0.488}}{0.016} & \meanstd{\textbf{63.2}}{1.8} 
            & \meanstd{\textbf{0.361}}{0.018} & \meanstd{\textbf{58.6}}{3.0} 
            & \meanstd{0.216}{0.027} & \meanstd{\underline{60.0}}{2.0} 
            & \meanstd{\textbf{0.342}}{0.011} & \meanstd{\textbf{60.8}}{1.5} 
            & \textbf{0.352} & \textbf{60.7}
            \\
        \midrule
        ABL1 
            & \meanstd{0.376}{0.018} & \meanstd{48.0}{2.0} 
            & \meanstd{0.210}{0.019} & \meanstd{41.1}{3.8} 
            & \meanstd{0.222}{0.014} & \meanstd{57.7}{3.8} 
            & \meanstd{0.260}{0.035} & \meanstd{52.2}{3.3} 
            & 0.267 & 49.8
            \\
        ABL2
            & \meanstd{0.388}{0.024} & \meanstd{50.7}{4.9} 
            & \meanstd{0.285}{0.038} & \meanstd{49.3}{5.2} 
            & \meanstd{0.191}{0.035} & \meanstd{58.5}{3.3} 
            & \meanstd{0.262}{0.044} & \meanstd{50.4}{4.4} 
            & 0.282 & 52.2
            \\
        \bottomrule
    \end{tabular}
    \label{tab: results_clustering}
\end{table*}

\begin{figure*}[!ht]
    \centering
    \begin{subfigure}{0.23\textwidth}
        \centering
        \includegraphics[width=\textwidth]{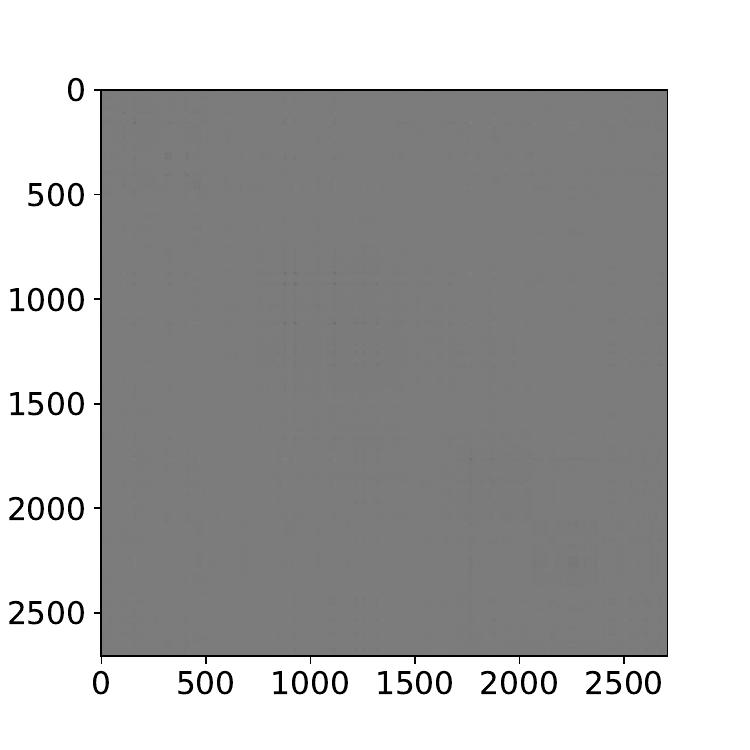}
        \caption{DiffPool}
    \end{subfigure}
    \begin{subfigure}{0.23\textwidth}
        \centering
        \includegraphics[width=\textwidth]{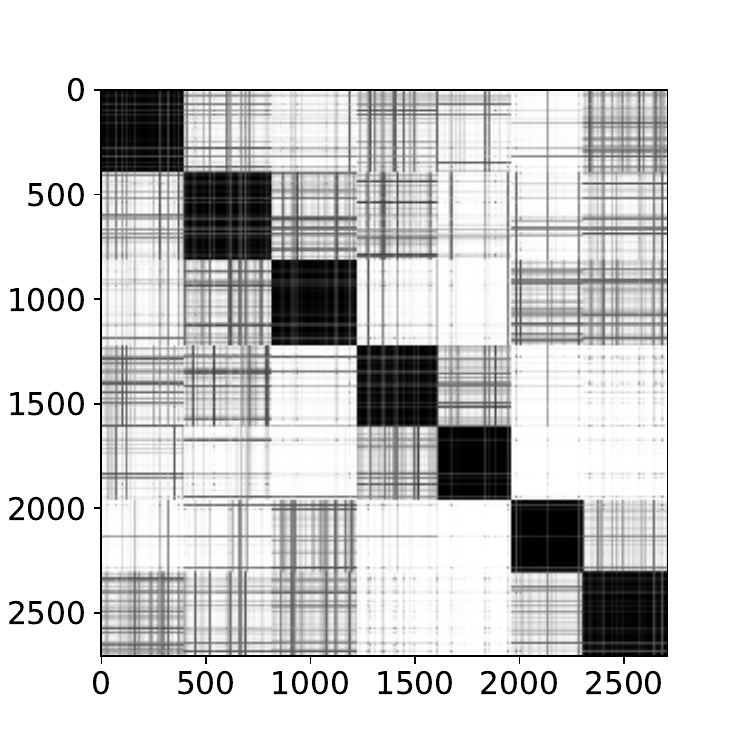}
        \caption{MinCutPool}
    \end{subfigure}
    \begin{subfigure}{0.23\textwidth}
        \centering
        \includegraphics[width=\textwidth]{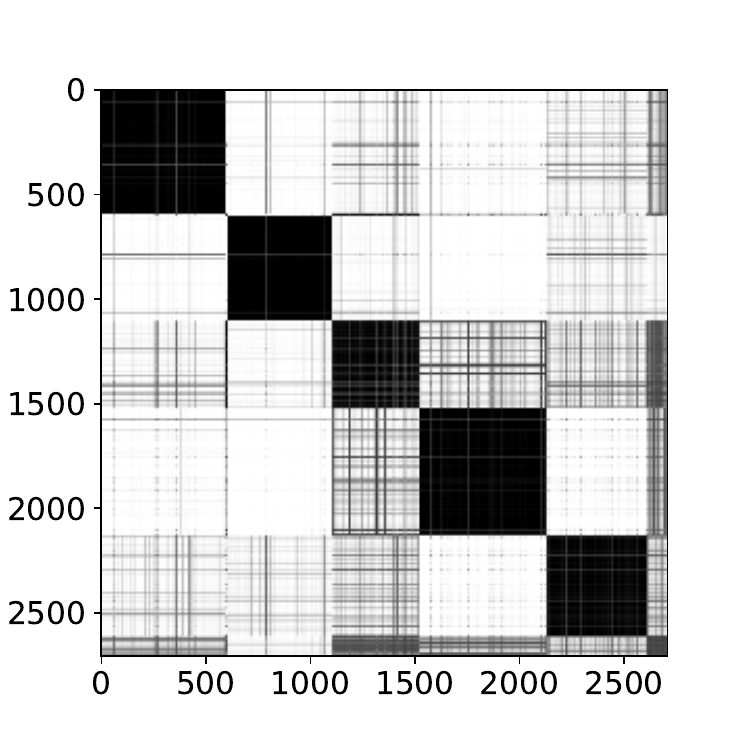}
        \caption{DMoN}
    \end{subfigure}
     \begin{subfigure}{0.23\textwidth}
        \centering
        \includegraphics[width=\textwidth]{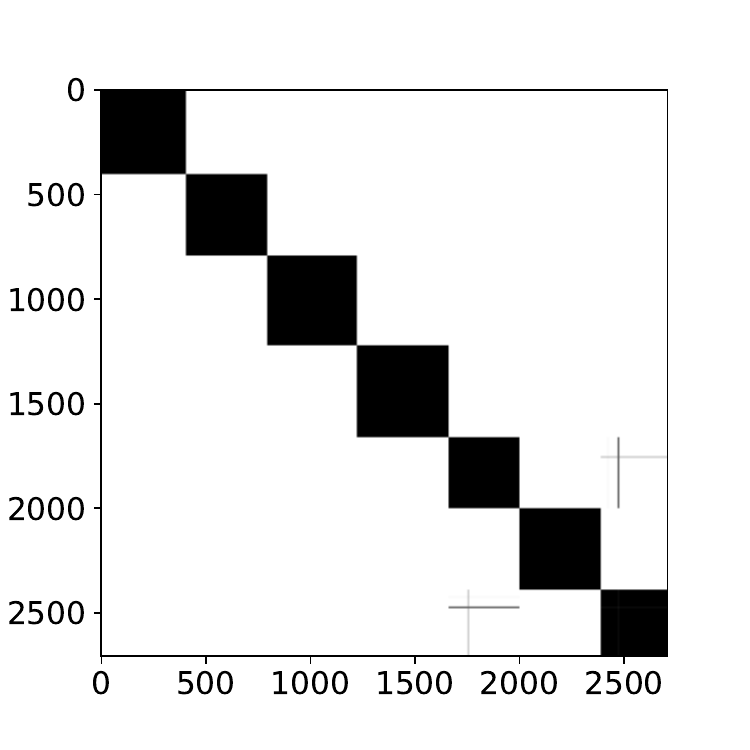}
        \caption{TVGNN}
    \end{subfigure}
    \caption{\small Visualization of the logarithm of $\mathbf{S}\mathbf{S}^T$ for Cora.}
    \label{fig: sharpness_plot_cora}
\end{figure*}

\begin{figure*}[!ht]
    \centering
    \small
    \begin{subfigure}{0.23\textwidth}
        \centering
        \includegraphics[width=\textwidth]{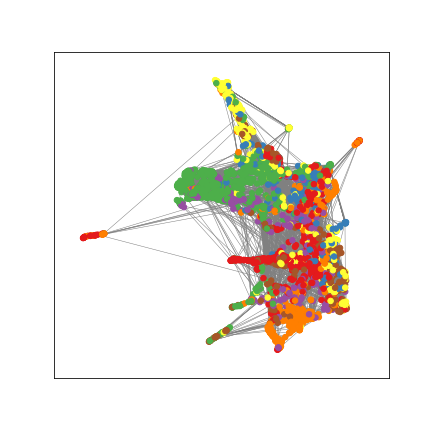}
        \vspace*{-0.8cm}
        \caption{DiffPool}
    \end{subfigure}
    \begin{subfigure}{0.23\textwidth}
        \centering
        \includegraphics[width=\textwidth]{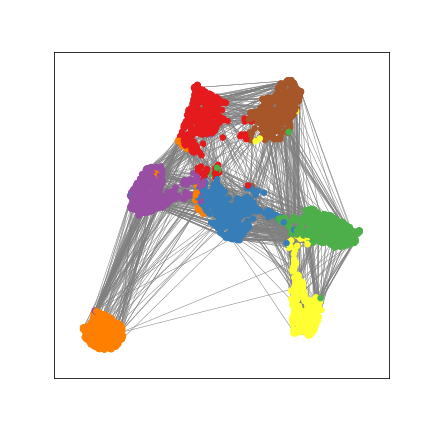}
        \vspace*{-0.8cm}
        \caption{MinCutPool}
    \end{subfigure}
    \begin{subfigure}{0.23\textwidth}
        \centering
        \includegraphics[width=\textwidth]{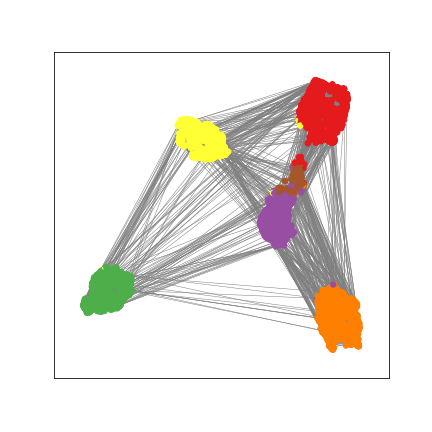}
        \vspace*{-0.8cm}
        \caption{DMoN}
    \end{subfigure}
    \begin{subfigure}{0.23\textwidth}
        \centering
        \includegraphics[width=\textwidth]{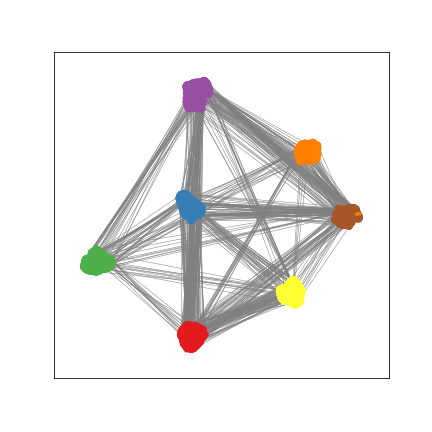}
        \vspace*{-0.8cm}
        \caption{TVGNN}
    \end{subfigure}
    \begin{subfigure}{0.23\textwidth}
        \centering
        \includegraphics[width=\textwidth]{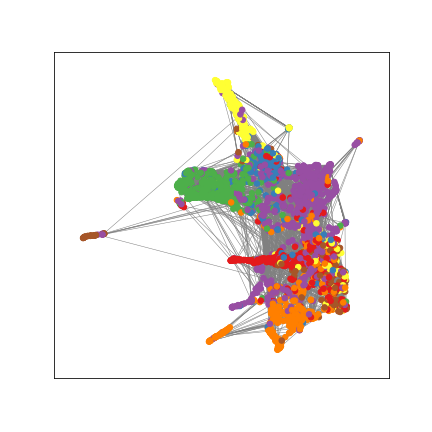}
        \vspace*{-0.8cm}
        \caption{DiffPool}
    \end{subfigure}
    \begin{subfigure}{0.23\textwidth}
        \centering
        \includegraphics[width=\textwidth]{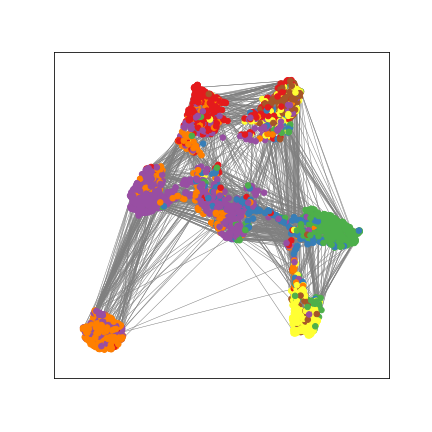}
        \vspace*{-0.8cm}
        \caption{MinCutPool}
    \end{subfigure}
    \begin{subfigure}{0.23\textwidth}
        \centering
        \includegraphics[width=\textwidth]{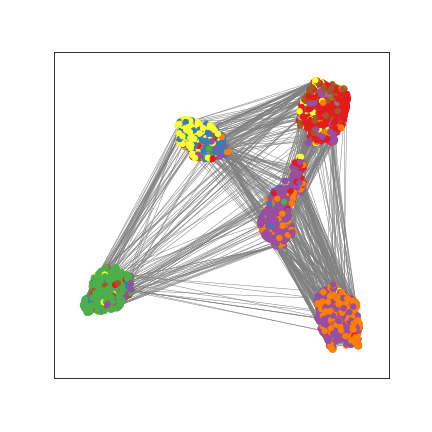}
        \vspace*{-0.8cm}
        \caption{DMoN}
    \end{subfigure}
    \begin{subfigure}{0.23\textwidth}
        \centering
        \includegraphics[width=\textwidth]{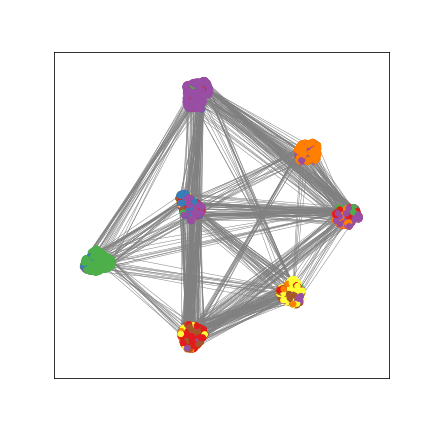}
        \vspace*{-0.8cm}
        \caption{TVGNN}
    \end{subfigure}
    \caption{\small 2D UMAP transform of $\X^{(L)}$ for Cora with the edges from the original adjacency matrix. In (a)-(d) vertex colors correspond to cluster assignments $\hat \y$ from Kuhn-Munkres and in (e)-(h) colors correspond to the true labels.}
    \label{fig: umap_transform}
\end{figure*}

In this experiment, we evaluate the capability of TVGNN to create cluster assignments that are sharp and match well the true vertex class.
The performance of TVGNN is compared against three classes of methods. 
The first, are algorithms that generate vertex embeddings based only on the adjacency matrix. 
The vertex embeddings are then clustered with $k$-means.
Representatives of this category are Spectral Clustering (SC), DeepWalk~\citep{perozzi2014deepwalk}, Node2vec~\citep{grover2016node2vec}, and NetMF~\citep{qiu2018network}.
The second class of methods generates vertex embeddings by accounting both for the adjacency matrix and for the vertex features. 
Afterward, the learned embeddings are clustered with $k$-means.
The chosen representatives for this category are the Graph AutoEncoder (GAE) and Variational Graph AutoEncoder (VGAE)~\citep{kipf2016variational}, TADW~\citep{10.5555/2832415.2832542}, BANE~\citep{yang2018binarized}, and TENE~\citep{yang2018enhanced}.
Finally, the last class of methods consists of end-to-end GNN models that directly generate soft cluster assignments $\Sb$ by accounting both for the graph connectivity and the vertex features. 
In this case, $k$-means is not required and the discrete cluster assignments are simply obtained as $\cb = \texttt{argmax}(\Sb)$.
DiffPool~\citep{ying2018hierarchical}, DMoN~\citep{tsitsulin2020graph}, MinCutPool~\citep{bianchi2020spectral}, and the proposed TVGNN belong to this class. 
The GNNs equipped with DiffPool, DMoN, and MinCutPool have the same general architecture depicted in Fig~\ref{fig:architectures}a: a stack of MP layers as in \eqref{eq: gcn_layer} followed by a layer that computes $\Sb$.
The GNNs are trained only by minimizing unsupervised losses, such as those in Eq.~\ref{eq:mincut} (MinCutPool), Eq.~\ref{eq:dmon} (DMoN), and Eq.~\ref{eq:objective_loss} (TVGNN). 
The hyperparameters of each model are in \ref{appendix:hyperparams}.


\begin{figure}[!ht]
    \centering
    \begin{subfigure}{0.23\textwidth}
        \centering
        \includegraphics[width=\textwidth]{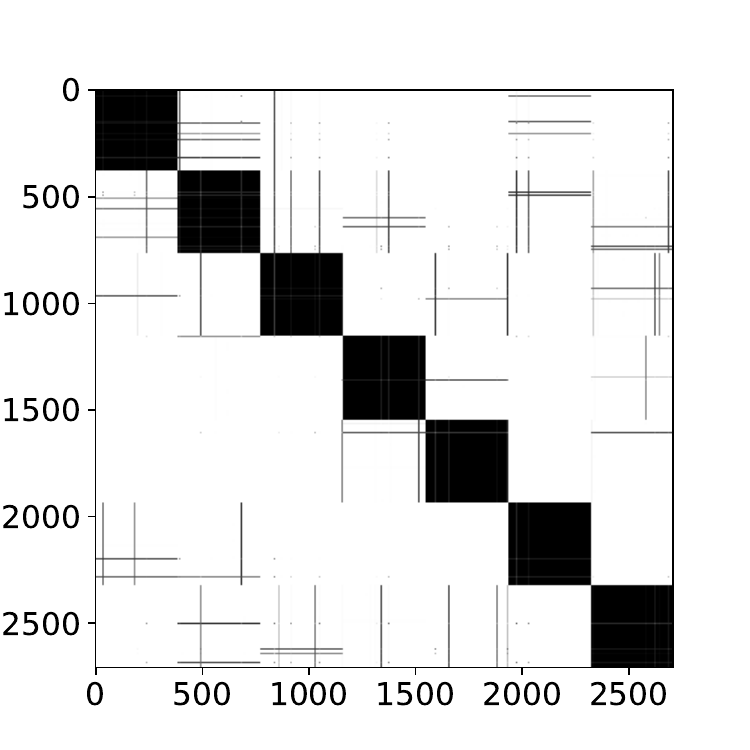}
        \caption{GTVConv + MinCut loss}
    \end{subfigure}
    \begin{subfigure}{0.23\textwidth}
        \centering
        \includegraphics[width=\textwidth]{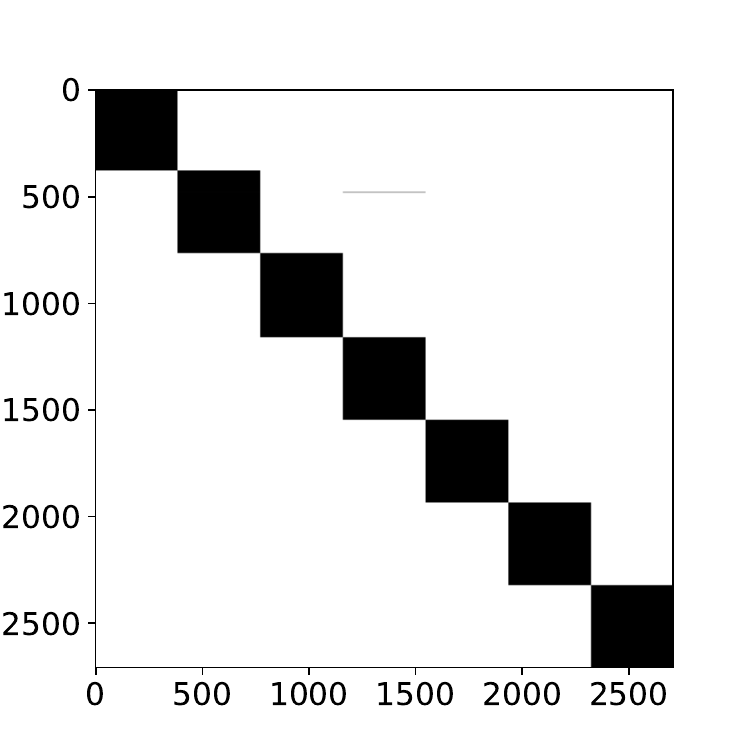}
        \caption{GCN + TVGNN loss} 
    \end{subfigure}
    \caption{\small Visualization of the logarithm of $\mathbf{S}\mathbf{S}^T$ for configurations used in the ablation study.}
    \label{fig: sharpness_plot_ablation_cora}
\end{figure}

\begin{figure}[!ht]
    \centering
    \small
    \begin{subfigure}{0.23\textwidth}
        \centering
        \includegraphics[width=\textwidth]{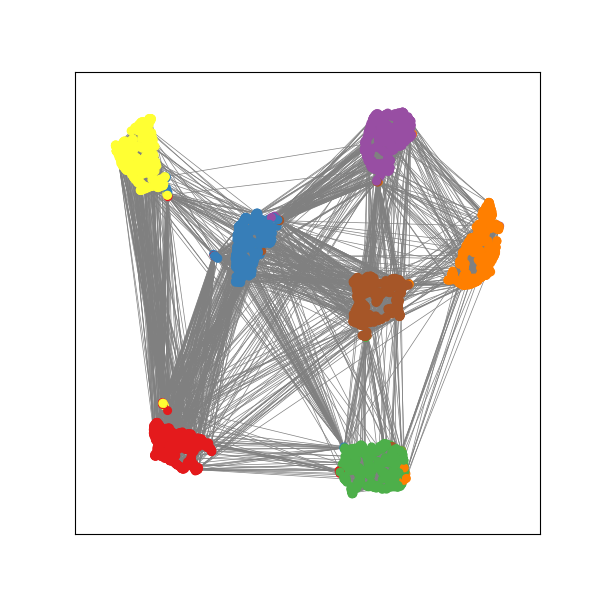}
        \vspace*{-0.8cm}
        \caption{GTVConv + MinCut loss}
    \end{subfigure}
    \begin{subfigure}{0.23\textwidth}
        \centering
        \includegraphics[width=\textwidth]{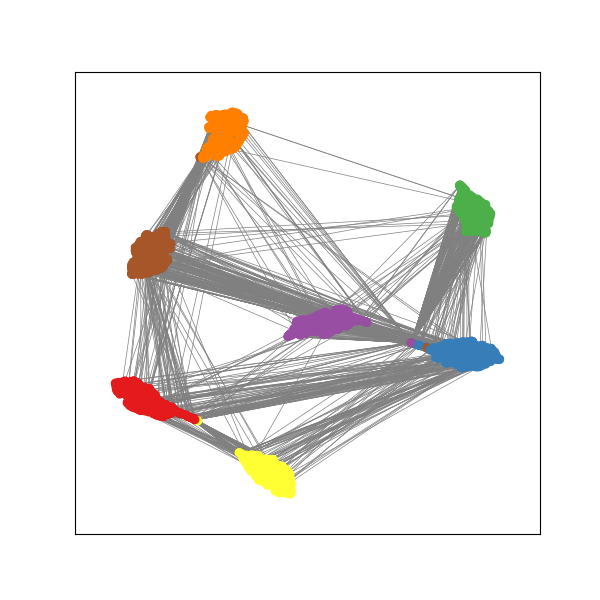}
        \vspace*{-0.8cm}
        \caption{GCN + TVGNN loss}
    \end{subfigure}
    \\
    \begin{subfigure}{0.23\textwidth}
        \centering
        \includegraphics[width=\textwidth]{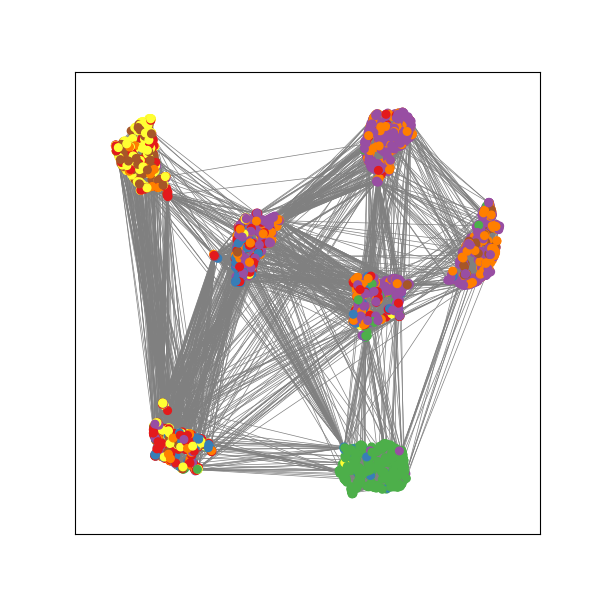}
        \vspace*{-0.8cm}
        \caption{GTVConv + MinCut loss}
    \end{subfigure}
    \begin{subfigure}{0.23\textwidth}
        \centering
        \includegraphics[width=\textwidth]{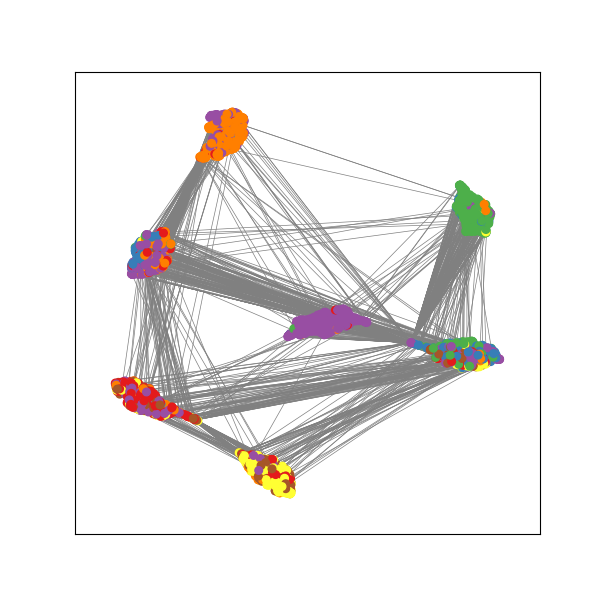}
        \vspace*{-0.8cm}
        \caption{GCN + TVGNN loss}
    \end{subfigure}
    \caption{\small Equivalent plots to \cref{fig: umap_transform} for the ablation study configurations. Colors correspond to the cluster assignments in the top row and the true labels in the bottom row.}
    \label{fig: umap_transform_ablation}
\end{figure}


\begin{table*}[!ht]
    \small
    \centering
    \caption{\small Graph classification accuracy. The highest mean accuracy for each dataset is in bold, and the second highest is underlined. We report the $p$-value of the difference between the two highest means ($^{*}$ and $^{**}$ denote significance at 95\% and 99\% confidence levels).}
    \begin{tabular}{l c c c c c c c}
        \toprule
         \textbf{Dataset} & \textbf{Top-$K$} & \textbf{SAGPool} & \textbf{DiffPool} & \textbf{MinCutPool} & \textbf{DMoN} & \textbf{TVGNN} & \textbf{\textit{p}-value} \\
         \midrule
         Bench-easy 
            & \meanstd{53.8}{31.8} & \meanstd{53.8}{31.8} & \meanstd{\underline{99.0}}{0.3} 
            & \meanstd{\underline{99.0}}{0.3} & \meanstd{98.8}{0.5} & \meanstd{\textbf{99.6}}{0.6} & .011$^{*}$ \\
         Bench-hard 
            & \meanstd{30.5}{0.7} & \meanstd{29.5}{0.0} & \meanstd{\underline{72.8}}{0.2}
            & \meanstd{70.9}{1.7} & \meanstd{71.8}{1.9} & \meanstd{\textbf{75.3}}{0.8} & <.001$^{**}$ \\
         MUTAG 
            & \meanstd{77.5}{8.4} & \meanstd{76.8}{9.7} & \meanstd{86.4}{7.6}
            & \meanstd{85.2}{7.2} & \meanstd{\underline{86.7}}{7.0} & \meanstd{\textbf{88.4}}{7.5} & .606 \\
         Mutagenicity 
            & \meanstd{68.4}{8.4} & \meanstd{68.2}{7.8} & \meanstd{\underline{78.5}}{1.5}
            & \meanstd{78.4}{1.4} & \meanstd{77.1}{1.3} & \meanstd{\textbf{80.0}}{1.3} & .028$^{*}$ \\
         NCI1 
            & \meanstd{54.0}{4.1} & \meanstd{59.2}{7.7} & \meanstd{74.1}{1.8}
            & \meanstd{\underline{75.2}}{1.8} & \meanstd{74.3}{1.3} & \meanstd{\textbf{77.3}}{1.8} & .018$^{*}$ \\
         Proteins 
            & \meanstd{69.6}{2.7} & \meanstd{70.4}{2.5} & \meanstd{74.6}{4.2}
            & \meanstd{\underline{75.7}}{3.0} & \meanstd{75.2}{3.3} & \meanstd{\textbf{77.1}}{2.9} & .302 \\
         D\&D 
            & \meanstd{62.0}{5.6} & \meanstd{64.2}{7.0} & \meanstd{77.7}{3.0}
            & \meanstd{\underline{78.2}}{3.4} & \meanstd{78.0}{3.3} & \meanstd{\textbf{79.5}}{2.2} & .323 \\
         COLLAB 
            & \meanstd{73.4}{6.9} & \meanstd{75.6}{2.5} & \meanstd{78.4}{1.6}
            & \meanstd{79.3}{1.1} & \meanstd{\underline{79.5}}{0.7} & \meanstd{\textbf{79.8}}{1.1} & .476 \\
        REDDIT-BINARY
            & \meanstd{54.0}{10.0} & \meanstd{50.0}{0.1} & \meanstd{80.9}{2.7} & \meanstd{82.3}{3.2} & \meanstd{\underline{82.6}}{2.9} & \meanstd{\textbf{86.5}}{2.8} & .007$^{**}$ \\
         \bottomrule
    \end{tabular}
    \label{tab: results_graph_clf}
\end{table*}

The methods are tested on 3 citation and 1 collaboration networks (details in \ref{appendix:datasets}). 
The number of clusters $K$ is set to be equal to the number of vertex classes. 
Averaged results from 10 independent runs of each method are in Tab.~\ref{tab: results_clustering}, which reports the Normalized Mutual Information (NMI) and the accuracy (ACC) between the vertex labels and the cluster assignments sorted with the Kuhn-Munkres algorithm. 
Overall, TVGNN outperforms every other method in terms of both NMI and ACC. 

While NMI and ACC quantify of how well the clusters match the true vertex labels, they do not measure the sharpness of the soft cluster assignments $\Sb$ generated by the GNN-based methods. 
To evaluate sharpness, we propose the following procedure.
Let $\hat \y$ be the assignments obtained from the Kuhn-Munkres algorithm, which minimizes the mismatch between the discrete cluster assignments $\cb$ and the labels $\y$. 
By sorting the rows of $\Sb$ according to the indices given by $\texttt{argsort}(\hat \y)$, $\Sb \Sb^T$ will exhibit a block-diagonal structure if the vertices are assigned with high confidence to only one cluster, i.e., if the cluster assignments are sharp. 
Instead, if the assignments in $\Sb$ are smooth, non-zero elements will appear on the off-diagonal of $\Sb \Sb^T$. 
In addition, the size of each block in $\Sb \Sb^T$ indicates the cluster size. 
Fig.~\ref{fig: sharpness_plot_cora} shows $\log(\Sb \Sb^T)$ for the cluster assignments obtained for the Cora dataset.
The soft assignments given by TVGNN are much sharper than those of DiffPool, MinCutPool, and DMoN, since most of the non-zero values lie within the blocks on the diagonal. 
Notably, the assignments given by DiffPool are so smooth that discerning any structure in $\Sb \Sb^T$ is impossible.

The cluster assignments are computed from vertex features $\X^{(L)}$ generated by the last MP layer according to \eqref{eq:mlp}.
To show the separation of the vertex communities before computing the assignments, we project $\X^{(L)}$ in two dimensions using UMAP~\citep{mcinnes2018umap}. 
Fig~\ref{fig: umap_transform} shows the projected features for Cora. 
In the first row, the vertices are colored according to the cluster assignments $\hat \y$ from Kuhn-Munkres; in the second row, according to the true labels $\y$.
Compared to the other methods, TVGNN yields clusters that are better separated and more compact. 
In addition, with TVGNN the class distribution is better aligned with the clustering partition.

Appendices \ref{appendix:additional_plots}, \ref{appendix:denoising}, and \ref{appendix:point_clouds} report additional results and experiments, which further highlight the capabilities of TVGNN in learning sharp and meaningful cluster assignments.

\subsection{Ablation experiment}
To verify the effectiveness of the proposed TVGNN architecture, which combines the loss function in~\eqref{eq:objective_loss} with the GTVConv layers in~\eqref{eq:gtvconv}, we conduct an ablation study with two modified configurations. 
The numerical results are reported at the bottom of Tab.~\ref{tab: results_clustering}. 
Here, ABL1 denotes a configuration where the proposed loss is replaced by the MinCutPool loss function in~\eqref{eq:mincut}.
The MinCutPool loss was chosen because it is closely related to the spectral clustering loss, which we aim at improving by basing our loss on the components of a tighter relaxation of the Asymmetric Cheeger cut. 
On the other hand, ABL2 denotes a configuration where the GTVConv layers, which minimize the GTV, are replaced by GCN layers, which minimize the LQV~\eqref{eq:lap_smooth_comb} instead. 
The plots of $\Sb\Sb^T$ and the UMAP transform of the latent representation are in Fig.~\ref{fig: sharpness_plot_ablation_cora} and \ref{fig: umap_transform_ablation}, respectively. 
Compared to the proposed model, in both ablation experiments, the clusters are less separated and compact.
In addition, there is a worse correspondence between clusters and class labels.

\subsection{Supervised graph classification}

This task consists in assigning each graph $\mathcal{G}_i$ to a class $y_i$.
We adopt the deep architecture described in Sec.~\ref{sec:architectures}, where MP layers are interleaved with a pooling layer.
As pooling methods, we consider DiffPool, MinCutPool, DMoN, SAGPool~\citep{lee2019self}, Top-$K$~\citep{cangea2018towards, gao2019graph}, and the proposed TVGNN.
All GNN architectures follow the same general configuration depicted in Fig.~\ref{fig:architectures}b, with the main difference that TVGNN adopts GTVConv rather than standard MP layers (hyperparameters and other details are in Appendix \ref{appendix:hyperparams}).
The unsupervised clustering losses in DiffPool, MinCutPool, DMoN, and TVGNN are computed at each pooling layer and then combined with the supervised cross-entropy loss at the end.
The GNNs with SAGPool and Top-$K$ do not have auxiliary losses and are trained only with the cross-entropy. 

We consider 9 graph classification datasets (details in \ref{appendix:datasets}).
Training and testing are done with a stratified 5-fold train/test split.
In addition, 10\% of the training set is used as a validation set using a random stratified split.
For each fold, we perform 3 independent runs and we train until we reach early stopping by measuring the validation loss.
To make the comparison fair, all methods are evaluated on the exact same splits.   
The classification accuracy for each dataset is reported in Table~\ref{tab: results_graph_clf}. 
The GNN models based on TVGNN achieve the highest mean accuracy on all datasets and the differences with the second-best performing method are statistically significant in most of the cases.

\section{Conclusions}
We introduced a novel graph neural network that significantly improves the performance of previous GNN models based on spectral clustering. 
To obtain compact and well-separated clusters, we derived an unsupervised loss from the Asymmetric Cheeger cut, which minimizes the graph total variation of the cluster assignments. 
Remarkably, this is the first attempt to adapt a tighter relaxation of the $K$-cut problem to neural networks and to apply the Asymmetric Cheeger cut relaxation for clustering vertices of attributed graphs.

To facilitate the minimization of the proposed loss function, we introduced GTVConv, a message-passing layer that updates the vertex features by following the gradient of their graph total variation.  
The formal derivation demands GTVConv to use a different connectivity matrix to process each one of the vertex features, which is intractable.
Therefore, we approximated the gradient descent step of GTV with a single connectivity matrix that accounts for all the vertex features at once, reducing the model complexity and facilitating its training.
An appealing property of the GTVConv layer is its capability to adjust the edge weights based on the vertex representations, which are learned in a data-driven fashion.
Our extensive experimental evaluation showed that TVGNN outperforms every other competing method in vertex clustering and graph classification tasks.
In particular, our model always separates well the vertex features and generates sharp cluster assignments.

In this work, the GTVConv was used in conjunction with the proposed unsupervised loss in GNN architectures for vertex clustering and graph classification.
However, GTVConv could also be used as a stand-alone MP layer in GNNs for tasks such as semi-supervised vertex classification.

\subsubsection*{Acknowledgements}
The authors gratefully acknowledge the support of Nvidia Corporation with the donation of the RTX A6000 GPUs used to perform the experimental evaluation.

\bibliography{biblio}
\bibliographystyle{icml2023}

\newpage
\appendix
\onecolumn
\section{Derivations and additional discussions}

\subsection{Graph cut in matrix form}
\label{appendix:cut}
To see that the cut between $C_k$ and its conjugate $\bar{C}_k$ can be expressed in matrix form, we first write the cut as 
\begin{equation*}
    \label{eq:cut}
    \text{cut}(C_k, \bar{C}_k) =  \mkern-18mu \sum \limits_{i \in C_k, j \in \bar C_k} \mkern-18mu a_{ij}(1- z_i z_j),
\end{equation*}
where $z_i, z_j \in \{ -1, 1 \}$ are cluster indicators, i.e., $z_i = 1$ if vertex $i\in C_k$ and $z_i = -1$ if  $i\notin C_k$.

Then,
\begin{equation*}
    \begin{aligned}
    & \sum \limits_{i \in C_k, j \in \bar C_k} a_{ij}(1- z_i z_j) = \sum \limits_{i \in C_k, j \in \bar C_k} a_{ij} \left(\frac{z_i^2 + z_j^2}{2} - z_i z_j \right) \\[5pt]
    & = \frac{1}{2} \sum \limits_{i \in C_k} \Bigg[ \sum \limits_{j \in \bar C_k} a_{ij} \Bigg]z_i^2 + \frac{1}{2} \sum \limits_{j \in \bar C_k} \Bigg[ \sum \limits_{i \in C_k} a_{ij} \Bigg] z_j^2 - \sum \limits_{i \in C_k, j \in \bar C_k} a_{ij} z_i z_j \\[5pt]
    & = \frac{1}{2} \sum \limits_{i \in C_k} d_{ii} z_i^2 + \frac{1}{2} \sum \limits_{j \in \bar C_k} d_{jj} z_j^2 - \z^T \A \z \\[5pt]
    & = \z^T \D \z - \z^T \A \z = \z^T \Lb \z.
    \end{aligned}
\end{equation*}


\subsection{Derivation of the GTVConv aggregation}
\label{appendix:GTVConv}
The graph total variation of a graph with univariate node attributes is defined as
\begin{equation*}
    ||\x||_\text{GTV} = \sum_{i=1}^N \sum_{j=1}^N a_{ij} |x_i-x_j|.
\end{equation*}
If the graph has no self-loops ($a_{ii}=0\;\; \forall\; i$) and $x_i\neq x_j\;\;\forall\; j\neq i$, the partial derivative of GTV with respect to $x_k$ is
\begin{equation*}
    \frac{\partial}{\partial x_k} (||\x||_\text{GTV}) = \sum_{\substack{j=1\\j\neq k}}^N a_{kj} \frac{x_k-x_j}{|x_k-x_j|} + \sum_{\substack{j=1\\j\neq k}}^N a_{jk} \frac{x_k-x_j}{|x_j-x_k|}
\end{equation*}
To achieve differentiability at all points, we define an approximated GTV function as
\begin{equation}
\label{eq:epsGTV}
    (||\x||_\text{GTV})_\epsilon = \sum_{i=1}^N \sum_{j=1}^N \left [I(|x_i - x_j| \geq \epsilon)\left( a_{ij} |x_i-x_j| \right) + I(|x_i - x_j| < \epsilon) \left(a_{ij}\frac{(x_i - x_j)^2}{2\epsilon} + \frac{a_{ij}\epsilon}{2}\right) \right],
\end{equation}
where $I(\cdot)$ is the indicator function. The function in \eqref{eq:epsGTV} is the one displayed on the left in Fig.~\ref{fig:functions}. Compared to the original GTV function, \eqref{eq:epsGTV} is smooth in the vicinity of $|x_i - x_j|=0$.
Importantly, this approximation removes the discontinuity in the derivative at $|x_i - x_j|=0$, which now becomes a piece-wise linear function, as shown on the right in  Fig.~\ref{fig:functions}. 

By defining the matrix $\mathbf{\Gamma}$ whose $(ij)$-th entry is
\begin{equation*}
    \gamma_{ij} = \frac{a_{ij}}{\max\{|x_i-x_j|, \epsilon\}},
\end{equation*}
the partial derivative of the approximate GTV can be expressed as 
\begin{align*}
    \frac{\partial}{\partial x_k} (||\x||_\text{GTV})_\epsilon &= \sum_{j=1}^N \gamma_{kj}(x_k-x_j) + \sum_{j=1}^N \gamma_{jk}(x_k-x_j)\\[5pt]
    &= x_k\sum_{j=1}^N \gamma_{kj} - \sum_{j=1}^N \gamma_{kj} x_j + x_k\sum_{j=1}^N \gamma_{jk} - \sum_{j=1}^N \gamma_{jk} x_j\\[5pt]
    &= x_k d_{k} - \sum_{j=1}^N \gamma_{kj} x_j + x_k d^*_{k} - \sum_{j=1}^N \gamma_{jk} x_j, \;\;\;\text{where}\; d_k=\sum_j \gamma_{kj},\;\; d^*_k=\sum_j \gamma_{jk} \\[5pt]
    &= \D_{\mathbf{\Gamma}_k} \x - \mathbf{\Gamma}_k \x + \D^*_{\mathbf{\Gamma}_k} \x - (\mathbf{\Gamma}^T)_k \x, 
\end{align*}
where $\mathbf{\Gamma}_k$ and $\D_{\mathbf{\Gamma}_k}$ represent the $k$-th row of $\mathbf{\Gamma}$ and $\D_{\mathbf{\Gamma}}=\text{diag}(\mathbf{\Gamma}\boldsymbol{1})$, respectively. The full gradient of the approximated GTV of $\x$ is furthermore
\begin{equation*}
    \nabla (||\x||_\text{GTV})_\epsilon = \D_\mathbf{\Gamma}\x - \mathbf{\Gamma}\x + \D^*_\mathbf{\Gamma}\x - \mathbf{\Gamma}^T\x.
\end{equation*}
If the graph is undirected, i.e., $\gamma_{ij} = \gamma_{ji}$, the gradient can be simplified as
\begin{equation*}
        \nabla (||\x||_\text{GTV})_\epsilon = 2(\D_\mathbf{\Gamma} - \mathbf{\Gamma})\x = 2\Lb_\mathbf{\Gamma}\x, 
\end{equation*}
where $\Lb_\mathbf{\Gamma}$ is the Laplacian associated with $\mathbf{\Gamma}$.
Finally, the gradient descent step update of the approximated GTV of $\x$ with step size $\delta$ becomes
\begin{align*}
    \x^{(t+1)} &= \x^{(t)} - \delta\nabla (||\x||_\text{GTV})_\epsilon^{(t)} \\[5pt]
    &= \x^{(t)} - 2\delta\Lb_\mathbf{\Gamma}^{(t)}\x^{(t)}\\[5pt]
    &= \left(\I - 2\delta\Lb_\mathbf{\Gamma}^{(t)} \right)\x^{(t)}.
\end{align*}

\subsection{Replacing multiple $\mathbf{\Gamma}^f$ with a single $\mathbf{\hat{\Gamma}}$}
\label{appendix:single_vs_multi}

\begin{figure*}[!ht]
    \centering
    \small
    \begin{subfigure}{\textwidth}
        \centering
        \includegraphics[width=.9\textwidth]{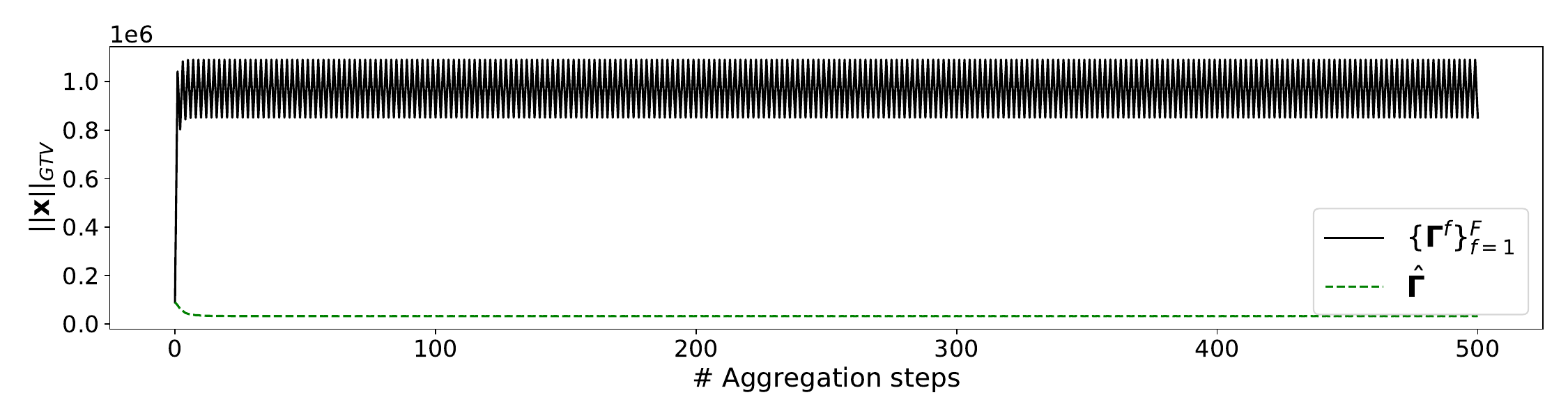}
        \caption{}
    \end{subfigure}
    \begin{subfigure}{\textwidth}
        \centering
        \includegraphics[width=.9\textwidth]{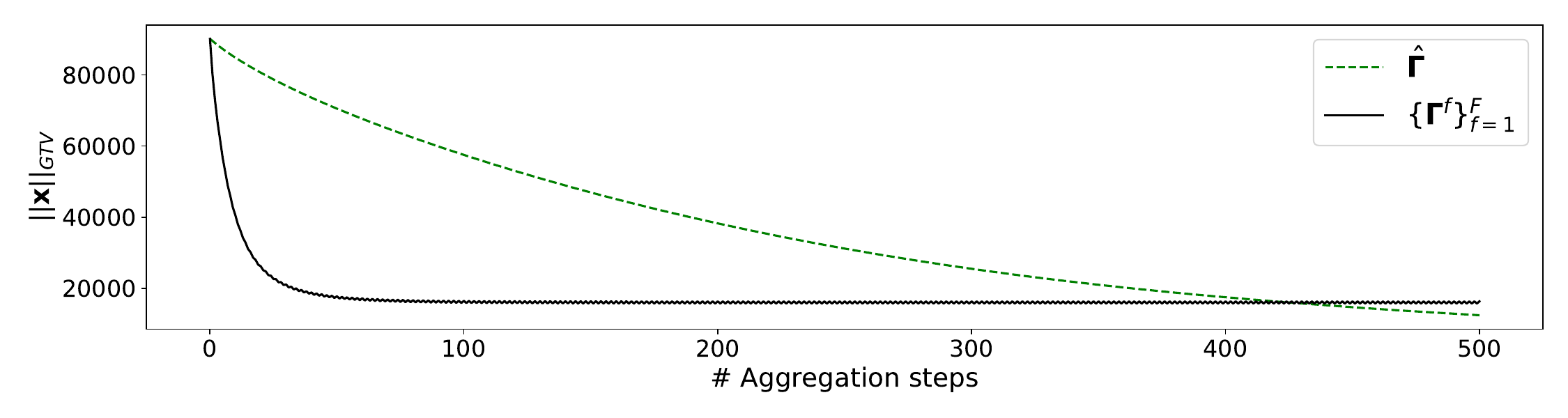}
        \caption{}
    \end{subfigure}
    \caption{Evolution of GTV for the vertex features of Cora with respect to the number of aggregation steps when using a connectivity matrix $\mathbf{\Gamma}^f$ for each feature or a single connectivity matrix $\mathbf{\hat{\Gamma}}$. In (a), we use gradient step $2\delta=0.311$; In (b), we use gradient step $2\delta=0.005$.}
    \label{fig:single_vs_multi}
\end{figure*}

As discussed in Sec.~\ref{sec:gtvconv_layer} and according to \eqref{eq:gamma_multi}, in the presence of multiple vertex features a different connectivity matrix  $\mathbf{\Gamma}^f, f=1, \dots, F$ must be used to independently aggregate each feature $f$.
We propose to simplify the procedure, by aggregating all features using the single connectivity matrix $\mathbf{\hat{\Gamma}}$ defined in \eqref{eq:gamma_single}.
The proposed simplification offers two important advantages.
The first and more obvious involves computational complexity: only one connectivity matrix needs to be stored and a single aggregation procedure must be performed.
A second advantage is that using $\mathbf{\hat{\Gamma}}$ facilitates the minimization of the GTV across the vertex features.
We illustrate this second point with an example.

In the following, we consider the Cora dataset but the same behavior is observed also for the other datasets.
To reproduce a typical GNN setting, the features $\X^{(0)}$ are constructed by mapping the original one-hot encoded features into a 32-dimensional space through the multiplication with a kernel, whose values are randomly sampled from a standard normal distribution.
Then, we apply several times the aggregation defined in equations \eqref{eq:gamma_multi} and \eqref{eq:gamma_single}.
Fig.~\ref{fig:single_vs_multi} depicts how the GTV of the vertex features, $\| \X \|_\text{GTV}$, decreases when the vertex features are aggregated using multiple matrices $\{\mathbf{\Gamma}^f\}_{f=1}^F$ (black line) or a single connectivity matrix $\mathbf{\hat{\Gamma}}$ (green dashed line).
In the first plot, Fig.~\ref{fig:single_vs_multi}(a), we use gradient step $2\delta=0.311$, which is the same used in the vertex clustering experiment.
In Fig.~\ref{fig:single_vs_multi}(b) we set $2\delta=0.005$.
First, we notice that for the larger gradient step, the aggregation based on multiple connectivity matrices performs poorly as the minimization gets stuck oscillating around some local minima. This can happen because the gradients of the GTV for each feature $f$ are pulling toward different directions.
Decreasing step size (Fig.~\ref{fig:single_vs_multi}(b)), which corresponds to lowering the contribution from neighbors when aggregating, makes the scheme based on multiple $\mathbf{\Gamma}^f$ more stable.
Nevertheless, even in this case the aggregation scheme based on the single $\mathbf{\hat{\Gamma}}$ matrix eventually converges to a better solution.

In conclusion, when using $\mathbf{\hat{\Gamma}}$ the optimization of GTV is generally more stable, allows to find better minima, and is more robust to the selection of the gradient step $\delta$.

\subsection{Weighted LQV and GTV}
\label{appendix:GTVConv_weight}

Let us define the LQV weighted by the vertex degrees as
\begin{equation*}
    ||\x||_{\text{LQV}_\text{W}} = \frac{1}{4}\sum_{i=1}^N \sum_{j=1}^N \left(\sqrt{\frac{a_{ij}}{d_i}}x_i - \sqrt{\frac{a_{ij}}{d_j}}x_j\right)^2 = \frac{1}{4}\sum_{i=1}^N \sum_{j=1}^N a_{ij} \left(\frac{x_i}{\sqrt{d_i}} - \frac{x_j}{\sqrt{d_j}}\right)^2.
\end{equation*}

Then, under the assumption that the graph is undirected, we have that

\begin{align*}
\frac{\partial}{\partial x_k}\left(||\x||_{\text{LQV}_\text{W}}\right) &= \frac{1}{\sqrt{d_k}} \sum_j a_{kj} \left( \frac{x_k}{\sqrt{d_k}} - \frac{x_j}{\sqrt{d_j}} \right) \\[5pt]
                             &= \frac{x_k}{d_k} \sum_j a_{kj} - \sum_j \frac{a_{kj}}{\sqrt{d_k}\sqrt{d_j}} x_j \\[5pt]
                             &= (\I \x)_k - (\D^{-1/2} \A \D^{-1/2} \x)_k \\[5pt]
                             \nabla\left(||\x||_{\text{LQV}_\text{W}}\right) &= (\I - \D^{-1/2} \A \D^{-1/2}) \x,
\end{align*}
where $(\I - \D^{-1/2} \A \D^{-1/2})$ is the symmetrically normalized Laplacian.

If we compute the gradient descent update we obtain
\begin{align*}
    \x^{(t+1)} &= \x^{(t)} - \delta  \nabla\left(||\x||_{\text{LQV}_\text{W}}\right)^{(t)}\\[5pt]
    &= \x^{(t)} - \delta  (\I - \D^{-1/2} \A \D^{-1/2}) \x^{(t)}.
\end{align*}

When the gradient step size is $\delta=1$, we get
\begin{equation*}
    \x^{(t+1)} = \D^{-1/2} \A \D^{-1/2}\x^{(t)},
\end{equation*}
which closely resembles the aggregation function used by a GCN to update the vertex features.

Now, let us define the degree-weighted GTV as
\begin{equation*}
    ||\x||_{\text{GTV}_\text{W}} = \frac{1}{2}\sum_{i=1}^N \sum_{j=1}^N \left|\sqrt{\frac{a_{ij}}{d_i}}x_i - \sqrt{\frac{a_{ij}}{d_j}}x_j\right| = \frac{1}{2}\sum_{i=1}^N \sum_{j=1}^N |q_{ij}|,
\end{equation*}
with $q_{ij} = \sqrt{\frac{a_{ij}}{d_i}}x_i - \sqrt{\frac{a_{ij}}{d_j}}x_j$ and $d_i = \D_{ii}$, where $\D = \text{diag}(\A\boldsymbol{1})$. The numerically stable approximation to the derivative of $q_{kj}$ w.r.t. $x_k$ is then given by
\begin{equation*}
     \frac{\partial}{\partial x_k}(q_{kj})_\epsilon = \begin{cases} \sqrt{\dfrac{a_{kj}}{d_k}}\cdot \dfrac{q_{kj}}{|q_{kj}|}, & |q_{kj}|\geq\epsilon\\[10pt]\sqrt{\dfrac{a_{kj}}{d_k}}\cdot \dfrac{q_{kj}}{\epsilon}, & |q_{kj}|<\epsilon \end{cases}
\end{equation*}
which can be rewritten in a more compact form as
\begin{equation*}
    \frac{\partial}{\partial x_k}(q_{kj})_\epsilon = \sum_j \sqrt{\frac{a_{kj}}{d_k}}\frac{q_{kj}}{\max\{\epsilon, |q_{kj}|\}}=\sum_j \frac{a_{kj}}{\sqrt{d_k}}\frac{\frac{x_k}{\sqrt{d_k}}-\frac{x_j}{\sqrt{d_j}}}{\max\{\epsilon, |q_{kj}|\}}.
\end{equation*}

Now let 
\begin{equation*}
    \gamma_{ij} = \frac{a_{ij}}{\max\{\epsilon, |q_{ij}|\}}
\end{equation*}

By substituting $\gamma_{kj}$ and assuming, once again, that the graph is undirected we get
\begin{align*}
     \frac{\partial}{\partial x_k}\left(|\x||_{\text{GTV}_\text{W}}\right)_\epsilon &= \sum_j \frac{1}{\sqrt{d_k}}\gamma_{kj} \left(\frac{x_k}{\sqrt{d_k}} - \frac{x_j}{\sqrt{d_j}}\right)\\[5pt]
     &= \frac{x_k}{d_k} \sum_j \gamma_{kj} - \sum_j \frac{\gamma_{kj}}{\sqrt{d_k}\sqrt{d_j}}x_j\\[5pt]
     &= (\D^{-1}\D_{\Gamma})_i\hspace{2pt}\x - (\D^{-1/2}\mathbf{\Gamma}\D^{-1/2})_i\hspace{2pt}\x\\[5pt]
     &= (\D^{-1/2}\D_{\mathbf{\Gamma}}\D^{-1/2})_i\hspace{2pt}\x - (\D^{-1/2}\mathbf{\Gamma}\D^{-1/2})_i\hspace{2pt}\x,
\end{align*}
where $\D_{\Gamma} = \text{diag}(\mathbf{\Gamma}\boldsymbol{1})$. 

The gradient of the weighted GTV is
\begin{align*}
    \nabla\left(||\x||_{\text{GTV}_\text{W}}\right) &= \D^{-1/2}\D_{\mathbf{\Gamma}}\D^{-1/2} \x - \D^{-1/2}\mathbf{\Gamma}\D^{-1/2} \x\\[5pt]
    &= \D^{-1/2} (\D_{\mathbf{\Gamma}} - \mathbf{\Gamma})\D^{-1/2} \x\\[5pt]
    &= \D^{-1/2} \Lb_{\mathbf{\Gamma}} \D^{-1/2} \x
\end{align*}

A gradient descent step for minimizing the weighted GTV is given by
\begin{align*}
    \label{eq: gradient_descent_GTV_alt}
    \x^{(t+1)} &= \x^{(t)} - \delta  \nabla\left(||\x||_{\text{GTV}_\text{W}}\right)^{(t)}\\[5pt]
    &= \x^{(t)} - \delta  \D^{-1/2} \Lb_{\mathbf{\Gamma}}^{(t)} \D^{-1/2} \x^{(t)}\\[5pt]
    &= (\I - \delta  \D^{-1/2} \Lb_{\mathbf{\Gamma}}^{(t)} \D^{-1/2})\x^{(t)} 
\end{align*}
which is equivalent to the aggregation step of a GCN with $\tilde{\A} = \I - \delta\D^{-1/2} \Lb_{\mathbf{\Gamma}}^{(t)} \D^{-1/2}$.

We notice the analogy between the updates derived from the degree-weighted LQV, which is 
$\x^{(t+1)} = \x^{(t)} - \delta  (\I - \D^{-1/2} \A \D^{-1/2}) \x^{(t)}$,
and from the degree-weighted GTV, which is
$\x^{(t+1)} = (\I - \delta  \D^{-1/2} \Lb_{\mathbf{\Gamma}}^{(t)} \D^{-1/2})\x^{(t)} $.

\section{Details of the experimental setting}

\subsection{Software libraries}
The GNN models were implemented using both Spektral\footnote{\url{https://graphneural.network}}~\citep{grattarola2020graph} and Pytorch Geometric\footnote{\url{https://pytorch-geometric.readthedocs.io}}~\citep{fey2019fast}. 
The methods for vertex embedding used in the vertex clustering experiment are based on the Karateclub\footnote{\url{https://karateclub.readthedocs.io}}~\citep{rozemberczki2020karateclub} implementation, and most of the datasets are taken from Pytorch Geometric.

\subsection{Datasets details}
\label{appendix:datasets}

\bgroup
\def\arraystretch{1.0} 
\setlength\tabcolsep{.2em} 
\begin{table}[!ht]
\footnotesize
\centering
\caption{Details of the vertex clustering datasets.} 
\label{tab:nc_dataset}
\begin{tabular}{lcccc}
\cmidrule[1.5pt]{1-5}
\textbf{Dataset} & \textbf{\#Vertices} & \textbf{\#Edges} & \textbf{\#Vertex attr.} & \textbf{\#Vertex classes} \\
\cmidrule[.5pt]{1-5}
Cora     & 2,708  & 10,556   & 1,433   & 7    \\
Citeseer & 3,327  & 9,104    & 3,703   & 6    \\
Pubmed   & 19,717 & 88,648   & 500    & 3    \\
DBLP     & 17,716 & 105,734  & 1,639   & 4    \\
\cmidrule[1.5pt]{1-5}
\end{tabular}
\end{table}
\egroup

\bgroup
\def\arraystretch{1.0} 
\begin{table*}[!ht]
\footnotesize
\centering
\caption{Details of the graph classification datasets.} 
\label{tab:gc_dataset}
\begin{tabular}{lcccccc}
\cmidrule[1.5pt]{1-7}
\textbf{Dataset} & \textbf{\#Samples} & \textbf{\#Classes} & \textbf{Avg. \#vertices} & \textbf{Avg. \#edges} & \textbf{Vertex attr.} & \textbf{Vertex labels} \\
\cmidrule[.5pt]{1-7}
Bench-easy    & 1,800  & 3  & 147.82 & 922.67    & -- & yes \\
Bench-hard    & 1,800  & 3  & 148.32 & 572.32    & -- & yes \\
MUTAG         & 188    & 2  & 17.93  & 19.79     & -- & yes \\
Mutagenicity  & 4,337   & 2  & 30.32  & 61.54    & -- & yes \\
NCI1          & 4,110   & 2  & 29.87  & 64.60    & -- & yes \\
Proteins      & 1,113  & 2  & 39.06  & 72.82     & 1  & yes  \\
D\&D          & 1,178  & 2  & 284.32 & 1,431.32  & -- & yes \\
COLLAB        & 5,000  & 3  & 74.49  & 4,914.43  & --  & no \\ 
REDDIT-BINARY & 2,000  & 2  & 429.63 & 995.51    & --  & no\\
\cmidrule[1.5pt]{1-7}
\end{tabular}
\end{table*}
\egroup


In the vertex clustering experiment, we considered the citation networks Cora, Pubmed, Citeseer~\citep{10.5555/3045390.3045396} and DBLP~\citep{fu2020magnn}.
In the graph classification experiment, we analyzed seven TUD datasets~\citep{Morris+2020} and two synthetic datasets, Bench-easy and Bench-hard~\citep{bianchi2022pyramidal}.

Details about the datasets are reported in Tab.~\ref{tab:nc_dataset} and \ref{tab:gc_dataset}. 
In the graph classification datasets, the vertex feature matrix $\X$ consists of vertex attributes, vertex labels, or a concatenation of both. For the datasets where neither the vertex attributes nor the vertex labels are available, a one-hot encoded vertex degree matrix was used as a surrogate feature for $\X$. 
Furthermore, motivated by the work of \cite{ivanov2019understanding}, the datasets were cleaned such that they only contained non-isomorphic graphs.

\subsection{Hyperparameters configuration}
\label{appendix:hyperparams}

\begin{table}[!ht]
    \centering
    \footnotesize
    \caption{Hyperparameters configuration for the vertex clustering and graph classification tasks. 
    $\sigma_\text{MP}$ indicates the activation of the MP layers, $\sigma_\text{MLP}$ is the activation of the MLP layers, $\delta$ is the step-size in the GTVConv layer, $\alpha_1$ is the coefficient for the total variation loss $\mathcal{L}_\text{GTV}$, $\alpha_2$ is the coefficient for the balance loss $\mathcal{L}_\text{AN}$, $\ell_2$ indicates the weight of the $\ell_2$ regularization on the GNN weight parameters. The values of \# MP layers and \# MLP layers are the numbers of MP and MLP layers, respectively, in each of the blocks of the architectures for clustering and classification presented in Section~\ref{sec:architectures}. For example, the architecture used for vertex clustering uses $2 \times 1$ MP and $1 \times 1$ MLP layers, while the architecture used in Proteins uses $3 \times 3$ MP and $3 \times 1$ MLP layers.}
    \begin{tabular}{l c c c c c}
        \cmidrule[1.5pt]{1-6}
        \textbf{Parameters} & 
            \textbf{Vertex Clustering} & \textbf{Bench-easy} & \textbf{Bench-hard} & \textbf{MUTAG} & \textbf{Mutagenicity} \\
        \midrule
        \# MP layers        & 2     & 1     & 1     & 1     & 3          \\
        \# MP channels      & 512   & 32    & 32    & 32    & 32        \\
        $\sigma_\text{MP}$  & ELU   & ReLU  & ReLU  & ELU   & ReLU    \\
        $2\delta$            & 0.311 & 0.724 & 2.288 & 1.644 & 3.077  \\
        \# MLP layers       & 1     & 3     & 1     & 3     & 2          \\
        \# MLP channels     & 256   & 64    & 64    & 64    & 32       \\
        $\sigma_\text{MLP}$ & ReLU  & ReLU  & ReLU  & ReLU  & ELU    \\
        $\alpha_1$          & 0.785 & 0.594 & 0.188 & 0.623 & 0.726  \\
        $\alpha_2$          & 0.514 & 0.974 & 0.737 & 0.832 & 0.982  \\
        $\ell_2$            & --    & 1e-5  & 0     & 1e-4  & 1e-5  \\
        Learning rate       & 1e-3  & 1e-3  & 5e-4  & 1e-2  & 5e-4 \\
        \cmidrule[1.5pt]{1-6}
    \end{tabular}
    
    \vspace{1cm}
    
    \begin{tabular}{l c c c c c}
        \cmidrule[1.5pt]{1-6}
        \textbf{Parameters} & \textbf{NCI1} & \textbf{Proteins} & \textbf{D\&D } & \textbf{COLLAB} & \textbf{REDDIT-BINARY}\\
        \midrule
        \# MP layers           & 3 & 3     & 1 & 2              & 3     \\
        \# MP channels         & 32 & 64    & 64 & 256          & 16    \\
        $\sigma_\text{MP}$     & ReLU & ELU   & ELU & ReLU      & ReLU  \\
        $2\delta$               & 2.411 & 2.073 & 0.622 & 0.554  & 1.896 \\
        \# MLP layers          & 3 & 1     & 2 & 2              & 3     \\
        \# MLP channels        & 32 & 16    & 32 & 128          & 16    \\
        $\sigma_\text{MLP}$    & ReLU & ELU   & ReLU & ReLU     & ReLU  \\
        $\alpha_1$             & 0.936 & 0.985 & 0.354 & 0.304  & 0.654 \\
        $\alpha_2$             & 0.639 & 0.751 & 0.323 & 0.801  & 0.962 \\
        $\ell_2$               & 1e-3 & 1e-3  & 1e-5 & 0        & 1e-3  \\
        Learning rate          & 5e-4 & 1e-3  & 1e-5 & 5e-5     & 1e-3  \\
        \cmidrule[1.5pt]{1-6}
    \end{tabular}
    \label{tab:hyperparams}
\end{table}

The hyperparameters for TVGNN for both the vertex clustering and graph classification tasks are reported in Tab.~\ref{tab:hyperparams}. The parameter $\epsilon$ which ensures numerical stability for $\mathbf{\Gamma}$ was set to 1e-3 in all experiments. For MinCutPool~\citep{bianchi2020spectral}, GAE and VGAE~\citep{kipf2016variational} the model configurations are those reported in the original papers. For DiffPool, Top-$K$, and SAGPool configurations were the same as in \citep{bianchi2020spectral}. The models with DMoN used the same hyperparameter configuration as for MinCutPool and the regularization term $\mathcal{L}_r$ in the auxiliary loss is weighted by 1e-1. In the case of DeepWalk, node2vec, NetMF, and TADW we used the default configurations from the Karateclub library~\citep{rozemberczki2020karateclub}.

For the graph classification task, the GNNs with TVGNN, MinCutPool, DiffPool, and DMoN were trained with a batch size of 8 for all datasets, except D\&D and REDDIT-BINARY, for which batch size was set to 1 due to memory constraints. The models with Top-$K$ and SAGPool were trained with a batch size of 1 for all datasets. 
For the vertex clustering task, the GNNs were trained for 10,000 epochs.
In the graph classification task, we performed early stopping on the validation set using patience of 20 epochs.

\subsection{Training times}
\label{appendix:training-times}
Table~\ref{tab: training_times_clustering} and Table~\ref{tab: training_times_graph_clf} report training times for the GNN-based methods for the vertex clustering task and graph classification task, respectively. 
All methods have been given similar capacities in terms of the number of layers and the size of the weight matrices. 
The reported times give a rough indication of the differences in the computational complexity, but they highly depend on how optimized is their implementation. 
Here, the different pooling methods were implemented using Pytorch Geometric~\citep{fey2019fast} and, thus, all methods besides Top-$K$ and SAGPool process dense representations of the graphs.
Overall, we notice no significant differences between the execution time of TVGNN and the other methods.

\begin{table*}[!ht]
    \footnotesize
    \centering
    \caption{\small Training times in milliseconds per epoch for the vertex clustering task.}
    \begin{tabular}{l c c c c}
    \cmidrule[1.5pt]{1-5}
    \textbf{Method} & \textbf{Cora} & \textbf{Citeseer} & \textbf{Pubmed} & \textbf{DBLP}\\
    \cmidrule[.5pt]{1-5}
    DiffPool &
        5.6 & 7.7 & 67.6 & 62.2\\
    MinCutPool &
        8.9 & 13.8 & 272.5 & 227.4\\
    DMoN &
        5.7 & 7.8 & 39.9 & 39.7\\
    TVGNN &
        8.3 & 9.4 & 56.4 & 57.2\\
    \cmidrule[1.5pt]{1-5}
    \end{tabular}
    \label{tab: training_times_clustering}
\end{table*}



\begin{table*}[!ht]
\footnotesize
\centering
\caption{\small Training times in seconds per epoch for the graph classification task.}
\begin{tabular}{l c c c c c c}
     \cmidrule[1.5pt]{1-7}
     \textbf{Dataset} & \textbf{Top-$K$} & \textbf{SAGPool} & \textbf{DiffPool} & \textbf{MinCutPool} & \textbf{DMoN} & \textbf{TVGNN} \\
     \cmidrule[.5pt]{1-7}
     Bench-easy &
         0.39 & 0.39 & 0.39 & 0.39 & 0.41 & 0.49\\
     Bench-hard &
        0.37 & 0.37 & 0.34 & 0.38 & 0.37 & 0.41\\
     MUTAG & 
       0.14 & 0.10 & 0.09 & 0.10 & 0.09 & 0.10\\
     Mutagenicity & 
        0.72 & 0.81 & 0.71 & 0.69 & 0.74 & 0.85\\
     NCI1 &
      0.69 & 0.70 & 0.63 & 0.70 & 0.71 & 0.76\\
     Proteins &
     0.28 & 0.23 & 0.21 & 0.24 & 0.24 & 0.25\\
     D\&D &
       0.32 & 0.34 & 0.61 & 0.63 & 0.45 & 0.50\\
     COLLAB &
        1.03 & 1.18 & 0.99 & 1.12 & 1.11 & 1.52\\
    REDDIT-BINARY &
    0.51 & 0.54 & 1.83 & 1.86 & 1.10 & 1.38\\
    \cmidrule[1.5pt]{1-7}
    \end{tabular}
    \label{tab: training_times_graph_clf}
\end{table*}

\section{Additional results}

\subsection{Additional plots}
\label{appendix:additional_plots}

\begin{figure*}[!ht]
    \centering
    \small
    \begin{subfigure}{0.24\textwidth}
        \centering
        \includegraphics[width=\textwidth]{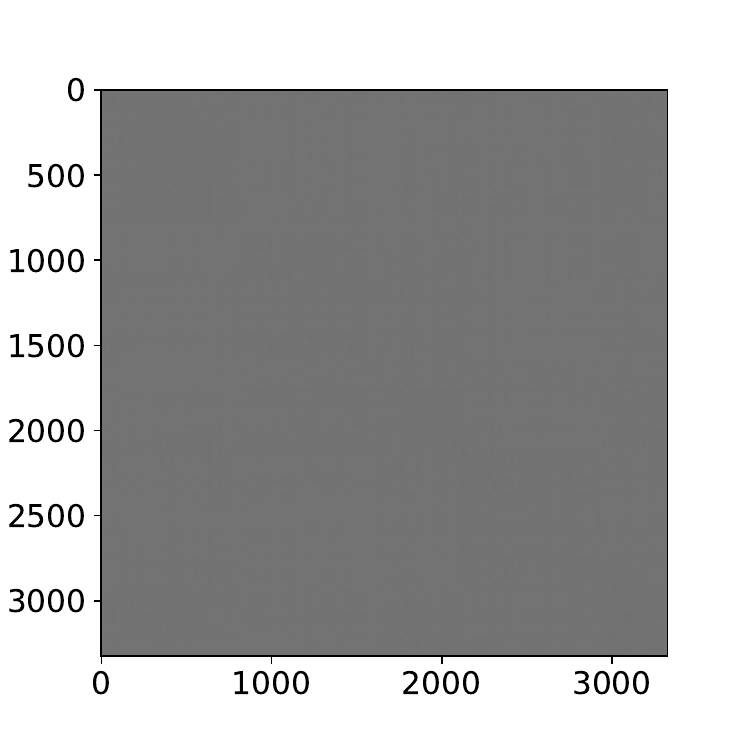}
        \vspace*{-0.7cm}
        \caption{\small DiffPool}
    \end{subfigure}
    \begin{subfigure}{0.24\textwidth}
        \centering
        \includegraphics[width=\textwidth]{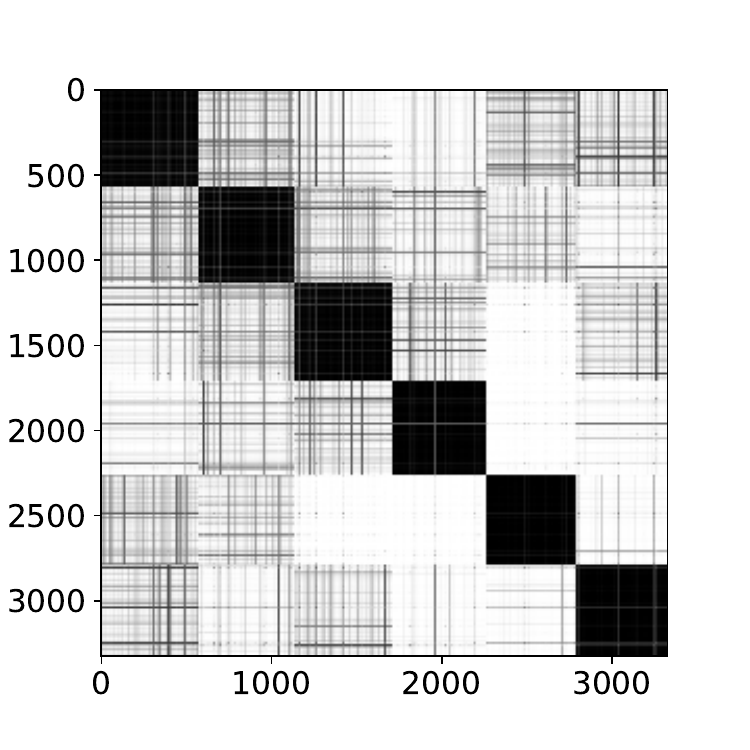}
        \vspace*{-0.7cm}
        \caption{\small MinCutPool}
    \end{subfigure}
    \begin{subfigure}{0.24\textwidth}
        \centering
        \includegraphics[width=\textwidth]{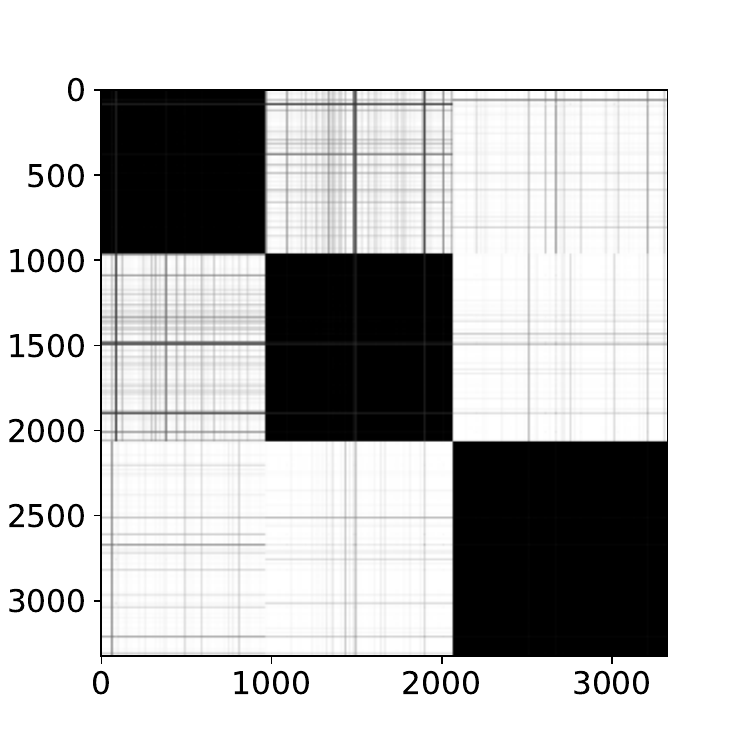}
        \vspace*{-0.7cm}
        \caption{\small DMoN}
    \end{subfigure}
     \begin{subfigure}{0.24\textwidth}
        \centering
        \includegraphics[width=\textwidth]{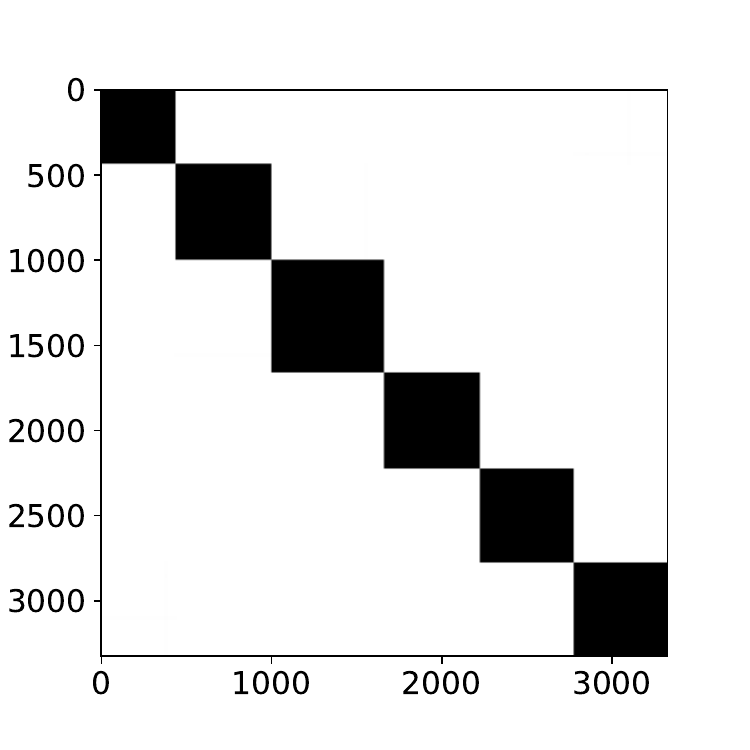}
        \vspace*{-0.7cm}
        \caption{\small TVGNN}
    \end{subfigure}
\vspace*{-0.3cm}
\caption{\small Logarithm of $\Sb\Sb^T$ for Citeseer.}
\label{fig: SST_plot_supplementary_citeseer}
\end{figure*}

\begin{figure*}[!ht]
    \centering
    \small
    \begin{subfigure}{0.24\textwidth}
        \centering
        \includegraphics[width=\textwidth]{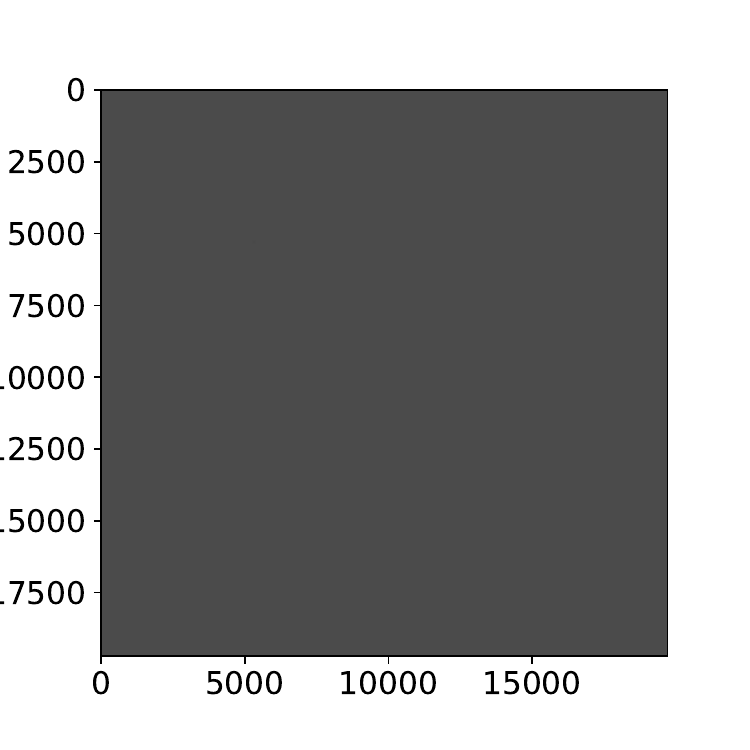}
        \vspace*{-0.7cm}
        \caption{\small DiffPool}
    \end{subfigure}
    \begin{subfigure}{0.24\textwidth}
        \centering
        \includegraphics[width=\textwidth]{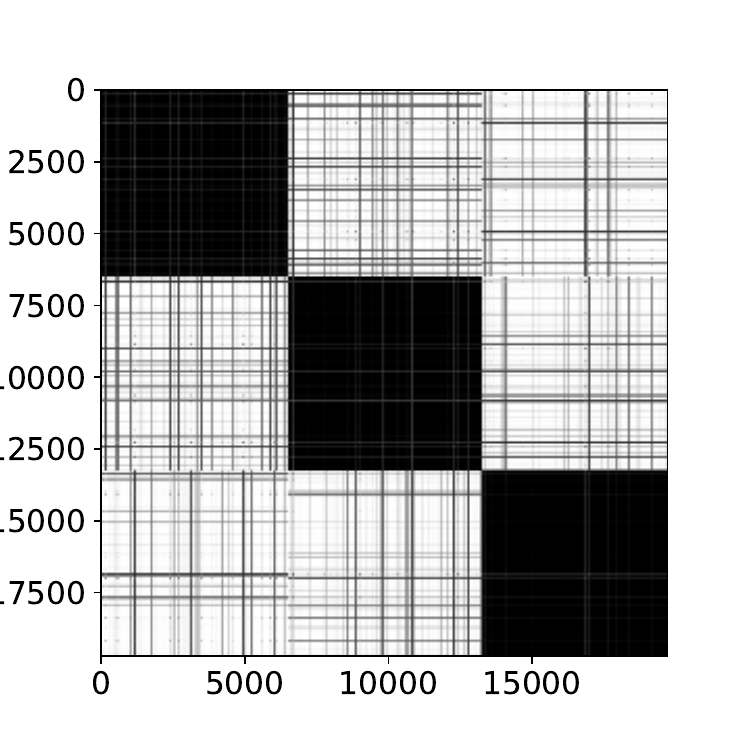}
        \vspace*{-0.7cm}
        \caption{\small MinCutPool}
    \end{subfigure}
    \begin{subfigure}{0.24\textwidth}
        \centering
        \includegraphics[width=\textwidth]{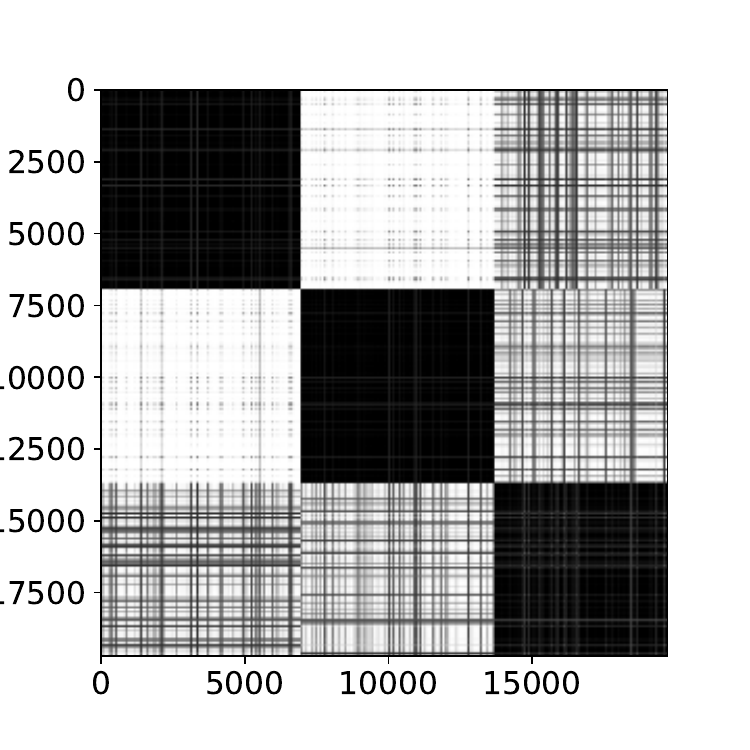}
        \vspace*{-0.7cm}
        \caption{\small DMoN}
    \end{subfigure}
     \begin{subfigure}{0.24\textwidth}
        \centering
        \includegraphics[width=\textwidth]{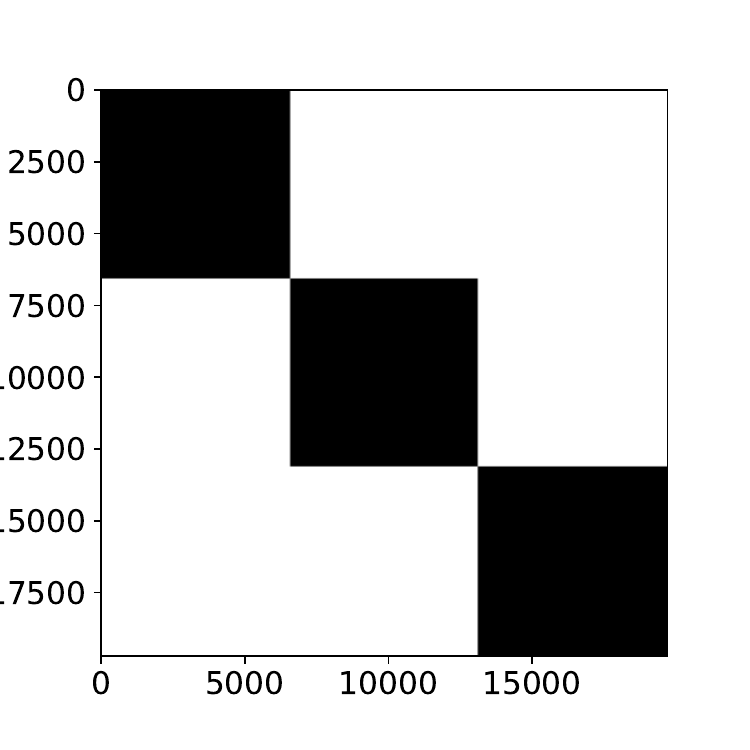}
        \vspace*{-0.7cm}
        \caption{\small TVGNN}
    \end{subfigure}
\vspace*{-0.3cm}
\caption{\small Logarithm of $\Sb\Sb^T$ for Pubmed.}
\label{fig: SST_plot_supplementary_pubmed}
\end{figure*}

\begin{figure*}[!ht]
    \centering
    \small
    \begin{subfigure}{0.24\textwidth}
        \centering
        \includegraphics[width=\textwidth]{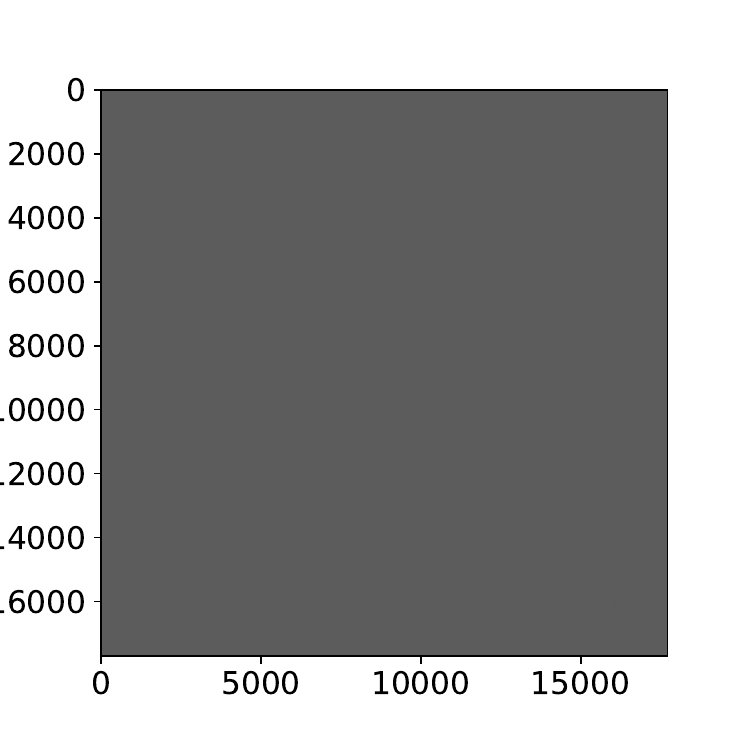}
        \vspace*{-0.7cm}
        \caption{\small DiffPool}
    \end{subfigure}
    \begin{subfigure}{0.24\textwidth}
        \centering
        \includegraphics[width=\textwidth]{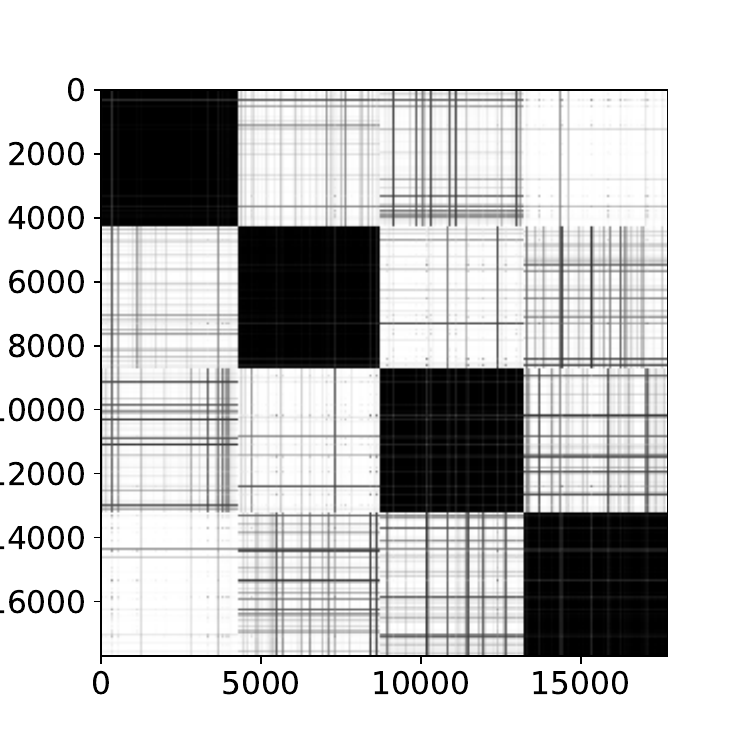}
        \vspace*{-0.7cm}
        \caption{\small MinCutPool}
    \end{subfigure}
    \begin{subfigure}{0.24\textwidth}
        \centering
        \includegraphics[width=\textwidth]{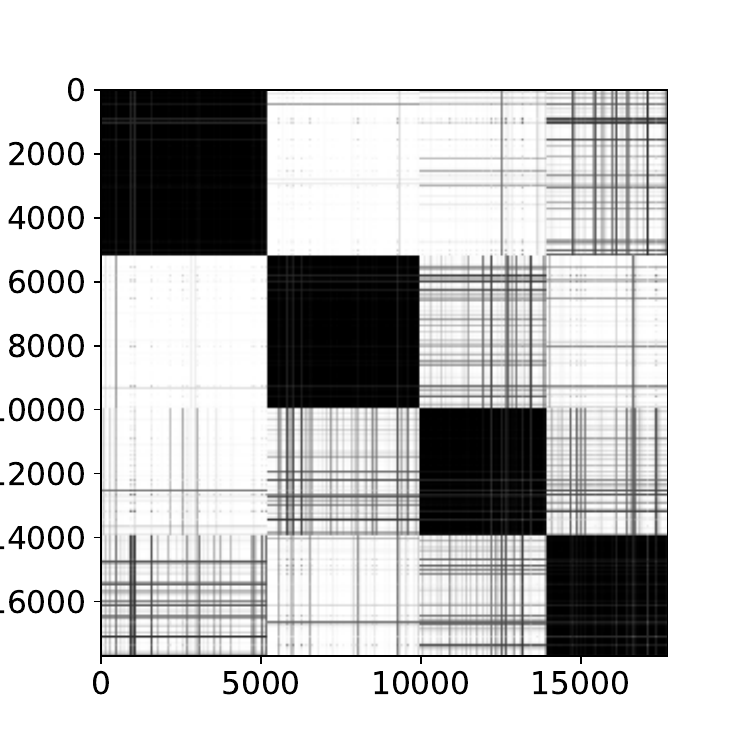}
        \vspace*{-0.7cm}
        \caption{\small DMoN}
    \end{subfigure}
     \begin{subfigure}{0.24\textwidth}
        \centering
        \includegraphics[width=\textwidth]{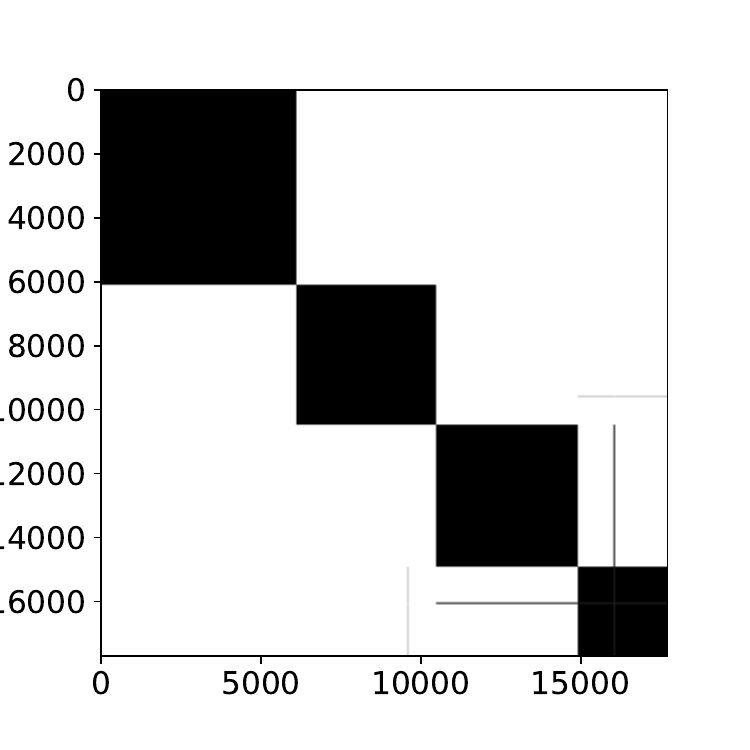}
        \vspace*{-0.7cm}
        \caption{\small TVGNN}
    \end{subfigure}
\vspace*{-0.3cm}
\caption{\small Logarithm of $\Sb\Sb^T$ for DBLP.}
\label{fig: SST_plot_supplementary_dblp}
\end{figure*}

\begin{figure*}[!ht]
    \centering
    \scriptsize
    \vspace*{-0.4cm}
    \begin{subfigure}{0.23\textwidth}
        \centering
        \includegraphics[width=\textwidth]{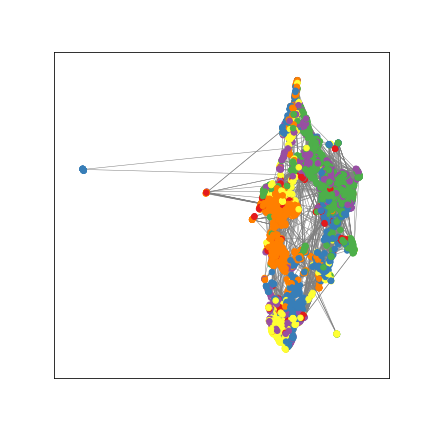}
        \vspace*{-0.8cm}
        \caption{\small DiffPool}
    \end{subfigure}
    \vspace*{-0.4cm}
    \begin{subfigure}{0.23\textwidth}
        \centering
        \includegraphics[width=\textwidth]{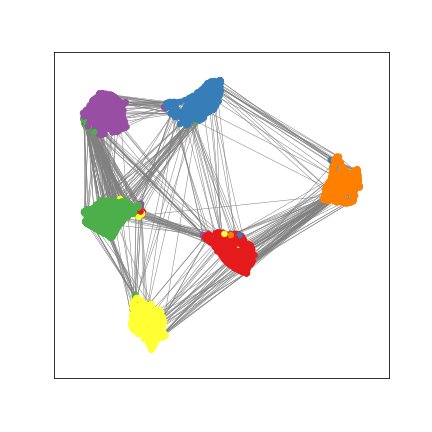}
        \vspace*{-0.8cm}
        \caption{\small MinCutPool}
    \end{subfigure}
    \begin{subfigure}{0.23\textwidth}
        \centering
        \includegraphics[width=\textwidth]{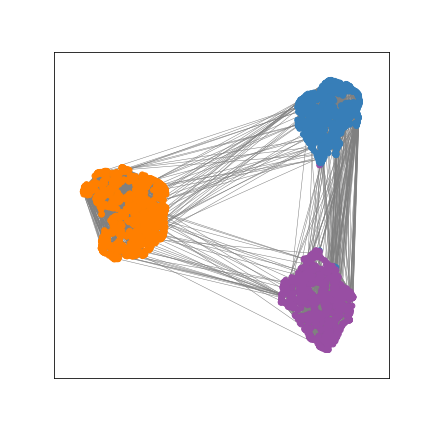}
        \vspace*{-0.8cm}
        \caption{\small DMoN}
    \end{subfigure}
    \begin{subfigure}{0.23\textwidth}
        \centering
        \includegraphics[width=\textwidth]{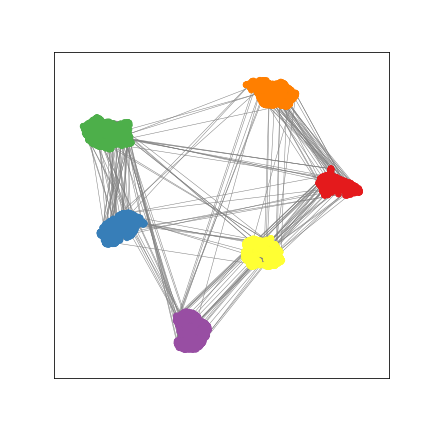}
        \vspace*{-0.8cm}
        \caption{\small TVGNN}
    \end{subfigure}
    \begin{subfigure}{0.23\textwidth}
        \centering
        \includegraphics[width=\textwidth]{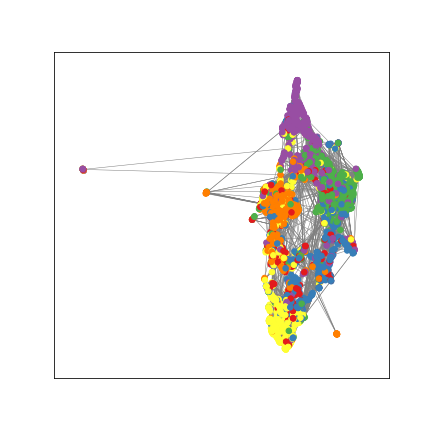}
        \vspace*{-0.8cm}
        \caption{\small DiffPool}
    \end{subfigure}
    \vspace*{-0.4cm}
    \begin{subfigure}{0.23\textwidth}
        \centering
        \includegraphics[width=\textwidth]{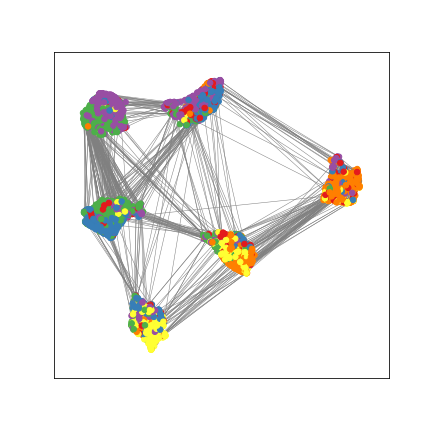}
        \vspace*{-0.8cm}
        \caption{\small MinCutPool}
    \end{subfigure}
    \begin{subfigure}{0.23\textwidth}
        \centering
        \includegraphics[width=\textwidth]{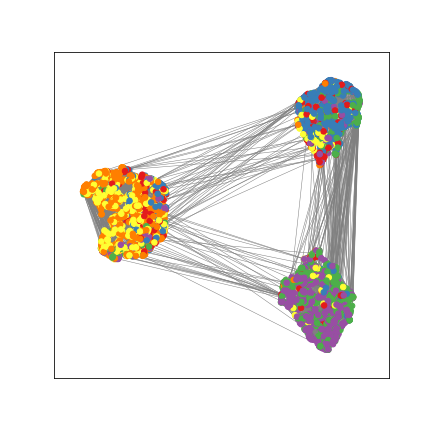}
        \vspace*{-0.8cm}
        \caption{\small DMoN}
    \end{subfigure}
    \begin{subfigure}{0.23\textwidth}
        \centering
        \includegraphics[width=\textwidth]{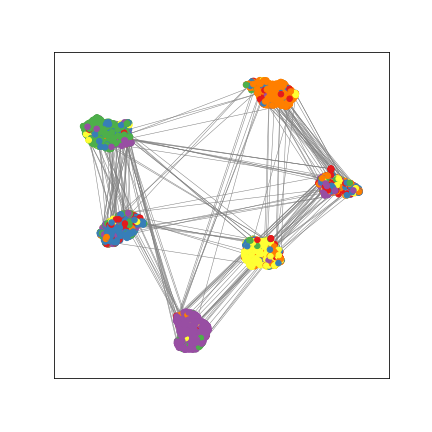}
        \vspace*{-0.8cm}
        \caption{\small TVGNN}
    \end{subfigure}
\caption{\small UMAP transforms of $\X^{(L)}$ for Citeseer. Colors in the top row of each dataset correspond to cluster assignments, while the colors in the bottom row correspond to true labels.}
\label{fig: umap_plot_supplementary_citeseer}
\end{figure*}

\begin{figure*}[!ht]
\centering
\scriptsize
    \begin{subfigure}{0.23\textwidth}
        \centering
        \includegraphics[width=\textwidth]{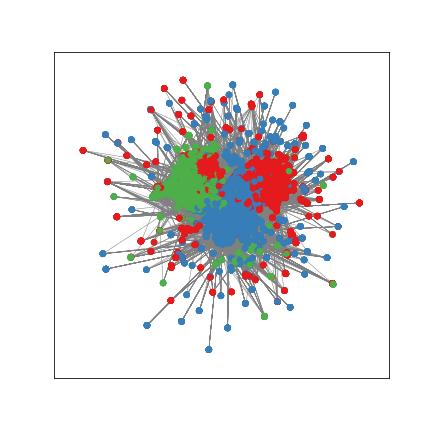}
        \vspace*{-0.8cm}
        \caption{\small DiffPool}
    \end{subfigure}
    \vspace*{-0.4cm}
    \begin{subfigure}{0.23\textwidth}
        \centering
        \includegraphics[width=\textwidth]{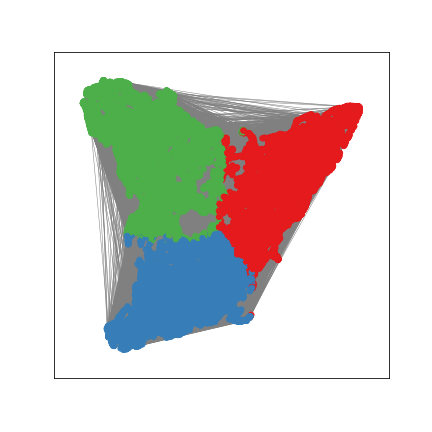}
        \vspace*{-0.8cm}
        \caption{\small MinCutPool}
    \end{subfigure}
    \begin{subfigure}{0.23\textwidth}
        \centering
        \includegraphics[width=\textwidth]{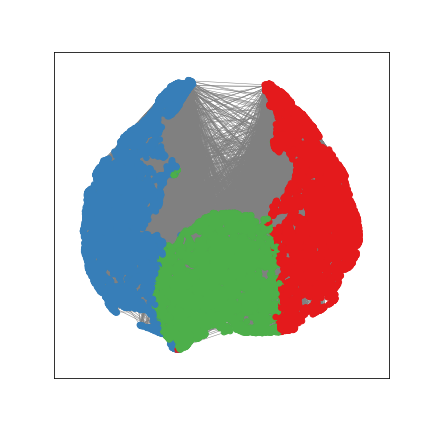}
        \vspace*{-0.8cm}
        \caption{\small DMoN}
    \end{subfigure}
    \begin{subfigure}{0.23\textwidth}
        \centering
        \includegraphics[width=\textwidth]{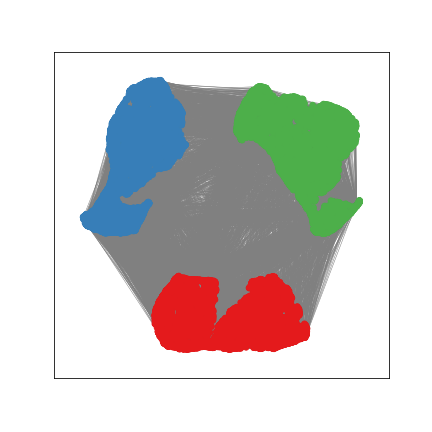}
        \vspace*{-0.8cm}
        \caption{\small TVGNN}
    \end{subfigure}
    \begin{subfigure}{0.23\textwidth}
        \centering
        \includegraphics[width=\textwidth]{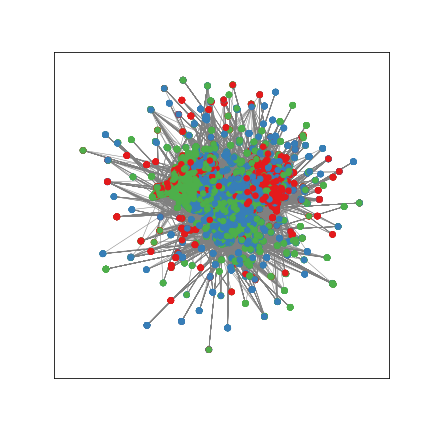}
        \vspace*{-0.8cm}
        \caption{\small DiffPool}
    \end{subfigure}
    \vspace*{-0.4cm}
    \begin{subfigure}{0.23\textwidth}
        \centering
        \includegraphics[width=\textwidth]{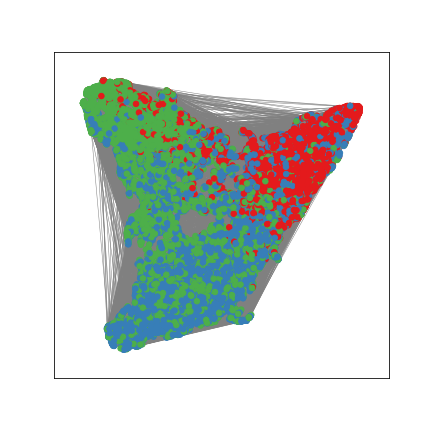}
        \vspace*{-0.8cm}
        \caption{\small MinCutPool}
    \end{subfigure}
    \begin{subfigure}{0.23\textwidth}
        \centering
        \includegraphics[width=\textwidth]{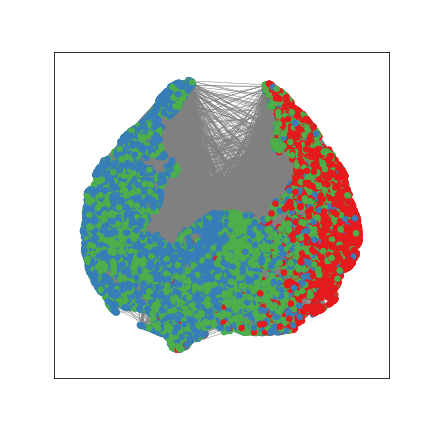}
        \vspace*{-0.8cm}
        \caption{\small DMoN}
    \end{subfigure}
    \begin{subfigure}{0.23\textwidth}
        \centering
        \includegraphics[width=\textwidth]{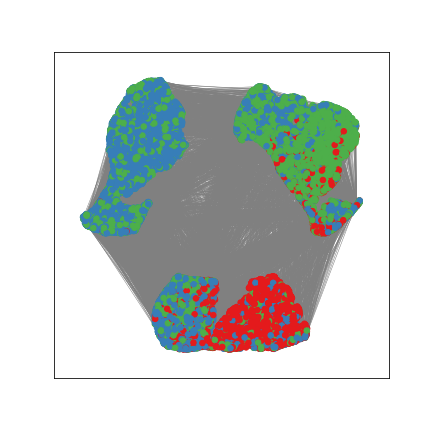}
        \vspace*{-0.8cm}
        \caption{\small TVGNN}
    \end{subfigure}
\caption{\small UMAP transforms of $\X^{(L)}$ for Pubmed. Colors in the top row of each dataset correspond to cluster assignments, while the colors in the bottom row correspond to true labels.}
\label{fig: umap_plot_supplementary_pubmed}
\end{figure*}

\begin{figure*}[!ht]
\centering
\scriptsize
    \begin{subfigure}{0.23\textwidth}
        \centering
        \includegraphics[width=\textwidth]{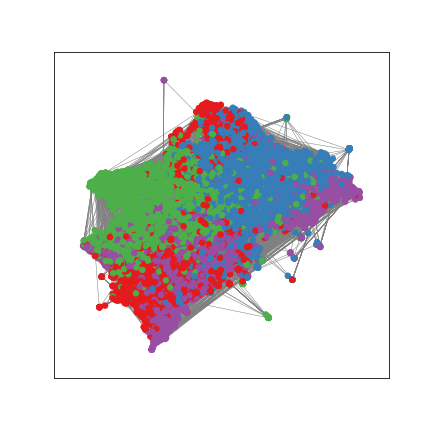}
        \vspace*{-0.8cm}
        \caption{\small DiffPool}
    \end{subfigure}
    \vspace*{-0.5cm}
    \begin{subfigure}{0.23\textwidth}
        \centering
        \includegraphics[width=\textwidth]{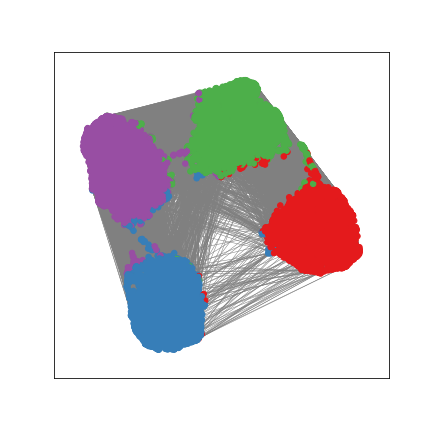}
        \vspace*{-0.8cm}
        \caption{\small MinCutPool}
    \end{subfigure}
    \begin{subfigure}{0.23\textwidth}
        \centering
        \includegraphics[width=\textwidth]{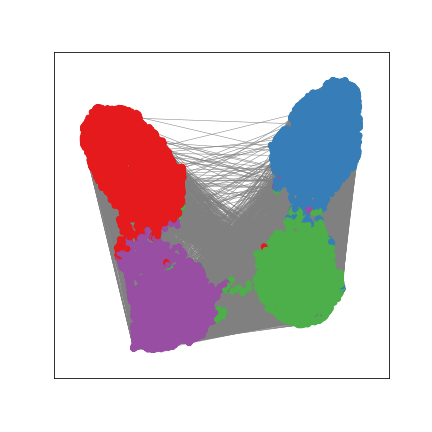}
        \vspace*{-0.8cm}
        \caption{\small DMoN}
    \end{subfigure}
    \begin{subfigure}{0.23\textwidth}
        \centering
        \includegraphics[width=\textwidth]{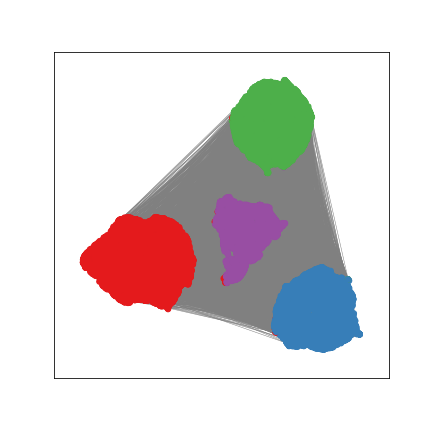}
        \vspace*{-0.8cm}
        \caption{\small TVGNN}
    \end{subfigure}
    \begin{subfigure}{0.23\textwidth}
        \centering
        \includegraphics[width=\textwidth]{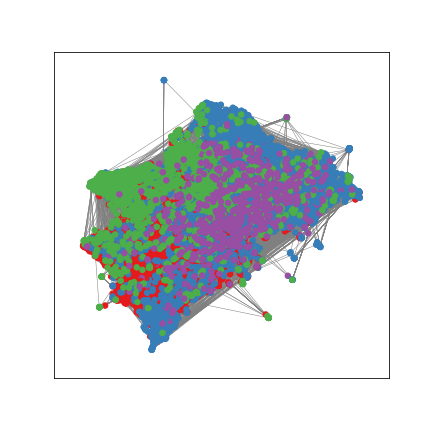}
        \vspace*{-0.8cm}
        \caption{\small DiffPool}
    \end{subfigure}
    \begin{subfigure}{0.23\textwidth}
        \centering
        \includegraphics[width=\textwidth]{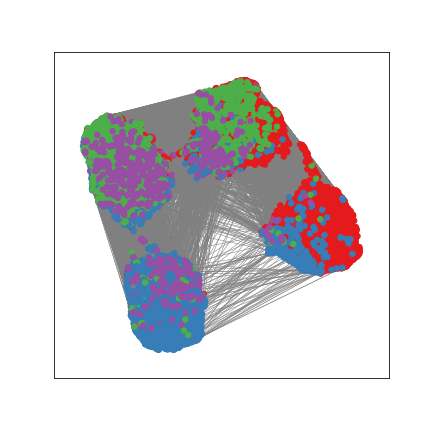}
        \vspace*{-0.8cm}
        \caption{\small MinCutPool}
    \end{subfigure}
    \begin{subfigure}{0.23\textwidth}
        \centering
        \includegraphics[width=\textwidth]{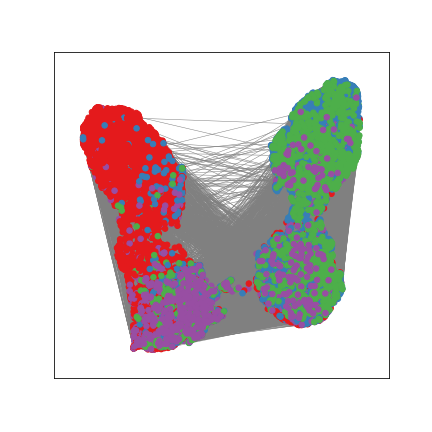}
        \vspace*{-0.8cm}
        \caption{\small DMoN}
    \end{subfigure}
    \begin{subfigure}{0.23\textwidth}
        \centering
        \includegraphics[width=\textwidth]{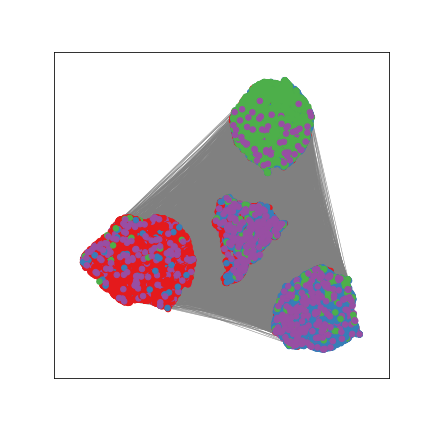}
        \vspace*{-0.8cm}
        \caption{\small TVGNN}
    \end{subfigure}
\caption{\small UMAP transforms of $\X^{(L)}$ for DBLP. Colors in the top row of each dataset correspond to cluster assignments, while the colors in the bottom row correspond to true labels.}
\label{fig: umap_plot_supplementary_dblp}
\end{figure*}

\begin{figure}[!ht]
    \centering
    \small
    \begin{subfigure}{0.24\textwidth}
        \centering
        \includegraphics[width=\textwidth]{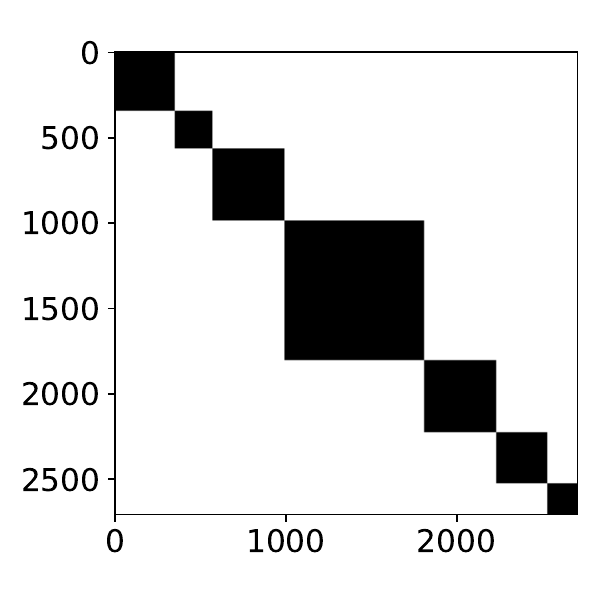}
        \vspace*{-0.7cm}
        \caption{\small Cora}
    \end{subfigure}
    \begin{subfigure}{0.24\textwidth}
        \centering
        \includegraphics[width=\textwidth]{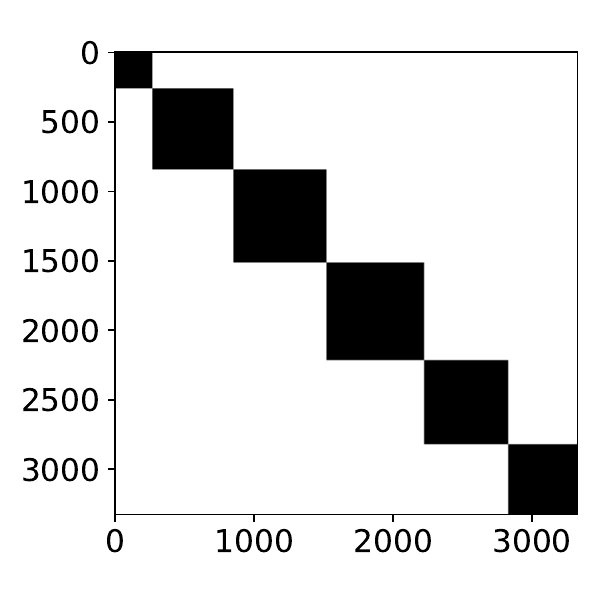}
        \vspace*{-0.7cm}
        \caption{\small Citeseer}
    \end{subfigure}
    \begin{subfigure}{0.24\textwidth}
        \centering
        \includegraphics[width=\textwidth]{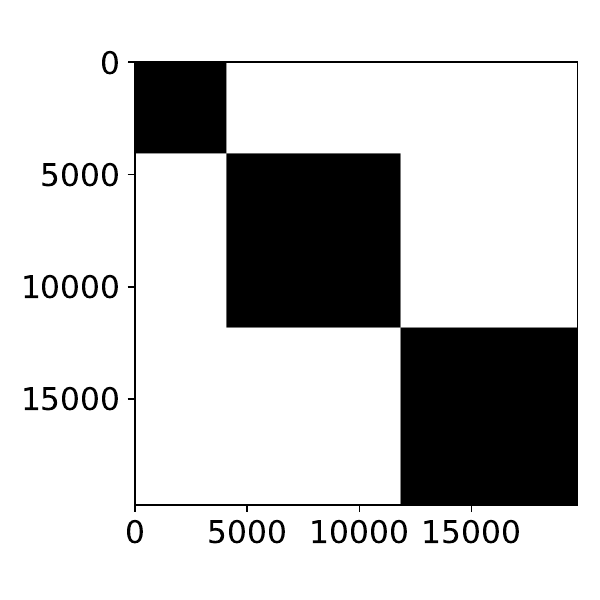}
        \vspace*{-0.7cm}
        \caption{\small Pubmed}
    \end{subfigure}
     \begin{subfigure}{0.24\textwidth}
        \centering
        \includegraphics[width=\textwidth]{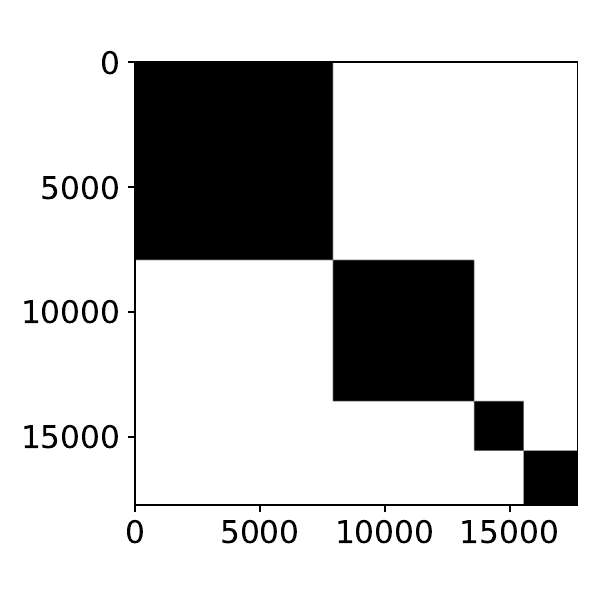}
        \vspace*{-0.7cm}
        \caption{\small DBLP}
    \end{subfigure}
    \caption{\small Logarithm of $\Sb\Sb^T$ for the true labels of each dataset.}
    \label{fig: SST_plot_labels_supplementary}
\end{figure}

\begin{figure}[!ht]
    \centering
    \small
    \begin{subfigure}{0.24\textwidth}
        \centering
        \includegraphics[width=\textwidth]{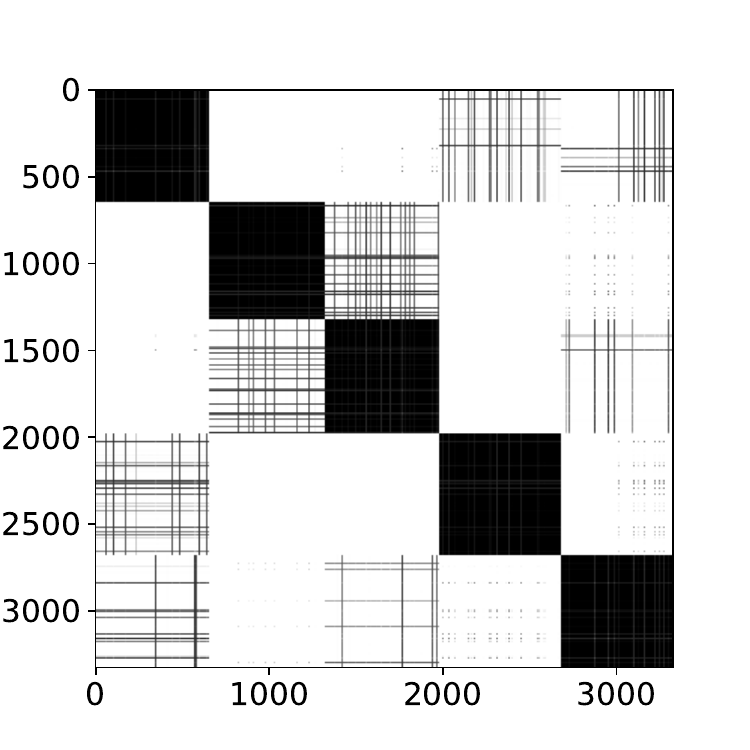}
        \vspace*{-0.7cm}
        \caption{\small GTVConv + MinCut loss}
    \end{subfigure}
    \begin{subfigure}{0.24\textwidth}
        \centering
        \includegraphics[width=\textwidth]{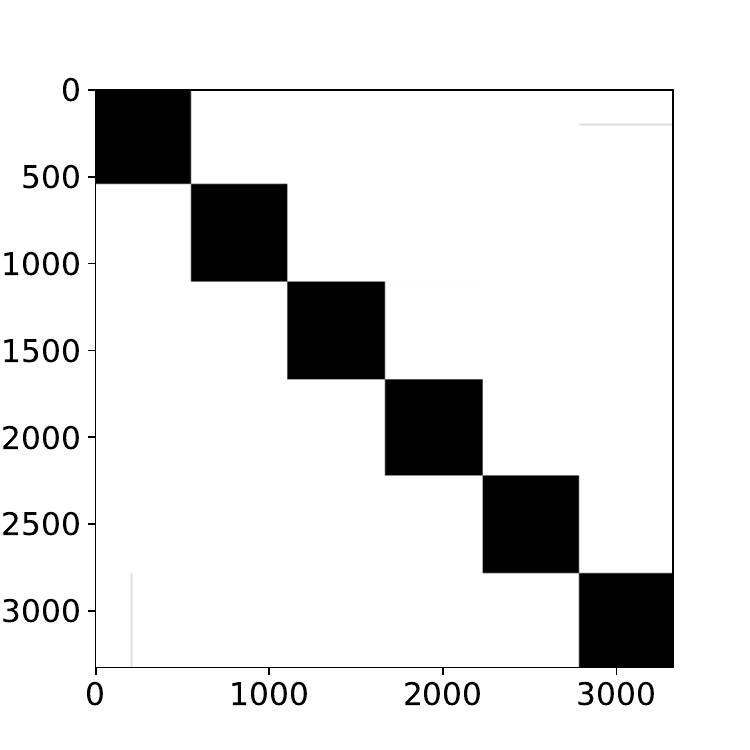}
        \vspace*{-0.7cm}
        \caption{\small GCN + TVGNN loss}
    \end{subfigure}
    \begin{subfigure}{0.24\textwidth}
        \centering
        \includegraphics[width=\textwidth]{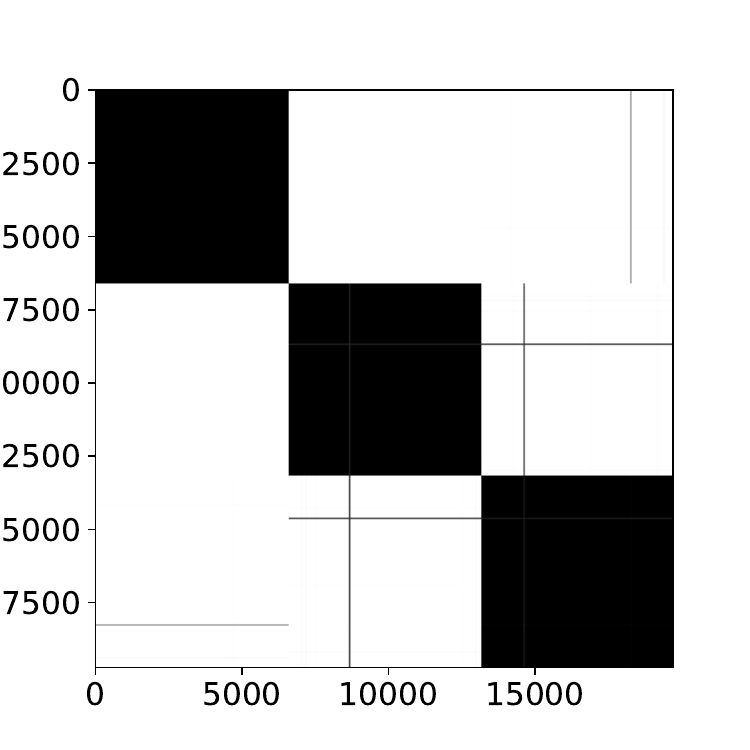}
        \vspace*{-0.7cm}
        \caption{\small GTVConv + MinCut loss}
    \end{subfigure}
    \begin{subfigure}{0.24\textwidth}
        \centering
        \includegraphics[width=\textwidth]{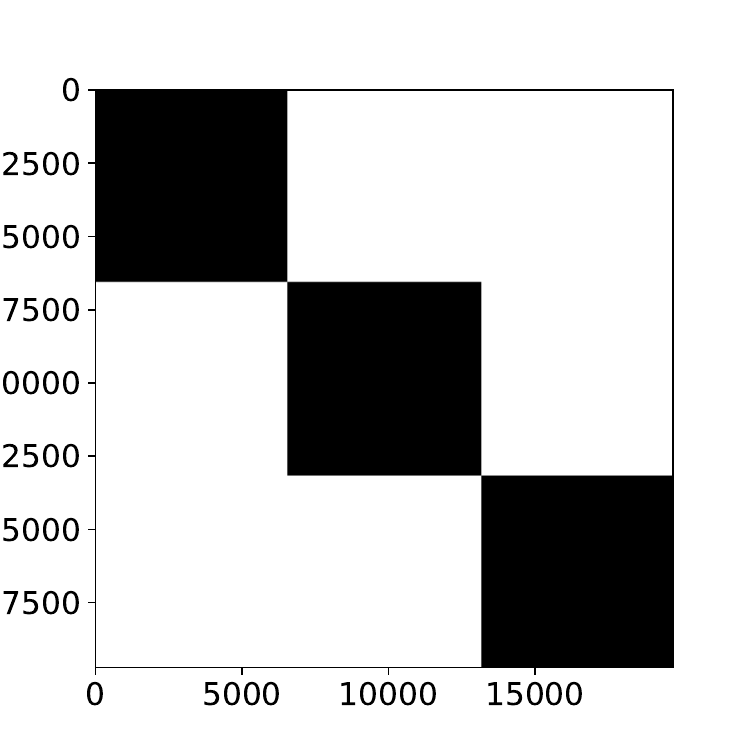}
        \vspace*{-0.7cm}
        \caption{\small GCN + TVGNN loss}
    \end{subfigure}
    \begin{subfigure}{0.24\textwidth}
        \centering
        \includegraphics[width=\textwidth]{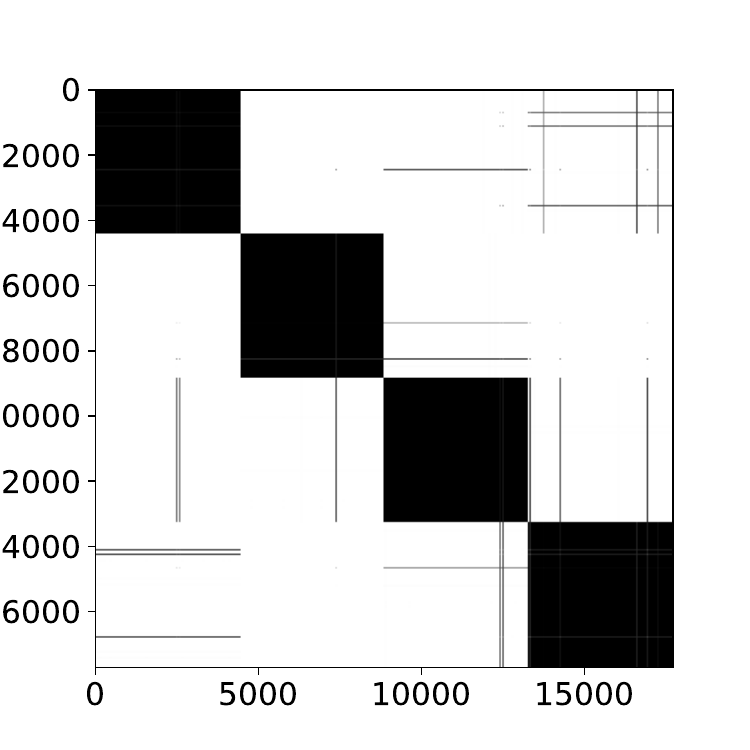}
        \vspace*{-0.7cm}
        \caption{\small GTVConv + MinCut loss}
    \end{subfigure}
    \begin{subfigure}{0.24\textwidth}
        \centering
        \includegraphics[width=\textwidth]{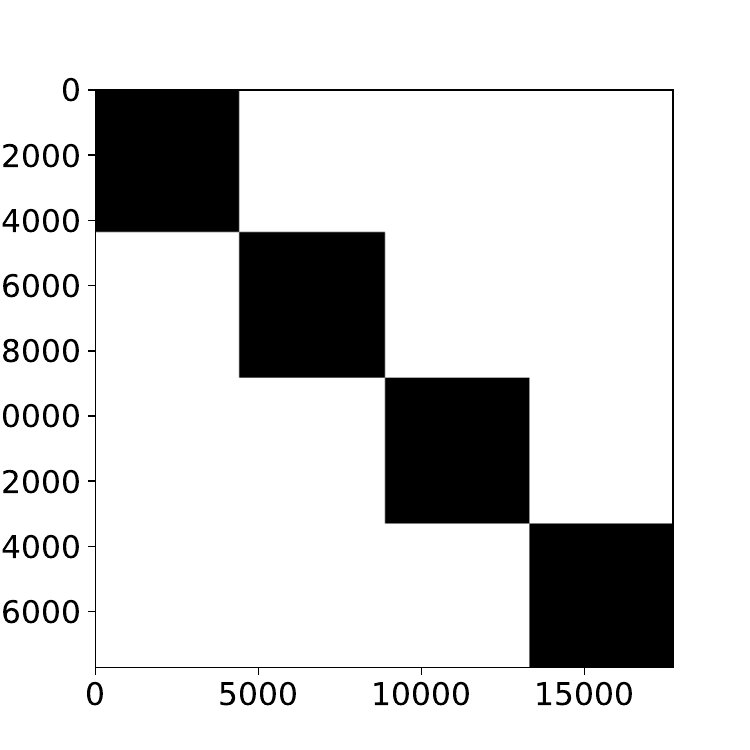}
        \vspace*{-0.7cm}
        \caption{\small GCN + TVGNN loss}
    \end{subfigure}
    \caption{\small Logarithm of $\Sb\Sb^T$ for the ablation study on Citeseer (a, b), Pubmed (c, d), and DBLP (e, f).}
    \label{fig: SST_plot_ablation}
\end{figure}

\begin{figure*}[!ht]
    \centering
    \scriptsize
    \begin{subfigure}{0.23\textwidth}
        \centering
        \includegraphics[width=\textwidth]{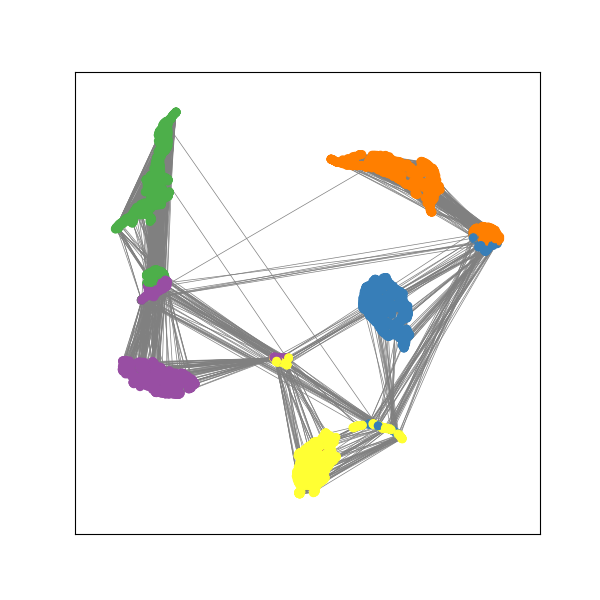}
        \vspace*{-0.8cm}
        \caption{\small GTVConv + MinCut loss}
    \end{subfigure}
    \vspace*{-0.1cm}
    \begin{subfigure}{0.23\textwidth}
        \centering
        \includegraphics[width=\textwidth]{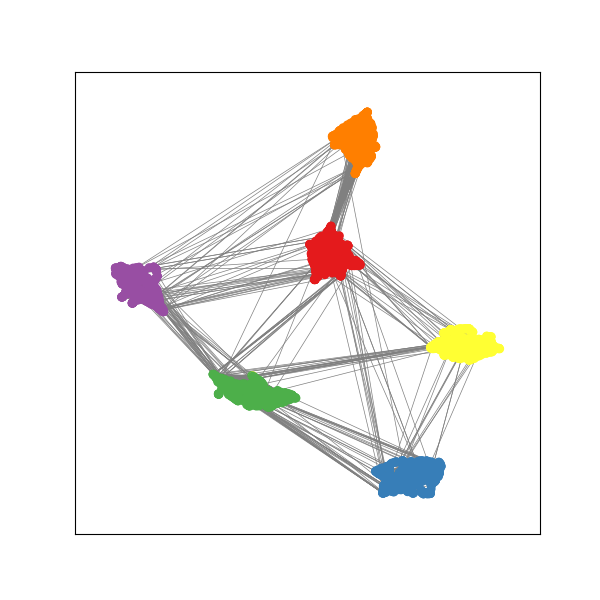}
        \vspace*{-0.8cm}
        \caption{\small GCN + TVGNN loss}
    \end{subfigure}
    \begin{subfigure}{0.23\textwidth}
        \centering
        \includegraphics[width=\textwidth]{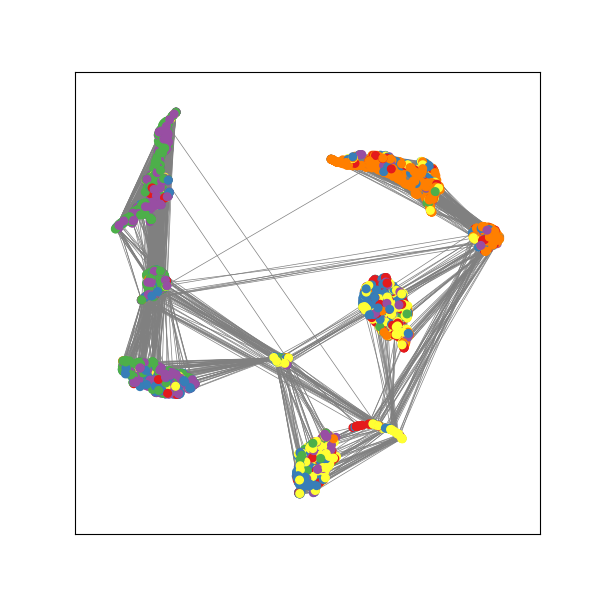}
        \vspace*{-0.8cm}
        \caption{\small GTVConv + MinCut loss}
    \end{subfigure}
    \begin{subfigure}{0.23\textwidth}
        \centering
        \includegraphics[width=\textwidth]{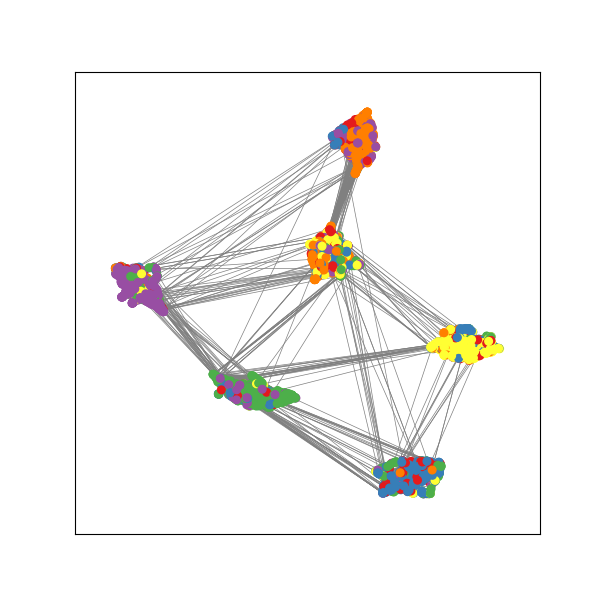}
        \vspace*{-0.8cm}
        \caption{\small GCN + TVGNN loss}
    \end{subfigure}
    \caption{\small UMAP plots for the ablation study on Citeseer.}
   \label{fig: umap_plot_ablation_citeseer}
\end{figure*}

\begin{figure*}[!ht]
    \centering
    \scriptsize
    \begin{subfigure}{0.23\textwidth}
        \centering
        \includegraphics[width=\textwidth]{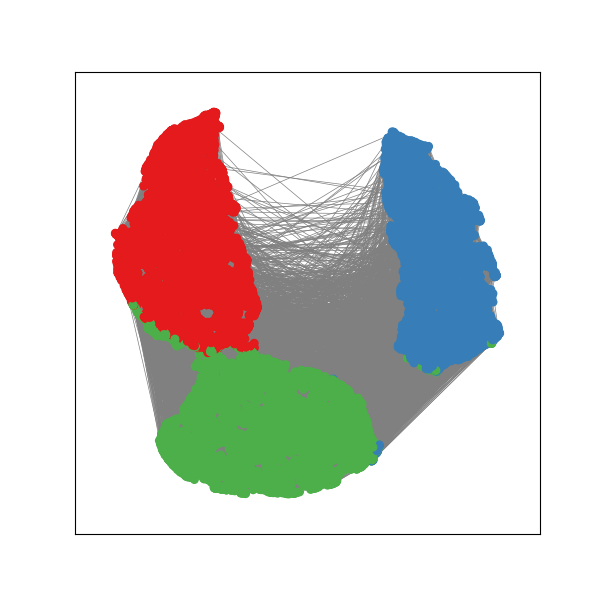}
        \vspace*{-0.8cm}
        \caption{\small GTVConv + MinCut loss}
    \end{subfigure}
    \begin{subfigure}{0.23\textwidth}
        \centering
        \includegraphics[width=\textwidth]{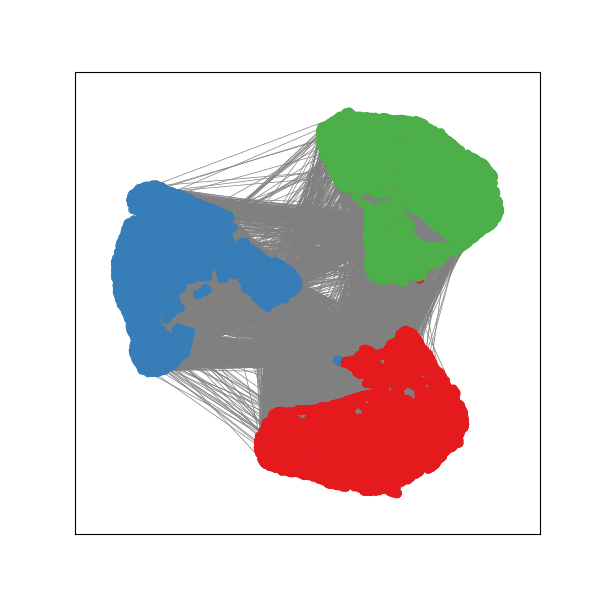}
        \vspace*{-0.8cm}
        \caption{\small GCN + TVGNN loss}
    \end{subfigure}
    \begin{subfigure}{0.23\textwidth}
        \centering
        \includegraphics[width=\textwidth]{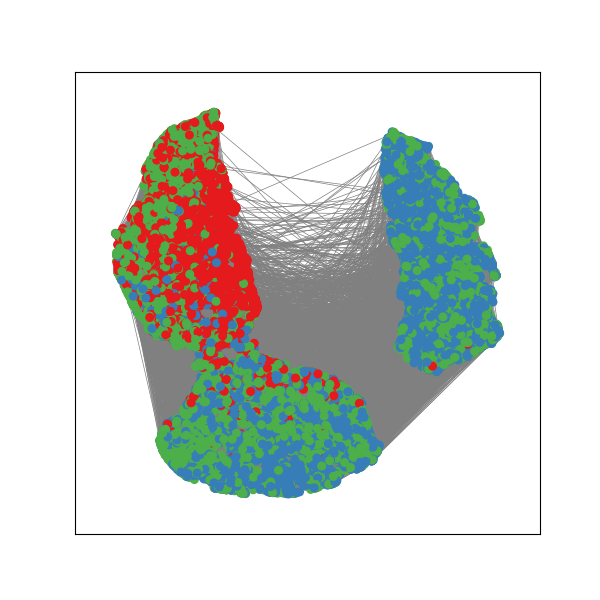}
        \vspace*{-0.8cm}
        \caption{\small GTVConv + MinCut loss}
    \end{subfigure}
    \begin{subfigure}{0.23\textwidth}
        \centering
        \includegraphics[width=\textwidth]{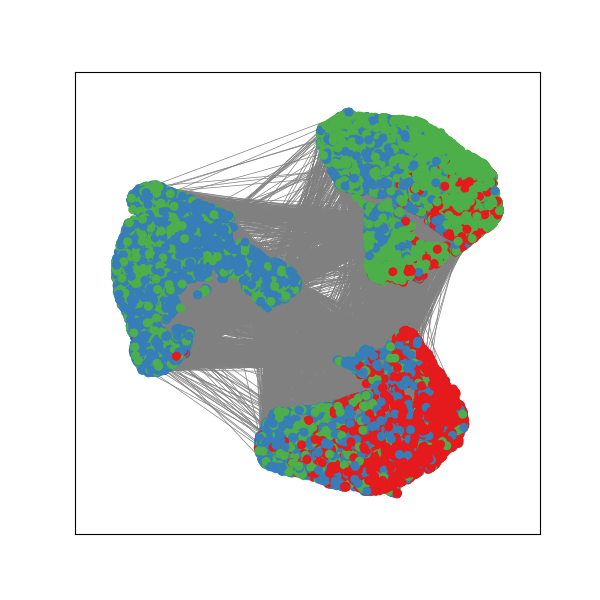}
        \vspace*{-0.8cm}
        \caption{\small GCN + TVGNN loss}
    \end{subfigure}
    \\
    \caption{\small UMAP plots for the ablation study on Pubmed.}
   \label{fig: umap_plot_ablation_pubmed}
\end{figure*}

\begin{figure*}[!ht]
    \centering
    \scriptsize
    \begin{subfigure}{0.23\textwidth}
        \centering
        \includegraphics[width=\textwidth]{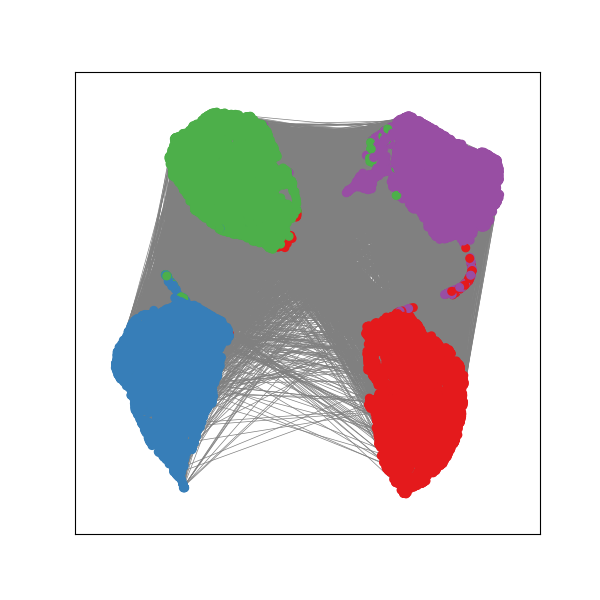}
        \vspace*{-0.8cm}
        \caption{\small GTVConv + MinCut loss}
    \end{subfigure}
    \begin{subfigure}{0.23\textwidth}
        \centering
        \includegraphics[width=\textwidth]{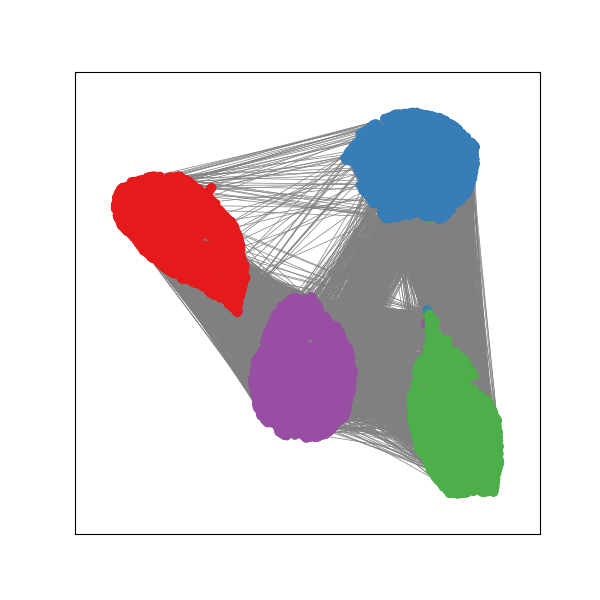}
        \vspace*{-0.8cm}
        \caption{\small GCN + TVGNN loss}
    \end{subfigure}
    \begin{subfigure}{0.23\textwidth}
        \centering
        \includegraphics[width=\textwidth]{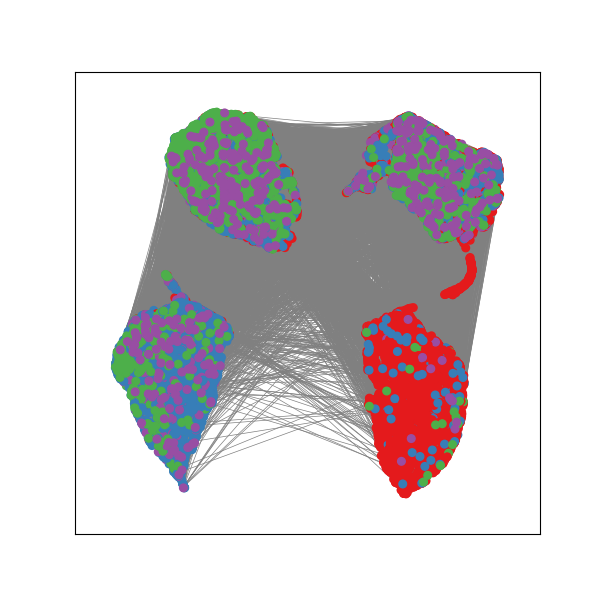}
        \vspace*{-0.8cm}
        \caption{\small GTVConv + MinCut loss}
    \end{subfigure}
    \begin{subfigure}{0.23\textwidth}
        \centering
        \includegraphics[width=\textwidth]{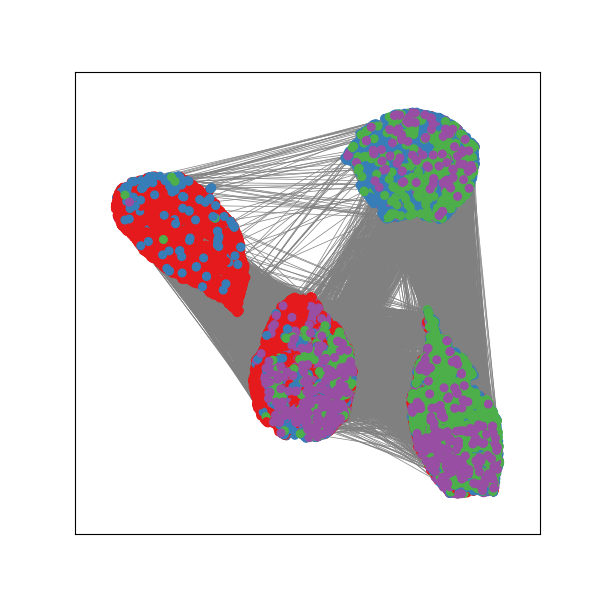}
        \vspace*{-0.8cm}
        \caption{\small GCN + TVGNN loss}
    \end{subfigure}
    \caption{\small UMAP plots for the ablation study on DBLP.}
   \label{fig: umap_plot_ablation_dblp}
\end{figure*}

Plots with the logarithm of $\Sb\Sb^T$ for Citeseer, Pubmed, and DBLP are presented in Fig.~\ref{fig: SST_plot_supplementary_citeseer}, Fig.~\ref{fig: SST_plot_supplementary_pubmed}, and Fig.~\ref{fig: SST_plot_supplementary_dblp} respectively.
The UMAP transform of $\X^{(L)}$ for Citeseer, Pubmed, and DBLP are presented in Fig.~\ref{fig: umap_plot_supplementary_citeseer}, Fig.~\ref{fig: umap_plot_supplementary_pubmed}, and Fig.~\ref{fig: umap_plot_supplementary_dblp} respectively. 

As for the case of Cora, TVGNN manages to give better-separated clusters with sharper assignments for all three graphs when compared to the other three GNN-based clustering methods that produce soft assignments. These plots also show that the cluster distribution given by TVGNN is not always balanced, see for instance Fig.~\ref{fig: SST_plot_supplementary_dblp}. In fact, with respect to the true labels, all four datasets are imbalanced, which can be seen from Fig.~\ref{fig: SST_plot_labels_supplementary}. 

The plots of $\Sb\Sb^T$ for the configurations used in the ablation study are presented in Fig.~\ref{fig: SST_plot_ablation}, while the UMAP plots for Citeseer, Pubmed, and DBLP are presented in Fig.~\ref{fig: umap_plot_ablation_citeseer}, Fig.~\ref{fig: umap_plot_ablation_pubmed}, and Fig.~\ref{fig: umap_plot_ablation_dblp}, respectively. 

\subsection{Denoising task}
\label{appendix:denoising}

\begin{figure*}[!ht]
    \centering
    \scriptsize
    \vspace*{-0.4cm}
    \begin{subfigure}{0.25\textwidth}
        \centering
        \includegraphics[width=\textwidth]{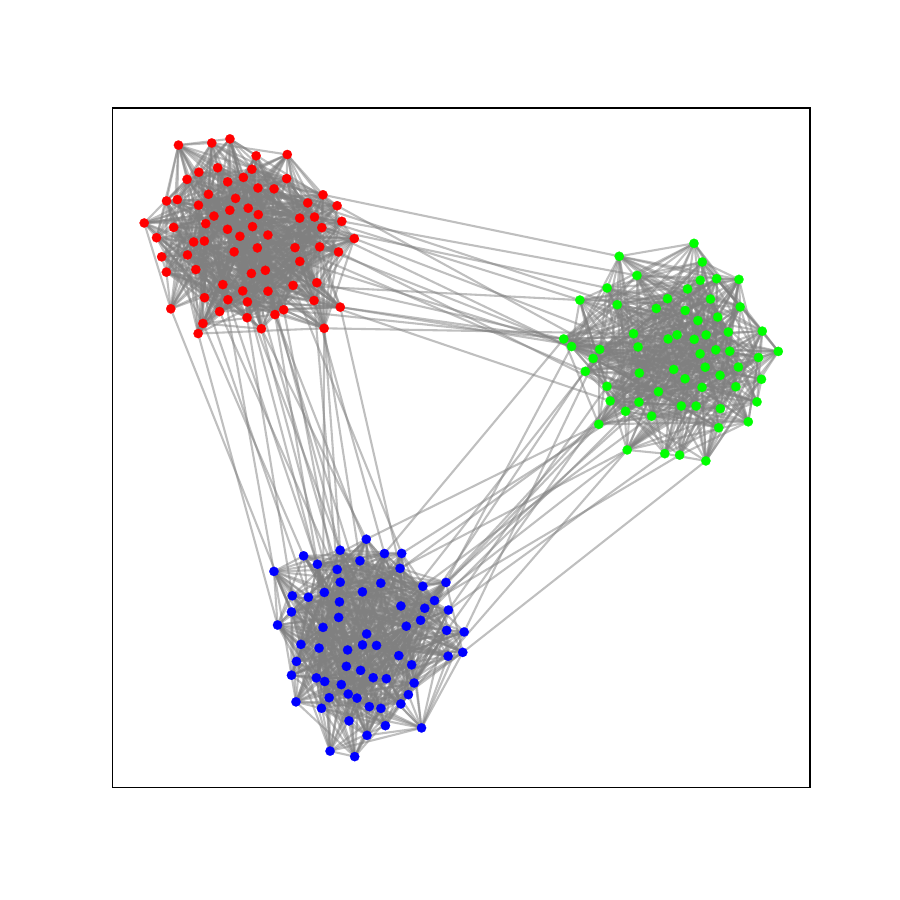}
        \vspace*{-0.8cm}
        \caption{\small Original}
    \end{subfigure}
    \vspace*{-0.4cm}
    \begin{subfigure}{0.25\textwidth}
        \centering
        \includegraphics[width=\textwidth]{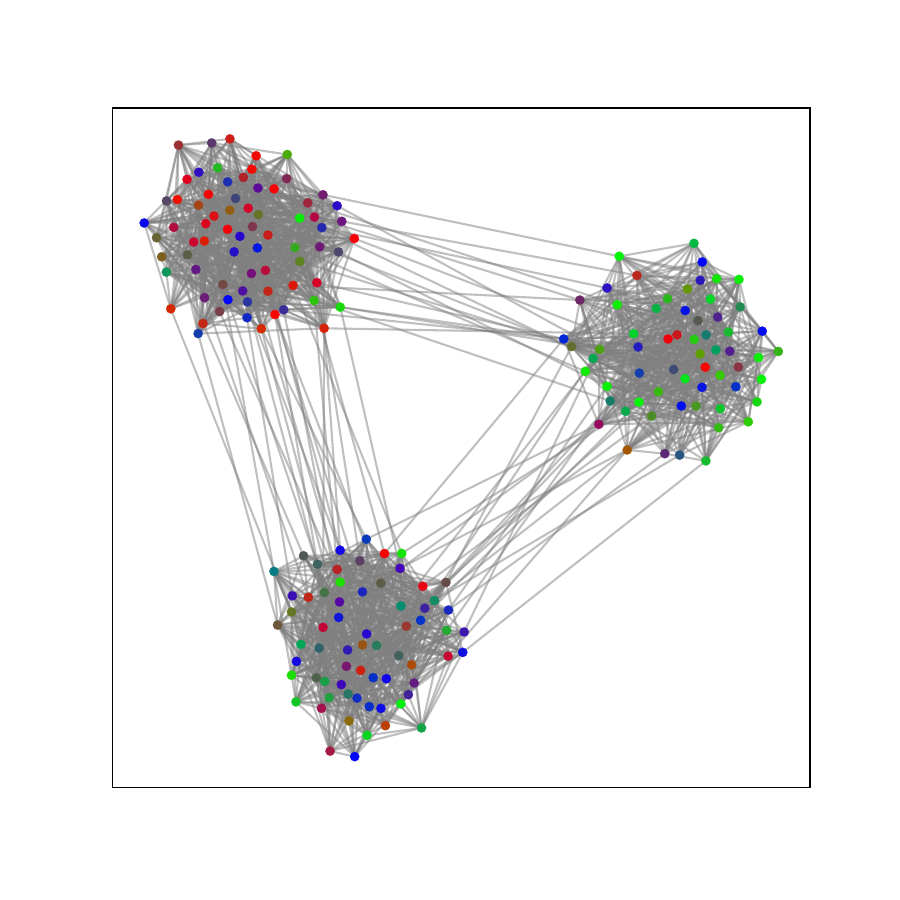}
        \vspace*{-0.8cm}
        \caption{\small Original + Noise}
    \end{subfigure}
    \begin{subfigure}{0.25\textwidth}
        \centering
        \includegraphics[width=\textwidth]{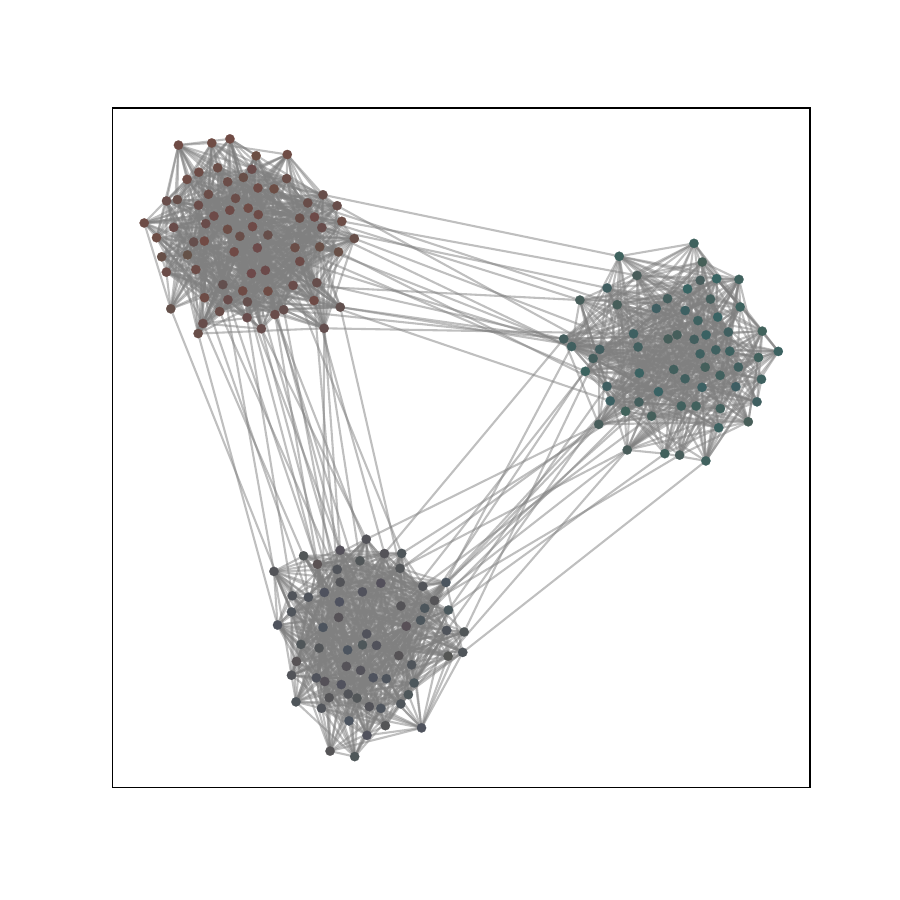}
        \vspace*{-0.8cm}
        \caption{\small Diffpool}
    \end{subfigure}
    \\
    \begin{subfigure}{0.25\textwidth}
        \centering
        \includegraphics[width=\textwidth]{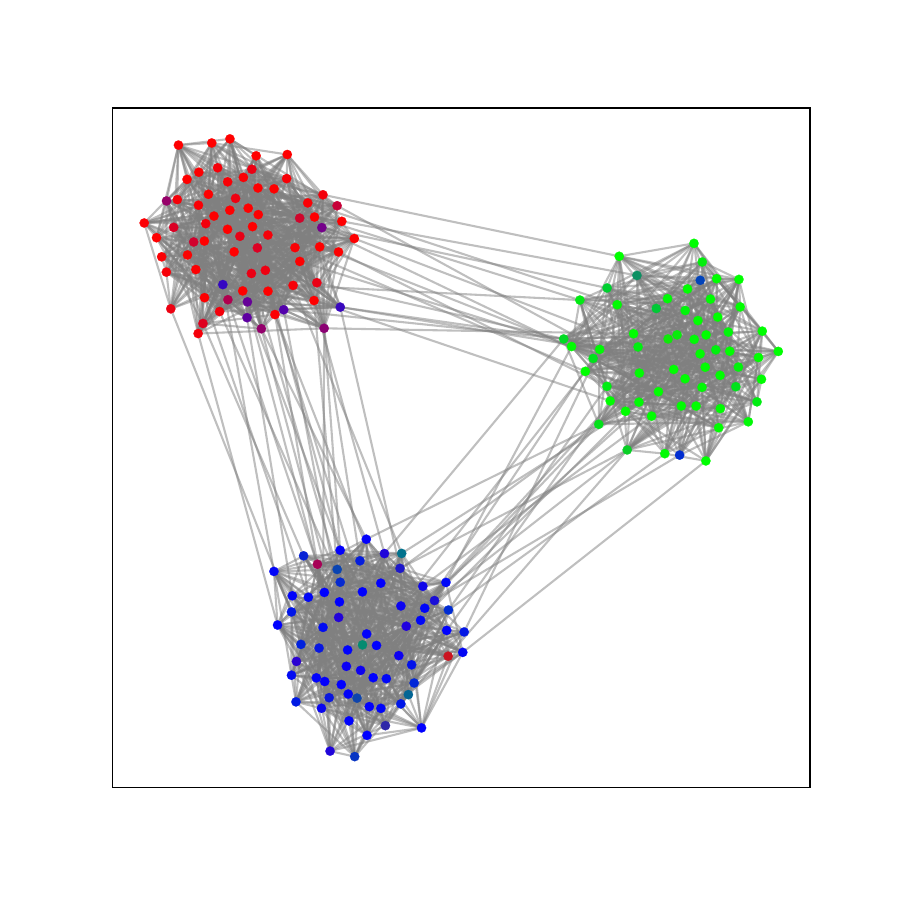}
        \vspace*{-0.8cm}
        \caption{\small MinCutPool}
    \end{subfigure}
    \begin{subfigure}{0.25\textwidth}
        \centering
        \includegraphics[width=\textwidth]{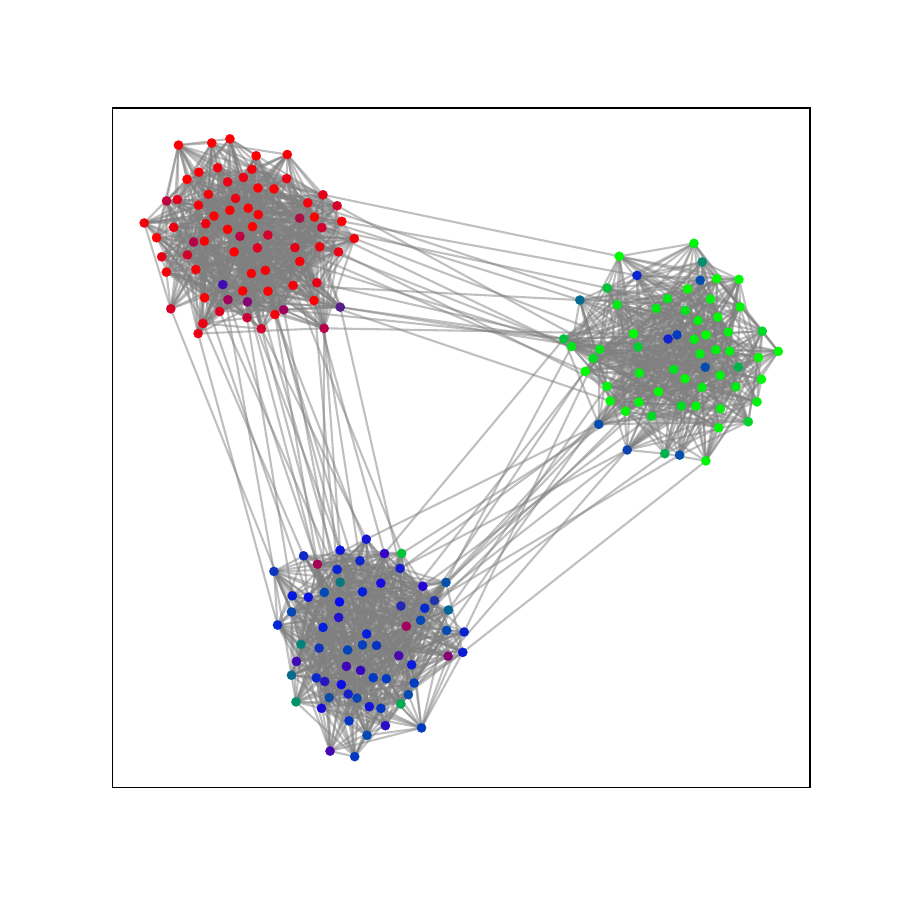}
        \vspace*{-0.8cm}
        \caption{\small DMoN}
    \end{subfigure}
    \begin{subfigure}{0.25\textwidth}
        \centering
        \includegraphics[width=\textwidth]{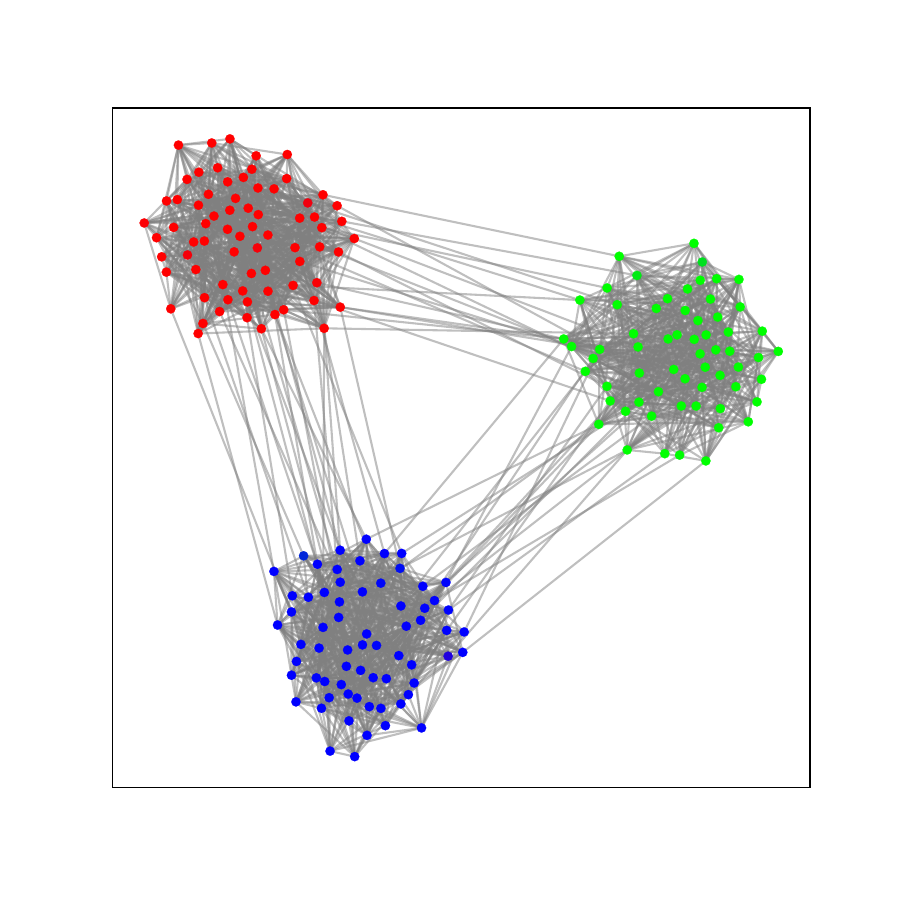}
        \vspace*{-0.8cm}
        \caption{\small TVGNN}
    \end{subfigure}
    \caption{\small Denoising task. (a) the original vertex features; (b) the vertex features corrupted with Gaussian noise; (c-f) cluster labels identified by each method.}
   \label{fig: denoising_task}
\end{figure*}

\begin{figure*}[!ht]
    \centering
    \scriptsize
    \vspace*{-0.4cm}
    \begin{subfigure}{0.2\textwidth}
        \centering
        \includegraphics[width=\textwidth]{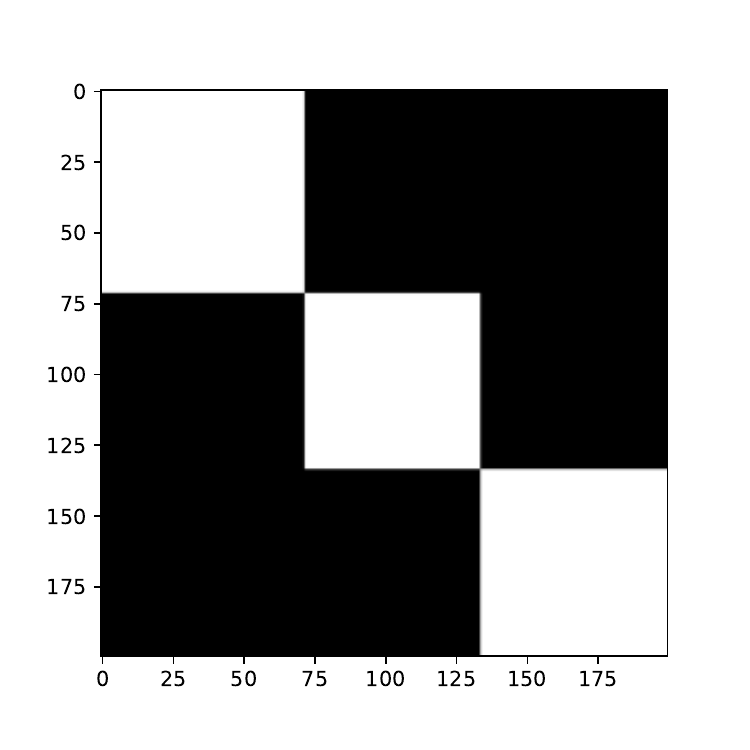}
        \vspace*{-0.7cm}
        \caption{\small Original}
    \end{subfigure}
    \vspace*{-0.4cm}
    \begin{subfigure}{0.2\textwidth}
        \centering
        \includegraphics[width=\textwidth]{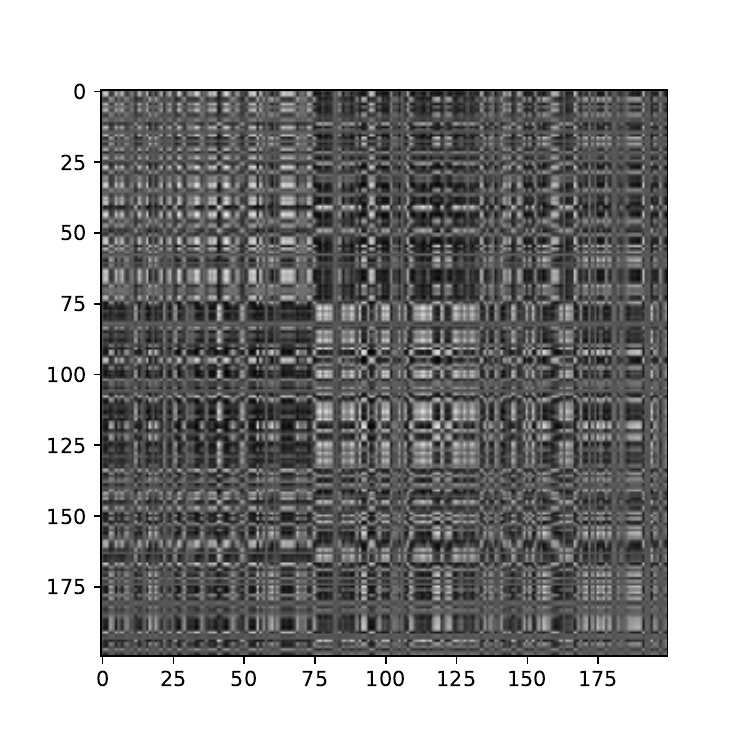}
        \vspace*{-0.7cm}
        \caption{\small Original + Noise}
    \end{subfigure}
    \begin{subfigure}{0.2\textwidth}
        \centering
        \includegraphics[width=\textwidth]{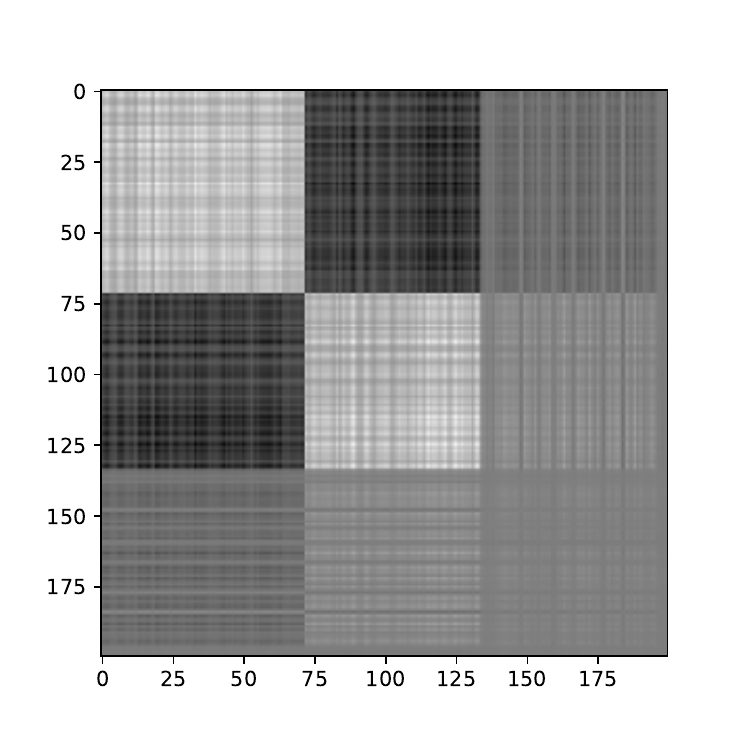}
        \vspace*{-0.7cm}
        \caption{\small Diffpool}
    \end{subfigure}
    \\
    \begin{subfigure}{0.2\textwidth}
        \centering
        \includegraphics[width=\textwidth]{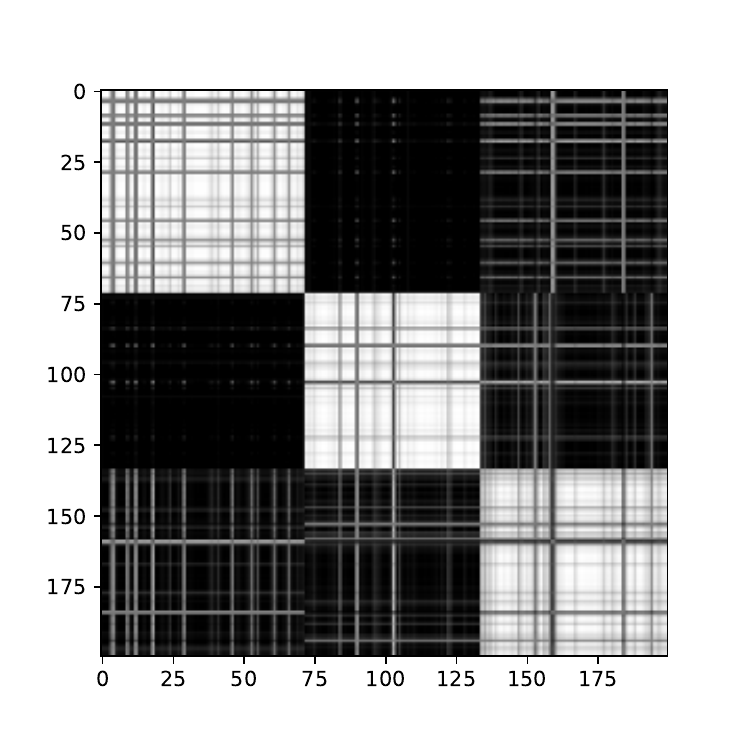}
        \vspace*{-0.7cm}
        \caption{\small MinCutPool}
    \end{subfigure}
    \begin{subfigure}{0.2\textwidth}
        \centering
        \includegraphics[width=\textwidth]{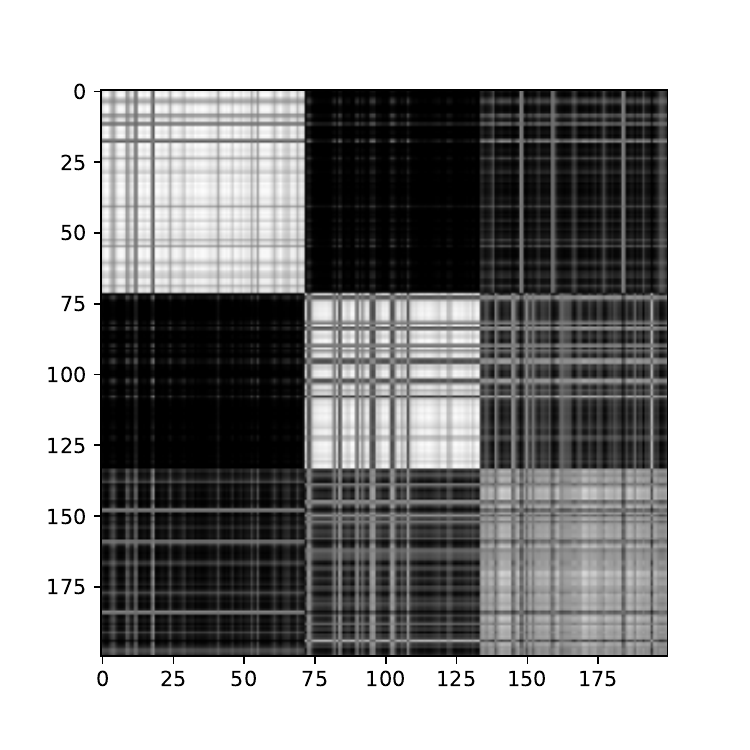}
        \vspace*{-0.7cm}
        \caption{\small DMoN}
    \end{subfigure}
    \begin{subfigure}{0.2\textwidth}
        \centering
        \includegraphics[width=\textwidth]{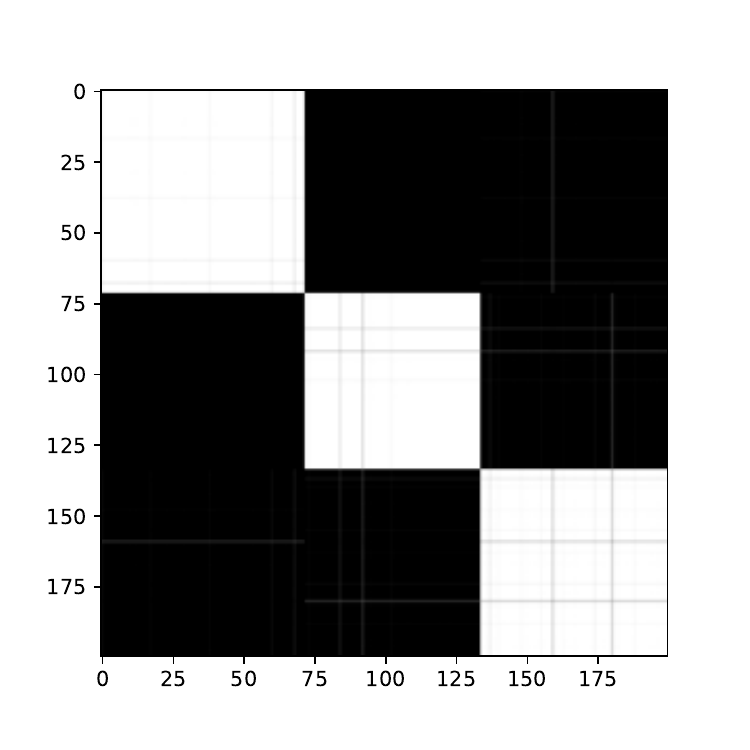}
        \vspace*{-0.7cm}
        \caption{\small TVGNN}
    \end{subfigure}
    \caption{\small Visualization of $\Sb\Sb^T$ for the denoising task. (a) the original vertex features; (b) the vertex features corrupted with Gaussian noise; (c-f) cluster labels identified by each method.}
   \label{fig: denoising_task_ss}
\end{figure*}

For this task, we generated a Stochastic-Block Model graph with three communities. 
The graph has 200 nodes, the probability of having a within-community edge is set to 0.3, and the probability of having an edge between communities is 0.005.
We assigned vertex features $\boldsymbol{x}_1 = [1,0,0]$, $\boldsymbol{x}_2 = [0,1,0]$, and $\boldsymbol{x}_3 = [0,0,1]$ to the vertices of the first, second, and third community, respectively.
Having three communities, allows us to use a convenient RGB color coding to visualize the vertex features.
Next, we corrupted the features by adding Gaussian noise from $\mathcal{N}(0, 1.5)$. 
Afterward, we performed vertex clustering with Diffpool, MinCut, DMoN, and TVGNN. 
The results are shown in Fig.~\ref{fig: denoising_task}.
Note that in panels (a-b), colors indicate node features $\X$, while in (c-f) the colors indicate the cluster assignments $\Sb$.
As we can see, TVGNN manages to perfectly recover the original clusters, while the other methods do not. 
As for the other experiments, in Fig.~\ref{fig: denoising_task_ss} we visualize the matrix $\Sb\Sb^T$ which shows that TVGNN generates cluster assignments that are much sharper than those produced by the other GNN-based clustering methods.

\subsection{Clustering two simple point clouds}
\label{appendix:point_clouds}

\begin{figure}[!ht]
    \centering
    \begin{subfigure}{0.23\textwidth}
        \centering
        \includegraphics[width=\textwidth]{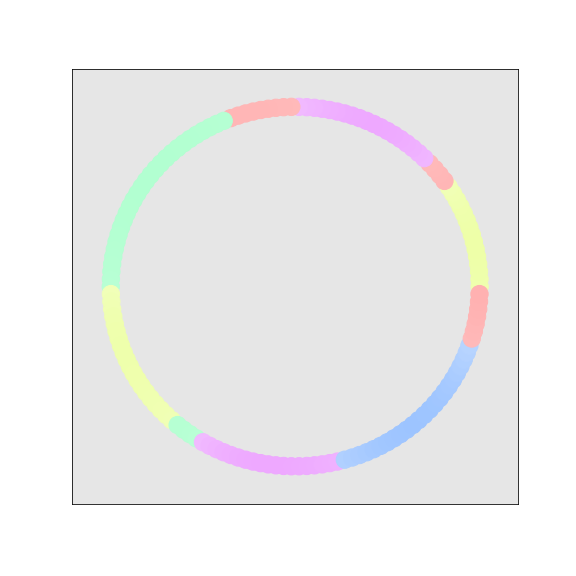}
        \vspace*{-0.8cm}
        \caption{DiffPool}
    \end{subfigure}
    \begin{subfigure}{0.23\textwidth}
        \centering
        \includegraphics[width=\textwidth]{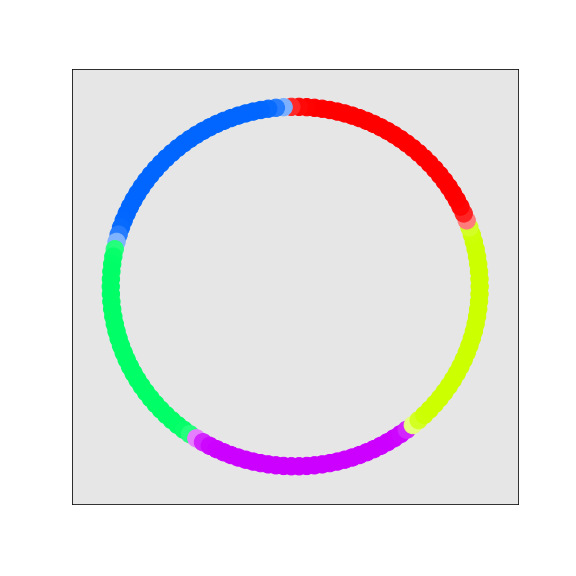}
        \vspace*{-0.8cm}
        \caption{MinCutPool}
    \end{subfigure}
    \begin{subfigure}{0.23\textwidth}
        \centering
        \includegraphics[width=\textwidth]{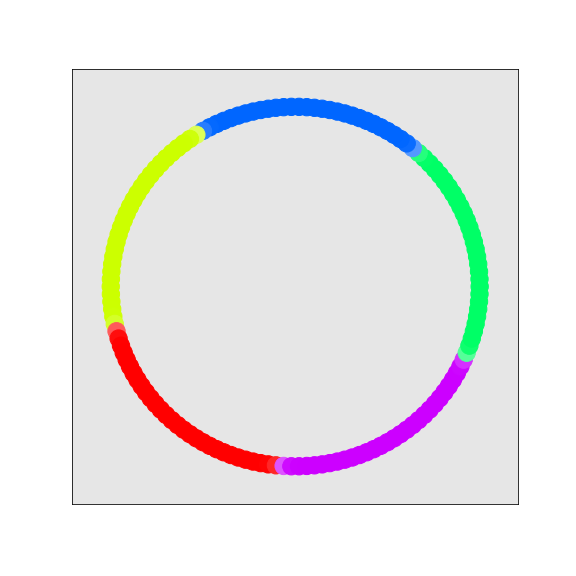}
        \vspace*{-0.8cm}
        \caption{DMoN}
    \end{subfigure}
    \begin{subfigure}{0.23\textwidth}
        \centering
        \includegraphics[width=\textwidth]{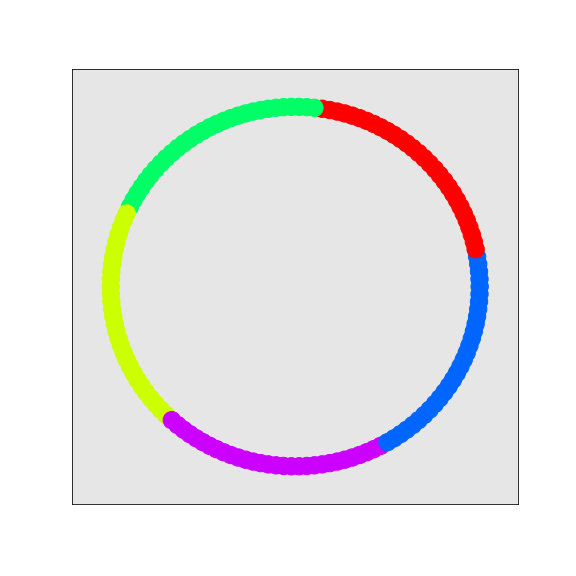}
        \vspace*{-0.8cm}
        \caption{TVGNN}
    \end{subfigure}
    \caption{Cluster assignments for the ring graph. The colors correspond to the index of the largest value in the soft cluster assignment vector. The brightness is proportional to the highest value in the assignment vector.}
    \label{fig: ring_graph_plot}
\end{figure}

\begin{figure}[!ht]
    \centering
    \begin{subfigure}{0.23\textwidth}
        \centering
        \includegraphics[width=\textwidth]{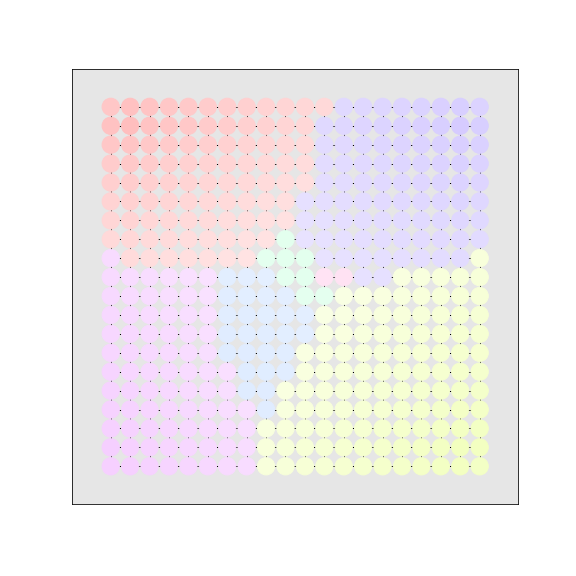}
        \vspace*{-0.8cm}
        \caption{DiffPool}
    \end{subfigure}
    \begin{subfigure}{0.23\textwidth}
        \centering
        \includegraphics[width=\textwidth]{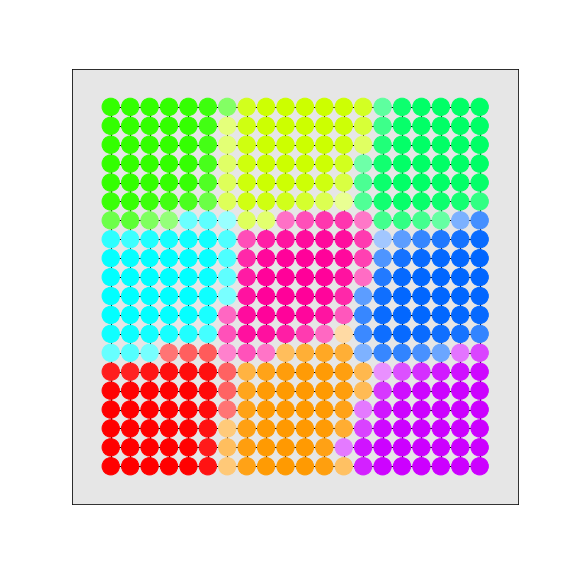}
        \vspace*{-0.8cm}
        \caption{MinCutPool}
    \end{subfigure}
    \begin{subfigure}{0.23\textwidth}
        \centering
        \includegraphics[width=\textwidth]{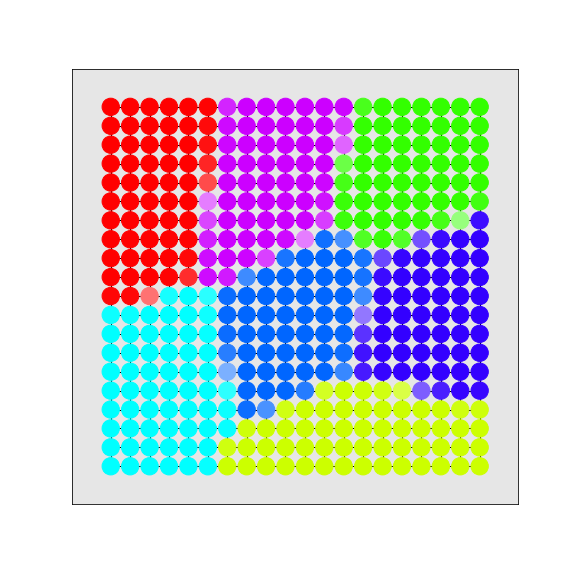}
        \vspace*{-0.8cm}
        \caption{DMoN}
    \end{subfigure}
    \begin{subfigure}{0.23\textwidth}
        \centering
        \includegraphics[width=\textwidth]{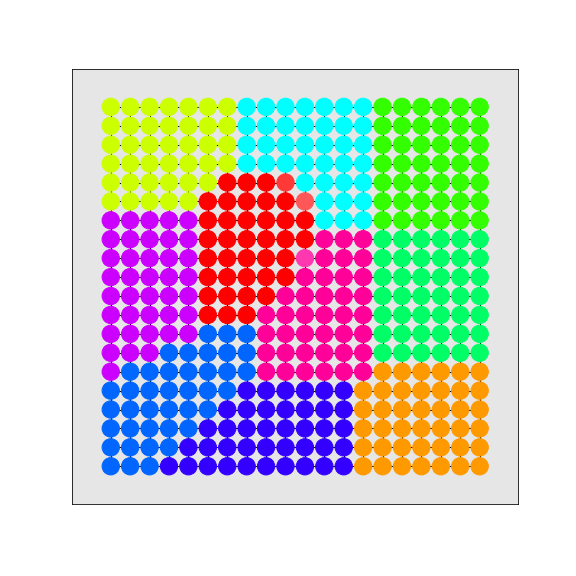}
        \vspace*{-0.8cm}
        \caption{TVGNN}
    \end{subfigure}
    \caption{Cluster assignments for the grid graph. The colors correspond to the index of the largest value in the soft cluster assignment vector. The brightness is proportional to the highest value in the assignment vector.}
    \label{fig: grid_2d_graph_plot}
\end{figure}

\begin{figure}[!ht]
    \centering
    \begin{subfigure}{0.49\textwidth}
        \centering
        \includegraphics[width=\textwidth]{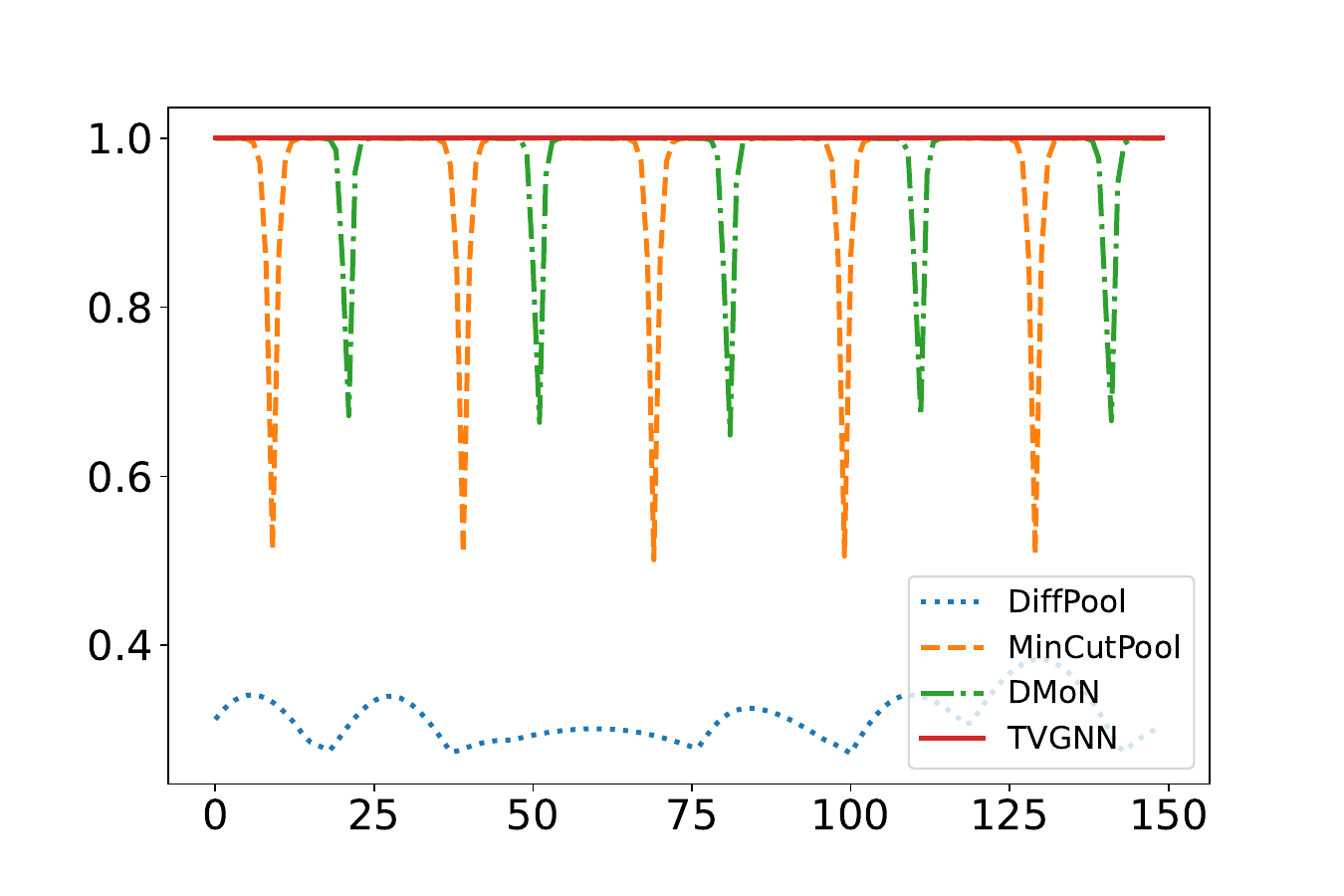}
        \vspace*{-0.7cm}
        \caption{Ring}
    \end{subfigure}
    \begin{subfigure}{0.49\textwidth}
        \centering
        \includegraphics[width=\textwidth]{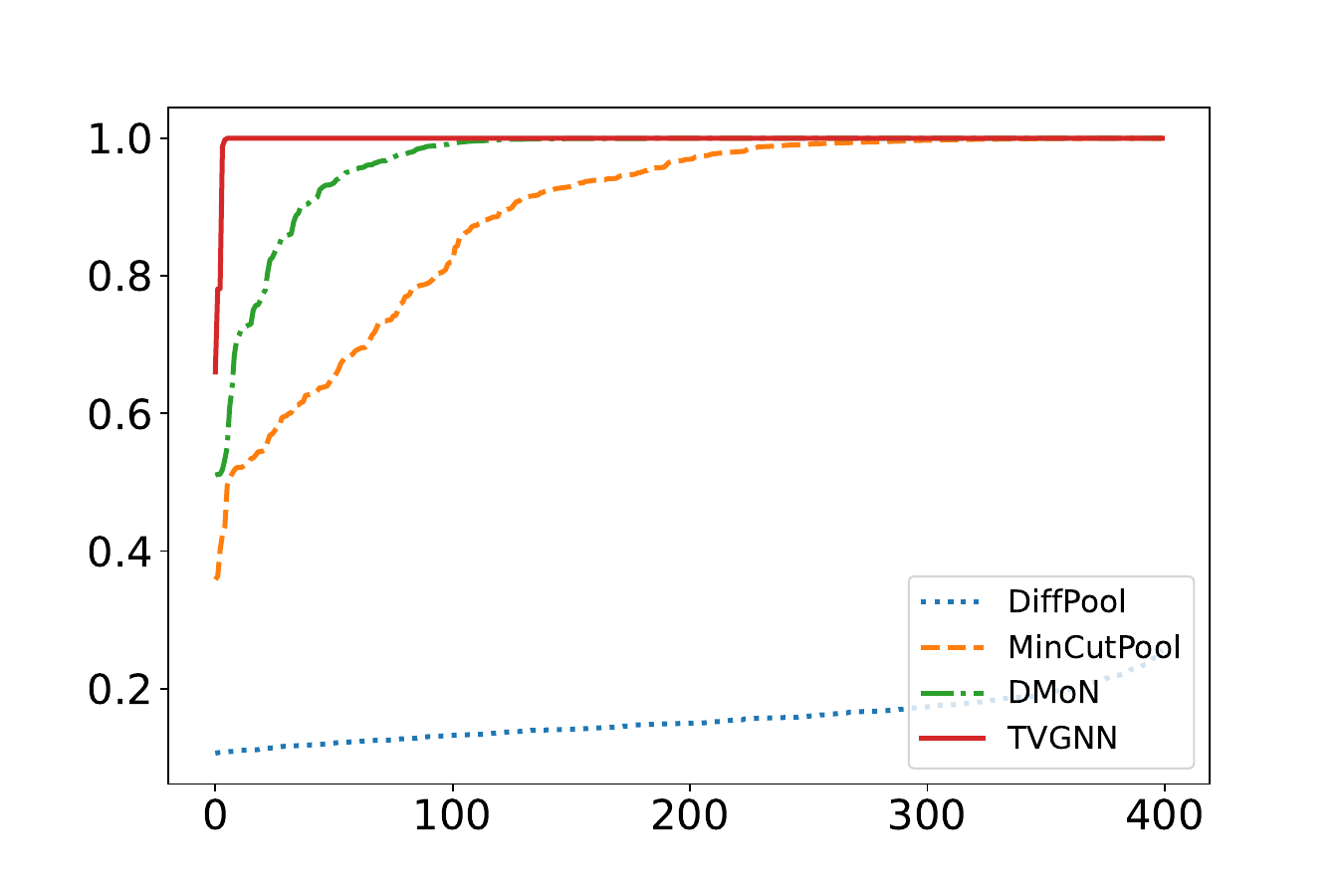}
        \vspace*{-0.7cm}
        \caption{Grid}
    \end{subfigure}
    \caption{Largest value in the soft cluster assignment vector as a function of the vertex index. For the ring in (a), the horizontal axis moves along the circumference of the ring. For the grid in (b), the node indices are sorted according to the largest value in the assignment vectors.}
    \label{fig: largest_soft_assignment}
\end{figure}

Fig.~\ref{fig: ring_graph_plot} and \ref{fig: grid_2d_graph_plot} show the largest soft assignment for each node when tasked with clustering a 2D ring graph and a 2D grid graph, respectively. The color of the node is chosen such that a sharp assignment of 1 gives a bright color (lightness equal to 0.5), while smoother assignments give paler colors, and an assignment of 0 is just white (lightness equal to 1). The number of desired clusters $K$ for the ring and grid was 5 and 10, respectively. The models were trained using the same hyperparameters as for the vertex classification task in the experiments. 

Again we see that DiffPool gives smooth assignments resulting in noticeably paler colors. In MinCutPool and DMoN we observe paler colors in the proximity of the cluster borders, while TVGNN exhibits sharp transitions from one cluster to the other. 
We also notice that TVGNN is the only method that on the grid generates a partition with the desired number of clusters $K=10$.

To better quantify the sharpness of the transition between different clusters, in Fig.~\ref{fig: largest_soft_assignment}a we show the largest value in the soft assignment vectors for the ring when moving along it. 
Here, the differences in cluster transitions are clear: the drops in the assignment value indicate the presence of smooth transitions. 
The cluster assignments of Diffpool are always very smooth; MinCutPool and DMoN exhibit smooth assignments only when crossing from one cluster to the other; with TVGNN the assignments are always sharp.

A similar plot for the grid is presented in Fig.~\ref{fig: largest_soft_assignment}b, but here the largest soft assignments for all nodes are sorted from lowest to highest, which indicates the overall proportion of smooth assignments.   
Also in this case, the cluster assignments of TVGNN are the sharpest, followed by DMoN, MinCutPool, and Diffpool.

\end{document}